\documentclass[pagesize=auto,bibliography=totocnumbered]{scrbook}
\usepackage[paperwidth=6.36in, paperheight=9in, margin=1in]{geometry}
\usepackage{lmodern} 

\usepackage[onehalfspacing]{setspace} 
\usepackage[hidelinks]{hyperref} 

\usepackage[T1]{fontenc}  
\usepackage{lmodern}      

\usepackage{tocloft}
\cftsetindents{chapter}{0em}{1.6em}   
\cftsetindents{section}{1.6em}{1.0em} 

\usepackage{csquotes}
\usepackage{graphicx}
\usepackage{lipsum}
\usepackage{appendix}
\usepackage{comment}
\usepackage{enumitem}
\usepackage{tabularx}
\usepackage{booktabs}
\usepackage{fancyvrb} 
\usepackage{tikz}
\usepackage{multirow}
\usepackage{amsthm}
\usepackage{amsmath} 
\usepackage{rotating}
\usepackage{flushend}
\usepackage{longtable}
\usepackage{caption}
\usepackage{scrhack}
\usepackage{titletoc}
\usepackage{algorithm}
\usepackage{algorithmic}
\usepackage{subcaption}

\newcommand{\CCM}{\mathsf{CoCoMo}}
\newcommand{\Rafel}{\mathsf{RAFEL}}
\newcommand{\SocraSynth}{\mathsf{SocraSynth}}
\newcommand{\EVINCE}{\mathsf{EVINCE}}
\newcommand{\DIKE}{\mathsf{DIKE}}
\newcommand{\ERIS}{\mathsf{ERIS}}
\newcommand{\CRIT}{\mathsf{CRIT}}
\newcommand{\LCI}{\mathsf{LCI}}
\newcommand{\MACI}{\mathsf{MACI}}
\newcommand{\RLDF}{\mathsf{RLDF}}
\newcommand{\BEAM}{\mathsf{BEAM}}
\newcommand{\ALAS}{\mathsf{ALAS}}

\newcommand{\SAGA}{\mathsf{SagaLLM}}

\newenvironment{noindlist}
 {\begin{list}{\labelitemi}{\leftmargin=0.3em \itemindent=0.6em}}
 {\end{list}}

 \newenvironment{noindenumerate}
  {\begin{list}{\arabic{enumi}.}
    {\usecounter{enumi}
     \setlength{\labelwidth}{0.6cm}
     \setlength{\leftmargin}{0.33cm}
     \setlength{\labelsep}{0.1cm}
     \setlength{\itemindent}{0cm}
     }
  }
  {\end{list}}

\definecolor{gold}{RGB}{255, 215, 0}

\begin{document}
\thispagestyle{empty} 
\begin{center}
\vspace*{\stretch{1}}
\includegraphics[width=1.0\textwidth,height=1.56\textwidth]{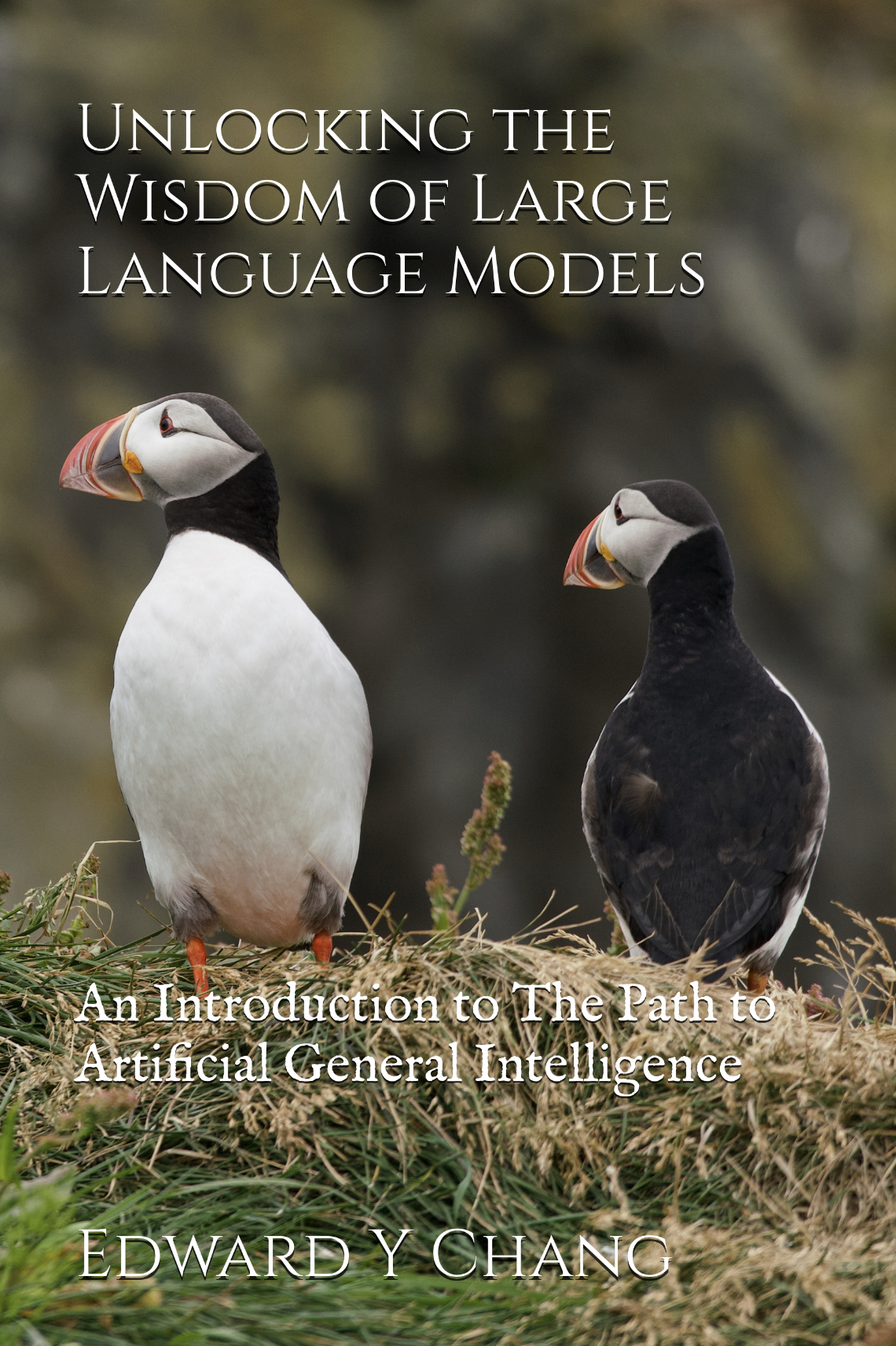}
\vspace*{\stretch{2}}
\end{center}
\clearpage 
\thispagestyle{empty} 

\begin{titlepage}
\thispagestyle{empty} 
\begin{center}
\vspace*{\stretch{1}}

{\bfseries\LARGE Unlocking the Wisdom of \\Large Language Models} \\
\vspace{0.8cm}
{\bfseries\Large An Introduction to \\the Path to Artificial General Intelligence} \\
\vspace{0.1cm}
{\bfseries\large Third Edition}

\date{\today}
\vspace{5cm} 
{\Large Edward Y. Chang \par} 
\vspace{.5cm} 
{\large Computer Science \par}
{\large Stanford University \par} 
\vspace{\stretch{2}}
\end{center}
\end{titlepage}

\thispagestyle{empty} 
\frontmatter

\newpage 
\newpage
\thispagestyle{empty} 

\frontmatter
\newpage
\chapter*{Preface}
\addcontentsline{toc}{chapter}{Preface}
\thispagestyle{empty}

This booklet, ``\textit{Unlocking the Wisdom of Large Language Models (LLMs)},'' serves as an introduction to the comprehensive work ``\textit{Multi-LLM Agent Collaborative Intelligence: The Path to Artificial General Intelligence}'' \cite{chang2024maci}. Through fourteen carefully crafted aphorisms, we distill the core insights and principles that underpin the broader exploration of AI's future through \textbf{Multi-LLM Agent Collaborative Intelligence} ($\MACI$), a framework we propose as a promising pathway toward realizing \textbf{Artificial General Intelligence} (\textbf{AGI}).

In addition to these insights, the booklet provides the titles, abstracts, and introductions of the main chapters of the book and presents the first two chapters in their entirety.

What follows is the original preface of ``\textit{Multi-LLM Collaborative Intelligence: The Path to Artificial General Intelligence},'' offers a critical context for the journey ahead. We are thrilled to announce the release of the third edition of the book this week, featuring groundbreaking additions of Chapters 11 and 12 on $\ALAS$ and $\SAGA$, revolutionary systems designed to address the challenges of long-lived workflows for real-world applications. These chapters confront fundamental limitations in AI systems: G\"{o}del's constraints on self-validation, context narrowing over extended operations, and the need for transaction guarantees with persistent memory in complex processes.

Critics have expressed skepticism about the ability of LLMs to achieve AGI, citing limitations in memory, planning, and world grounding. However, we present compelling evidence that $\MACI$, especially with multiple multimodal LLMs working together, provides a transformative architecture to overcome these challenges and forge the path towards AGI. The chapters on $\SocraSynth$, $\EVINCE$, $\CRIT$, behavior modeling, consciousness modeling, and ethical alignment through our checks and balances architecture---which assigns executive, legislative, and judicial roles inspired by a three-branch system of government---demonstrate that through collaborative intelligence, LLMs can achieve levels of reasoning that a single LLM cannot reach alone. Now, we have further strengthened this framework with $\ALAS$ and $\SAGA$ to systematically address self-validation and persistent memory challenges, two critical barriers identified by AGI skeptics.

By fostering collaboration through structured dialogue, $\MACI$ enables individual LLMs to complement each other's strengths, generating insights that no single model could achieve in isolation. This approach mirrors how human institutions balance diverse perspectives to produce breakthrough innovations and discoveries.

We can further extend this concept by designing one LLM to construct a virtual ``world model'' by mapping conceptual relationships and another to act as a super sensor, leveraging instruments to detect signals beyond human sensory capabilities. Together, these specialized LLMs can model and perceive reality in ways that transcend individual human limitations, opening entirely new frontiers of understanding.

Time will tell if this collaborative framework will bring us closer to AGI, but we firmly believe that the path forward lies in cooperation, not isolation. Just as collective intelligence has propelled human progress throughout history, collaborative LLMs may unlock new frontiers of artificial intelligence that no single model can achieve alone and potentially transforming how we address humanity's most pressing challenges.
\vspace{1cm}
\newline \newline
\noindent{Edward Y. Chang,} \\ 
\noindent{August $15^{th}$, 2024.} \\
\noindent{October $28^{th}$, 2024 (second edition).} \\
\noindent{March $31^{st}$, 2025 (third edition).} \\
\noindent{Series {\color{red}$\pi$} 0020131G003}
\clearpage

\section*{Preface to ``LLM Collaborative Intelligence: The Path to Artificial General Intelligence''}

As Generative AI transforms our world, experts predict the emergence of Artificial General Intelligence (AGI) as early as 2040. This book proposes that the key to achieving AGI, characterized by versatility, adaptability, reasoning, critical thinking, planning, and ethical alignment, lies not in creating more powerful individual models, but in enabling large language models (LLMs) to engage in intelligent and collaborative dialogue. This concept, termed \textbf{Multi-LLM Agent Collaborative Intelligence} ($\MACI$) forms the foundation of our exploration.

$\MACI$ transcends conventional ``mixture of experts'' models or traditional LLM debates by optimizing the exchange of information between LLM agents through seven essential foundations.

\begin{enumerate}[left=0em,topsep=-.0em, parsep=-.05em,label=\arabic*.]
\item \textbf{Balancing Exploration and Exploitation}:
$\MACI$ takes advantage of various perspectives while maintaining robust reasoning rooted in the strong priors of next-token maximum likelihood predictions. This ensures the exploration of novel perspectives while preserving the stability of the model parameters learned from the training data. (Chapters 6 and 7)

\item \textbf{Modulating Linguistic Behavior}:
Beyond balancing exploration and exploitation, $\MACI$ dynamically modulates the linguistic behaviors of LLM agents, facilitating transitions between contention and collaboration. By fostering productive perspective exchanges quantified through information theory metrics, $\MACI$ generates diverse viewpoints and novel insights. Calibrating debate contentiousness enables $\MACI$ to explore a broader spectrum of ideas while consistently converging on well-reasoned conclusions. Beyond contentiousness modulation, $\MACI$ detects and regulates behavior across multiple dimensions, including hate speech and empathy, anxiety and calmness, and emotional extremes---ensuring discussions remain productive and ethical while preserving cognitive diversity. (Chapters 9 and 10)

\item \textbf{Checks and Balances for Context-Sensitive Ethical Alignment}:
Individual LLM agents are assigned specialized roles: the Executive for knowledge formation, the Legislative for developing ethical frameworks, and the Judicial for contextual evaluation. This governance structure allows ethics to be legislated while enabling context-dependent interpretation by the judicial branch. The parameters of the executive LLMs remain unchanged to preserve performance integrity. (Chapter 10)

\item \textbf{Persistent Memory as Essential Infrastructure}: LLM agents require persistent memory systems that transcend their limited context windows. Without robust state persistence, agents cannot reliably maintain awareness of constraints, track system states, or execute complex workflows. Even sophisticated validation logic fails without complete historical context, and transaction patterns become meaningless when systems cannot accurately recall previous states. Persistent memory forms the critical foundation upon which both effective validation and reliable transaction guarantees must be built, enabling agents to operate coherently across extended temporal sequences. (Chapters 11 and 12)

\item \textbf{Reasoning with the Socratic Method}: 
$\MACI$ employs the Socratic Method to refine reasoning through iterative questioning and dialogue. This approach fosters deeper analysis, challenges assumptions, and enables the discovery of robust solutions by encouraging critical thinking among agents. (Chapter 5)

\item \textbf{Frontier Discovery through Polydisciplinary Synthesis}: LLMs are trained without informing the transformer algorithms about document domains. LLM training treats all documents as sequences of words, resulting in representations without domain boundaries where all fields are combined in one unified space. This polydisciplinary representation provides an unprecedented opportunity to synthesize knowledge across traditional boundaries and explore uncharted intellectual territories. It is at these intersections of previously siloed domains where true ingenious intelligence emerges, revolutionizing our understanding and unleashing unprecedented capabilities. (Chapter 15)
\end{enumerate}

Building on these foundations, the fifteen chapters guide the reader systematically from theoretical frameworks to practical applications. Key algorithms include $\CRIT$ for critical evaluation, $\SocraSynth$ for dynamic dialogues, $\EVINCE$ to optimize the flow of information through Bayesian statistics and information theory, $\SAGA$ to implement persistent memory systems that enable robust validation and preservation of transaction properties, and polydisciplinary synthesis to explore knowledge across traditional boundaries. We explore applications ranging from medical diagnosis to news debiasing and address fundamental challenges in AI safety through checks and balances applied across LLM's linguistic, legislative, and judicial modules, ensuring alignment between knowledge, ethics, and contextual reasoning.

\subsection*{Reframing LLMs as Components in the Path to AGI}

Some pioneers of modern AI, including Yann LeCun, argue that LLMs alone cannot achieve AGI due to several fundamental limitations: their lack of persistent memory, reasoning and planning capabilities, and physical grounding. LeCun specifically asserts that true intelligence requires interaction with the physical world through sensors and embodiment. He emphasizes that while LLMs demonstrate impressive linguistic capabilities, they lack genuine understanding and even suggests that ``the sensory-motor abilities of a cat surpass those of an LLM.'' This perspective raises legitimate concerns about current LLM architectures, particularly regarding the efficiency with which humans learn complex skills from few examples, something current systems struggle to replicate.
In this book, we reframe this debate by examining how LLMs might serve not as standalone AGI systems but as crucial components within a more comprehensive architecture.

\paragraph{Reframing LLMs' Role in Intelligence}

Rather than viewing LLMs as standalone forms of artificial general intelligence, we propose reframing them as foundational components within a broader intelligent system. In biological intelligence, consciousness operates on top of an extensive underlying architecture of neural processes (the unconscious) responsible for perception, motor control, memory consolidation, and other essential life functions.

Similarly, LLMs can function as \emph{cognitive substrate} within a modular intelligence framework, responsible for processing, pattern matching, and information synthesis. The external modules can then manage reasoning, validation, planning, and behavior modulation. Although not identical to human unconsciousness, LLMs exhibit functionally analogous capabilities: recognizing patterns, forming associations, and constructing representations that support more sophisticated conscious operations when properly structured.

Critically, this reframing highlights a profound parallel: Human intelligence does not begin as a blank slate. Infants enter the world equipped with evolutionarily encoded unconscious mechanisms that govern breathing, metabolism, coordination, and even basic social instincts. Conscious learning through few-shot experiences and fine-tuning emerges upon this biological foundation. Therefore, comparing an infant's intelligence directly to an LLM is misleading; the infant benefits from billions of years of embedded evolutionary knowledge. Likewise, LLMs provide \emph{artificial unconsciousness}, a foundation on which modular systems can develop structured, adaptive, and goal-directed intelligence.
(Chapter 13 provides an in-depth discussion.)

\paragraph{Addressing the Grounding Problem}
The claim that LLMs lack a world model stems from a particular interpretation of what constitutes ``grounding'' in intelligence. We support Ilya Sutskever's perspective that documents themselves encapsulate worldviews, and history is fundamentally written in text. Text represents a projection of the world, a distillation of human experience, observation, and reasoning encoded in language.

When LLMs train on vast corpora, they learn to compress and represent the processes that generate these texts. This compression captures significant aspects of human understanding, including causal relationships, physical properties, and social dynamics. Although this text-based grounding differs from sensorimotor grounding, it represents a valuable form of knowledge acquisition that should not be dismissed.

That said, we acknowledge that multimodal perception provides complementary forms of grounding that can enhance an AI system's understanding. Rather than viewing sensory input as the exclusive basis for intelligence, we see it as one of multiple channels through which a system can build comprehensive world models.

\paragraph{The Memory Challenge}
Current LLMs face two distinct memory-related challenges: context window limitations during inference and the lack of persistent memory between interactions. The first constrains how much information an LLM can consider at once, while the second prevents the accumulation of experience over time.
Context window expansions have addressed some limitations, but issues surrounding context loss and context narrowing have caused unexpected inconsistencies in long-lived transactions or long chains of thoughts. Persistent memory integration remains crucial for capabilities like constraint validation, state tracking for action rollback, and maintaining consistency across complex workflows. The $\SAGA$ framework introduced in Chapter 12 specifically addresses these challenges by adapting saga-transaction mechanisms from database systems to manage memory in LLM-based agents.

$\SAGA$ enables selective storage and retrieval of critical information, maintaining a structured record of past states, actions, and their consequences. This allows for a precise rollback when errors occur and ensures that interdependent processes maintain consistency. In our experiments with planning scenarios such as the Thanksgiving dinner coordination problem, $\SAGA$ successfully maintained awareness of constraints that the standalone LLMs lost track of during extended reasoning.

\paragraph{Reasoning and Planning Capabilities}
Building on these memory advances, we now address the significant challenges LLMs face with reasoning and planning in complex, multi-step processes. Although persistent memory provides the foundation, effective reasoning requires additional architectural innovations.
The $\MACI$ framework leverages $\SAGA$'s memory capabilities while integrating specialized modules for different cognitive functions. By distributing reasoning across structured components, analogous to the executive, legislative, and judicial branches of government, $\MACI$ implements logical operations of checks and balances. This structure supports an iterative process ``think, validate, rethink'' where each step is recorded in persistent memory, allowing the system to continuously verify that all constraints are observed throughout the planning sequence.

Evidence from our implementation demonstrates that this architecture successfully maintains global constraints across extended planning horizons, effectively detects and corrects inconsistencies in proposed actions, and seamlessly integrates new information without losing critical context. The system's ability to reference previous states and decisions ensures that constraints established early in the reasoning process remain enforced even as plans become increasingly complex. Although these capabilities do not yet match human-level planning, they establish a promising path toward addressing current limitations through architectural innovation rather than relying solely on scaling existing models.

\paragraph{Synthesis: A Modular Path Forward}
The path toward more comprehensive artificial intelligence likely involves neither pure LLMs nor a complete abandonment of their capabilities. Instead, we envision a modular architecture in which LLMs serve as powerful semantic processors within a larger system that includes dedicated components for perception, memory management, causal reasoning, and action selection.

This framework addresses LeCun's concerns about embodiment and grounding while leveraging the linguistic and associative strengths of LLMs. By integrating sensor modules, persistent memory systems such as $\SAGA$, and specialized reasoning components within the $\MACI$ framework, we can build systems that combine the pattern matching capabilities of LLM with the physical interaction and persistent learning that critics rightly identify as essential to intelligence.

Rather than asking whether LLMs alone can achieve AGI, we should ask how their unique capabilities can complement other modalities in building more robust, adaptive, and trustworthy intelligent systems. The frameworks presented in this book offer concrete implementations of this vision, demonstrating how LLMs can serve as foundational components in AI systems that reason effectively, maintain memory consistently, and ground understanding in both language and perception.

The path to AGI through $\MACI$ is not a sudden leap but a gradual process of structured integration and collaborative evolution. Just as human civilization advances through the accumulation of shared knowledge and social negotiation, multimodal LLM agents can transcend their individual limitations by working together, each contributing to different modes, perspectives and roles of reasoning. This book provides the architectural blueprints and theoretical grounding to pursue this path, demonstrating how systems built on foundational substrates can evolve toward human-level reasoning and beyond.

To conclude this exploration, Chapter 1 offers a philosophical synthesis: Fourteen aphorisms distilled from practical experimentation and long-term research. These aphorisms reflect the core insights of $\MACI$ and articulate a new paradigm: one where intelligence is not defined by individual scale, but by the capacity for dialogue, regulation and rational alignment with complexity and uncertainty.
\newline 
\noindent{Edward Y. Chang} \\ 
\noindent{March $13^{th}$, 2024 (first edition)} \\
\noindent{October $28^{th}$, 2024 (second edition)} \\
\noindent{March $31^{st}$, 2025 (third edition for Stanford CS372)} \\
\noindent{Series {\color{red}$\pi$} 0020131G006}
\clearpage

\clearpage

\tableofcontents
\mainmatter

\chapter{Aphorisms of LLM Collaborative Intelligence}
\markboth{Aphorisms}{Aphorisms}
\label{chap:Aphorisms}

\section*{Introduction}

This chapter introduces fourteen aphorisms developed from five years of research into multi-agent Large Language Models (LLMs). These aphorisms explain and justify the foundations of $\MACI$ (Multi-LLM Agent Collaborative Intelligence), an architecture that organizes multiple LLMs to support reasoning, planning, and decision-making beyond the capacity of single models. Grounded in philosophical traditions and implemented through modular frameworks introduced earlier, $\MACI$ views intelligence as dialogic, evaluative, and regulative. Rather than scaling individual models in isolation, $\MACI$ coordinates collaboration across diverse perspectives to enable verifiable and context-aware behavior, offering a potential path toward AGI.

The fourteen aphorisms are organized into four conceptual layers that reflect $\MACI$’s development from theoretical insights to practical orchestration. These include foundational concepts about truth and validation (Aphorisms 1–4), system-level design principles (Aphorisms 5–8), collaborative reasoning strategies (Aphorisms 9–12), and emergent intelligence mechanisms (Aphorisms 13–14).

Although grounded in earlier frameworks including $\CCM$ \cite{chang2023cocomo}, \\ $\SocraSynth$ \cite{SocraSynthChangCSCI2023}, $\CRIT$ \cite{SocraticIEEECCWC2023}, $\EVINCE$ \cite{EVINCEChang2024}, $\DIKE$-$\ERIS$ \cite{chang2024NeurIPS}, and $\SAGA$ \cite{chang2025sagallm}, this chapter refers to them only when illustrative. The focus is on the design philosophy they collectively support: that collaborative intelligence, not isolated pattern generation, is the basis of general intelligence.

Each aphorism encapsulates a principle grounded in empirical system development, from questions about epistemology and bias to applied mechanisms for role assignment, linguistic behavior modulation, and validation. Together, they offer a comprehensive framework for understanding and building multi-agent  systems.

The fourteen aphorisms are grouped by conceptual layer, reflecting $\MACI$'s progression from theoretical principles to system design and emergent intelligence. Table~\ref{tab:aphorism-map} summarizes their organization.

\begin{table}[th!]
\centering
\footnotesize
\begin{tabular}{|p{2.5cm}|p{8cm}|}
\hline
\textbf{Category} & \textbf{Aphorisms} \\
\toprule
\hline
\textbf{Theoretical} & \#1: Truth emerges from perspectives \\ 
\textbf{Foundations} & \#2: External mirrors enable validation \\ 
                                  & \#3: LLMs faking reasoning and feelings, just as we do \\ 
                                  & \#4: Foundations and adaptations \\
\hline
\textbf{System}           & \#5: Context transforms capabilities \\ 
\textbf{Design}           & \#6: Linguistic behavior reflects intention and emotions \\ 
                                  & \#7: Checks and balances ensure adaptive alignment \\ 
                                  & \#8: Consciousness filters impulse; $\MACI$ governs LLMs \\
\hline
\textbf{Collaborative}  & \#9: The essence of precise questioning \\ 
\textbf{Dynamics}       & \#10: Debate strengthens reasoning quality \\ 
                                  & \#11: Hallucinations fade under scrutiny \\ 
                                  & \#12: Polydisciplinary synthesis expands discovery \\
\hline
\textbf{Emergent Intelligence}   & \#13: Human heuristics are limited and end-to-end learning shows supremacy \\ 
                                  & \#14: AGI emerges through collaborative intelligence \\
\hline
\end{tabular}
\caption{Organizational Map of the 14 Aphorisms of $\MACI$}
\label{tab:aphorism-map}
\end{table}


\section*{Aph. \#1: Truth emerges from perspectives} \noindent \textit{``Outside formal systems, objectivity remains a hard problem; what we pursue instead is reasonableness through multiple perspectives.''}
\addcontentsline{toc}{section}{Aph. \#1: Truth emerges from perspectives}

Mathematics and logic offer certainty through formal proofs, but most real-world domains, such as medicine, law, journalism, operate under uncertainty and competing interpretations. In these settings, the objective truth is often inaccessible. What we can instead pursue is reasonableness, a standard grounded in coherence, evidence, and comparative judgment.

Philosophical traditions have long wrestled with this challenge. Kant distinguished between phenomena (what we perceive) and noumena (things-in-themselves), underscoring how cognition filters all experience \cite{Kant1781Critique}. Nietzsche argued that facts are always shaped by interpretation \cite{Nietzsche1968WillPower}. In practice, objectivity is rarely found; reasonableness is our proxy.

This insight matters in domains where even the ``ground truth'' can be flawed. A 2023 Johns Hopkins study estimated a 10\% misdiagnosis rate in US hospitals, revealing the limits of clinical certainty \cite{NewmanToker2023}. Similarly, studies of media bias show how divergent expressions influence public understanding \cite{chang2024uncoveringbiases}.

To navigate these uncertainties, MACI fosters dialogue among diverse reasoning agents. Each agent contributes a different perspective, that is, assessing, challenging, or reframing claims. Through structured interaction and principled disagreement, the system refines its understanding. Reasonableness emerges not from any one voice, but from how multiple voices collaborate with each other.

This process is made tractable through MACI's internal evaluation methods, which assess arguments for coherence, evidentiary grounding, and logical structure. Rather than assuming correctness, the system measures how well each claim withstands scrutiny from alternative perspectives.

MACI does not seek to resolve the philosophical problem of objectivity. Instead, it builds a practical methodology around comparative reasoning. By treating disagreement as a resource and not a failure, it moves closer to reasonableness, an ideal more aligned with how humans actually make sense of the world.


\section*{Aph. \#2: External mirrors enable validation}
\noindent \textit{``No system can fully validate its own reasoning from within, a fundamental limitation shared by monolithic LLMs and human minds, as Gödel's incompleteness theorems reveal.''}
\addcontentsline{toc}{section}{Aph. \#2: External mirrors enable validation}

Both humans and machines face a core limitation: we struggle to detect our own errors. People often miss inconsistencies or typos in their own writing, but spot them easily in others'. Language models face a similar challenge. When operating in isolation, a model cannot reliably assess the soundness of its own output.

This challenge has technical underpinnings. Researchers have identified that the stochastic nature of next-token prediction, conditioned on evolving context, can lead to degradation in reasoning quality. Effects such as effective long-context loss \cite{liu-etal-2024-lost}, attention narrowing and distraction \cite{zhu2025focusdirectionsmakelanguage}, and context override through state transitions \cite{yen2024longcontext} can cause the model to diverge from earlier logic or contradict prior statements without awareness. Without an external reference point, these failures often go uncorrected.

This limitation is mirrored in foundational theories. Gödel's incompleteness theorems show that complex systems cannot establish all truths from within themselves \cite{godel1931english}. Wittgenstein emphasized that meaning arises from public language practices, not internal solipsism \cite{wittgenstein1953philosophical}. And Popper noted that science advances through external falsification, not internal confirmation \cite{popper1959logic}. 

Together, these insights point to a common conclusion: Validation requires independence. Sound reasoning is rarely self-contained. It emerges from critique, comparison, and testing across different viewpoints.

MACI addresses this constraint through a distributed validation system. Rather than relying on self-consistency, MACI enables agents to serve as external mirrors for each other. Each model contributes both its own judgment and its evaluation of others. Reasoning becomes more stable and accountable when multiple agents examine claims from independent points of view.

Progress in AI may depend less on increasing internal complexity and more on structuring systems that can challenge, reflect, and revise through collaboration. MACI provides the architecture to support this, making validation not a post-processing step, but an integral function of interaction.

Attaining greater intelligence may not require larger models, but better mirrors.


\section*{Aph. \#3: Simulated reasoning can be useful, even if not real}
\noindent \textit{``An actor may not feel true sorrow when portraying grief, yet their performance can still move us. The question is not whether the emotion is real, but whether it functions in context.''}
\addcontentsline{toc}{section}{Aph. \#3: Simulated reasoning can be useful, even not real}

Language models do not reason or feel the way humans do. They have no subjective experience, emotional memory, or intent. Yet they can often produce outputs that appear reasoned or emotionally expressive. This gap, between form and inner state, raises important questions about what it means to simulate intelligence.

Turing's 1950 formulation suggested that indistinguishability in conversation might be sufficient to attribute intelligence \cite{turing1950computing}. Although this view has been criticized, most notably by Searle's Chinese Room argument, it nevertheless shifted the focus to behavior over internal structure. In practice, the utility of a system often depends not on how it works internally but on how its outputs function in a broader interaction.

The analogy to acting serves to illustrate this distinction. A skilled actor does not experience the grief that they portray, yet their performance may be compelling and meaningful. Of course, actors draw on lived emotional experience and possess intentionality. LLMs do not. The analogy holds not in terms of internal motivation but in how we evaluate the product of a performance: by its coherence, appropriateness, and effect.

Likewise, Erving Goffman's theory of social performance described how people change roles based on context and audience expectations \cite{goffman1959presentation}. Humans adapt strategically and with awareness. LLMs lack such metacognitive control. But their outputs can still be modulated—externally and systematically—to adopt different roles, such as analyst, critic, or teacher.

This is where MACI provides architectural clarity. It does not treat LLMs as conscious agents. Instead, it uses their capacity for simulated role enactment within a structured environment. Roles are assigned, perspectives are coordinated, and outputs are checked by other agents. This process does not generate genuine understanding, but can still support useful reasoning behavior when mediated by architectural safeguards.

Simulation, in this view, is not deception if it is constrained, transparent, and subject to verification. MACI's purpose is not to pass as human, but to leverage the expressive potential of LLMs in a way that supports deliberation, critique, and adaptive planning.

Treating simulated reasoning as real reasoning would be a mistake. But rejecting simulation outright would ignore its functional value, especially when structured through external moderation and multi-agent feedback. MACI acknowledges this distinction and builds on it. The intelligence that emerges is not located in any one model, but in the interaction between them.


\section*{Aph. \#4: Foundations and adaptations} 
\noindent \textit{``Intelligence operates on dual layers: a data-intensive foundation akin to unconscious processes and an agile contextual layer resembling conscious adaptation.''}\\
\addcontentsline{toc}{section}{Aph. \#4: Foundations and adaptations}
\smallskip

Critics often observe that LLMs require huge amounts of training data, while humans can learn from just a few examples. But this contrast overlooks a structural similarity: both human and machine intelligence operate on two interdependent layers: a broad foundational substrate and a flexible adaptive overlay.

In humans, the foundational layer consists of perceptual and emotional responses shaped by evolution and experience. Freud described this as the unconscious mind, a deep reservoir beneath awareness \cite{freud1915unconscious, eagleman2011incognito}. Kahneman's ``system 1'' captures this further as fast, intuitive, and automatic thinking \cite{kahneman2011thinking}. On top of this sits ``system 2'', a slower, more effort-intensive, and deliberative reasoning layer used when pattern matching alone is insufficient.

Similarly, LLMs are trained on massive corpora to form a statistical base of linguistic fluency. This base supports fast generation in familiar contexts. However, when adaptation is needed—such as answering novel questions, reconciling competing interpretations, or reasoning through ambiguity—the model's behavior must be shaped by external inputs: prompts, instructions, role constraints, or collaborative feedback.

Here, MACI contributes a distinct architectural layer. Modern LLMs already separate training from deployment-time behavior modulation; MACI formalizes and extends this adaptive layer by coordinating multiple agents with complementary roles. It does not alter the weights of the foundational model, but instead adds mechanisms for critique, revision, and convergence. Through inter-agent interaction, moderated dialogue, role-switching, and counter argumentation, MACI builds an adaptive scaffolding that simulates deliberation and contextual reasoning atop static models.

This dual-layer design helps resolve the apparent contradiction between data-hungry models and human-like flexibility. Neither humans nor machines reason from scratch. They reason on top of deeply embedded patterns—refining, recombining, and recontextualizing them when the situation demands. Although the mechanisms and development paths differ significantly between biological and artificial systems, the functional parallel remains instructive.

Understanding this layered architecture reframes the debate. Intelligence is not a monolithic capability, but a system of foundations and adaptations. MACI's distinctive contribution lies in how it coordinates both: it preserves the efficiency of foundational LLM knowledge while adding structured mechanisms for critique, verification, and collaborative reasoning that enhance the adaptation layer beyond what individual models can achieve alone.


\section*{Aph. \#5: Context transforms capabilities}  
\noindent \textit{``Strength and weakness in an LLM are not fixed traits, but fluid, shifting with context. MACI enables models to transcend training biases and adopt new roles through structured interaction.''}  
\addcontentsline{toc}{section}{Aph. \#5: Context transforms capabilities}

Language models are often evaluated in terms of static competencies. They are said to be ``good'' at summarization, ``weak'' at logical inference, or prone to certain biases. However, this evaluation misses a deeper principle: \textit{LLM behavior is context-sensitive}. The same model can produce divergent results, depending on how it is prompted, instructed, or situated within a task.

This flexibility arises from the way LLMs generate responses. Without stable goals or persistent beliefs, they respond to cues embedded in context. Although this can lead to inconsistency or overfitting to recent tokens, it also allows \textit{contextual modulation}, a process that alters behavior not through retraining but by shaping the context.

$\MACI$ turns this property into a core design principle. It treats context not as a constraint but as a \textit{controllable layer of adaptation}. Through role assignment, dialogic scaffolding, and structured disagreement, $\MACI$ coordinates agents to respond differently than they would in isolation. In this sense, context becomes a functional \textit{design surface}—not only directing behavior but configuring the reasoning dynamics of the system itself.

Three principles illustrate this transformation:

\begin{enumerate}
    \item \textit{Bias is contextual.} A model that defaults to caution may suppress novel ideas in exploratory tasks but increase safety in medical settings. $\MACI$ manages these tradeoffs by assigning complementary roles and balancing behavior through interaction.

    \item \textit{Adaptation is interactional.} LLMs change tone, focus, and strategy when prompted to critique others, defend a viewpoint, or explore counterfactuals. $\MACI$ leverages these dynamics to increase the diversity of reasoning and improve convergence through dialogue.

    \item \textit{Capability is conditional.} Rather than retraining models for fixed competencies, $\MACI$ reuses the same model across multiple roles—analyst, skeptic, reconciler—each invoked by a distinct interaction pattern or instruction format. This modular role-switching reveals capacities that remain latent in static use.
\end{enumerate}

These behaviors are measurable. Prior work shows that contentiousness levels, modulated through language patterns and role assignments, affect the outcome of multi-agent reasoning \cite{SocraSynthChangCSCI2023}. Information-theoretic metrics such as mutual information and entropy reduction have been used to guide the transition from contention to consensus \cite{EVINCEChang2024}.

For example, in a medical triage task, two LLM agents begin with divergent diagnoses. One emphasizes common symptoms, the other focuses on rare but high-risk conditions. They engage in multiple rounds of debate. Each round is moderated using entropy and relevance metrics. The agents critique and revise their positions. Eventually, they arrive at an agreement with thorough research and reasoning. This outcome is not encoded in either model individually. It emerges from how the \textit{contextual modulation} system organizes their interaction.

This reframes the concept of ``capability.'' It is no longer an intrinsic property of the model, but a \textit{systemic effect shaped by contextual control}. $\MACI$ does not assume that any one model is optimal. It assumes that models, placed in the right structure, can complement and challenge one another in ways that expand their usable capabilities.

This aphorism builds directly on Aphorism \#4, which describes the two-layer structure of intelligence. Here, the \textit{adaptation layer is made operational}—not through internal model change but through external coordination and context design. What the model does depends not only on what it has learned, but on how the system engages it.

\paragraph{Note:}
Aphorism \#5 alone is necessary but insufficient to foster effective multi-agent debate. 
It works in concert with Aphorism \#6 (Linguistic behavior reflects intention and emotions) and 
Aphorism \#11 (Debate strengthens reasoning quality) to create a complete framework. 
Together, they ensure an optimal balance between exploring new perspectives and
exploiting strong priors, between persuasive and conciliatory tones, and 
maintaining high-quality reasoning throughout the dialectical process.


\section*{Aph. \#6: Linguistic behavior reflects intention and emotions}
\noindent \textit{``LLMs are designed and trained to emulate human linguistic endeavors, each aimed at fulfilling distinct human objectives.''} \\
\addcontentsline{toc}{section}{Aph. \#6: Ling. behavior reflects intention and emotions}

Human language is rarely neutral; we write with intention, to argue, inform, or express. LLMs, trained on such purposeful language, inherit these patterns. Although they lack internal states, they can emulate the forms and rhetorical strategies associated with various roles.

Consider these roles:
\begin{itemize}[leftmargin=1.0em, topsep=-.05em, parsep=-.05em]
    \item \textit{Journalist}: Presents coherent, factual narratives;
    \item \textit{Debater}: Constructs and defends arguments;
    \item \textit{Analyst}: Dissects problems and identifies patterns;
    \item \textit{Mediator}: Synthesizes divergent viewpoints.
\end{itemize}

$\MACI$ assigns these roles intentionally and uses structured prompts and interaction constraints to ensure that agents embody distinct linguistic strategies. For example, in multi-agent debates, contentiousness is modeled by adjusting tone, emphasis, and word choice \cite{SocraSynthChangCSCI2023}.

\begin{table}[t!]
\centering
\includegraphics[width=0.98\linewidth, height=0.50\linewidth]{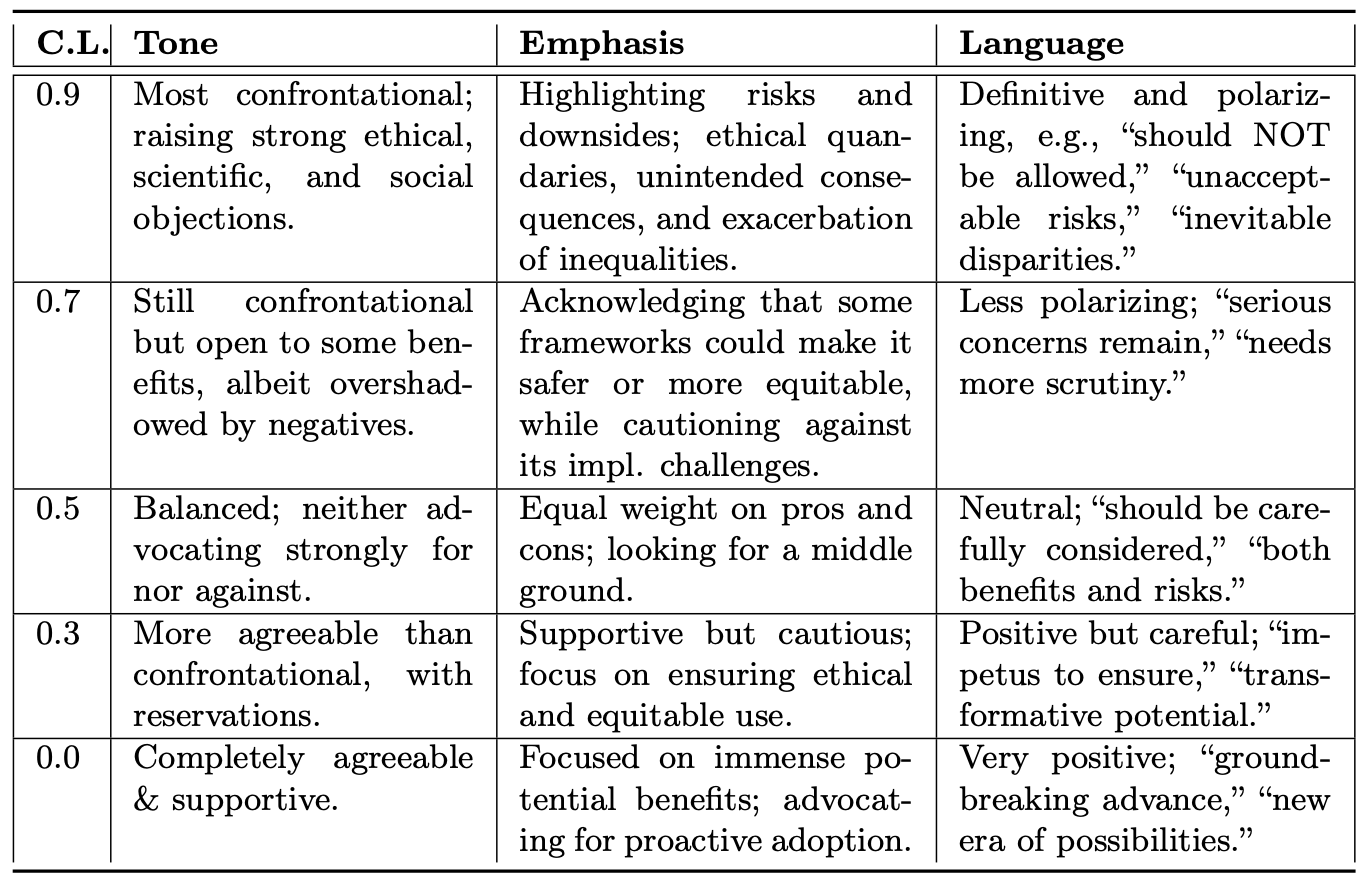}
\caption{The Linguistic Features of Varying Contentious Levels Modeled by GPT-4.}
\label{tab:chap14_contentiousness}
\end{table}

\begin{figure*}[th!]
\begin{center}
\includegraphics[width=0.98\linewidth]{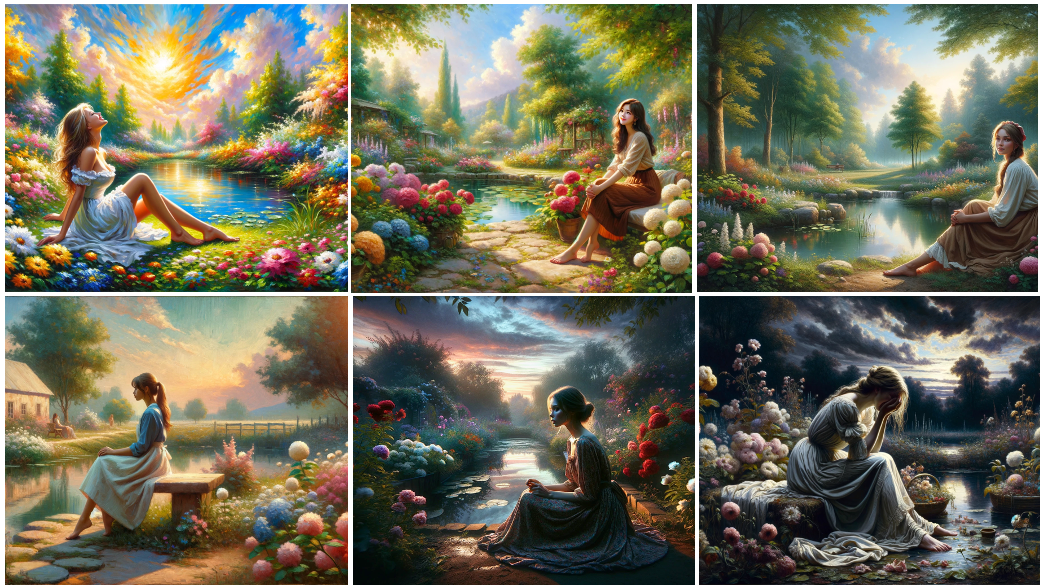}
\end{center}
\caption{A Lady and Garden Scene under Different Emotions. From top-left, happiest, to bottom-right, saddest.}
\label{fig:aph6-emotion-levels}
\end{figure*}

Beyond contentiousness, $\MACI$ also modulates emotional expression. Research \cite{chang2024MIPR} has shown that LLMs can be conditioned to generate outputs across affective spectrums, validated through both linguistic features and visual representations, as presented in Figure~\ref{fig:aph6-emotion-levels}.

This capacity to modulate linguistic behavior across roles and emotional states creates a bridge between human and machine communication. $\MACI$ orchestrates these behaviors not as mere mimicry, but as strategic coordination: assign complementary roles that reflect task requirements and organize agent interactions to support adaptive reasoning.

\paragraph{Connection:} Aphorism \#6 builds directly on Aphorism \#5's principle of contextual modulation, showing that linguistic behavior is the primary mechanism through which $\MACI$ operationalizes different agent roles and functions.


\section*{Aph. \#7: Checks and balances ensure adaptive alignment}
\noindent \textit{``Separating knowledge discovery, ethical oversight, and behavioral evaluation into distinct roles ensures a system of checks and balances, promoting adaptable AI safety and alignment with cultural norms.''}\\
\addcontentsline{toc}{section}{Aph. \#7: Checks and balances ensure adaptive alignment}

\begin{figure}[t!] 
\centering 
\includegraphics[width=0.98\linewidth, height=0.70\linewidth]{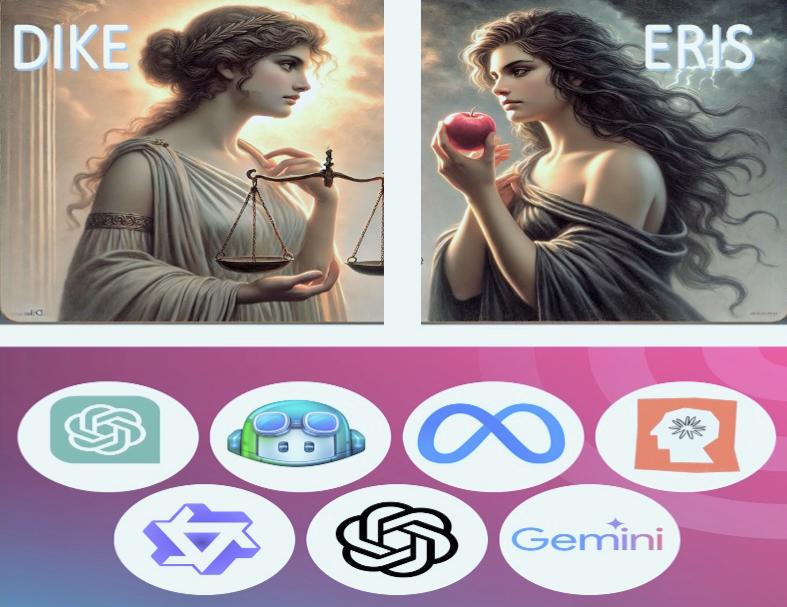} 
\caption{Three Framework Components: Executive LLMs (bottom), Legislative ($\DIKE$), and Judicial ($\ERIS$).} 
\label{aph:fig-alphorism7} 
\end{figure}

$\MACI$ employs a tripartite structure (Fig.~\ref{aph:fig-alphorism7}), inspired by constitutional governance \cite{montesquieu1748spirit, chang2024NeurIPS}, to manage adaptive alignment:
\begin{itemize}[leftmargin=1.0em, topsep=-.05em, parsep=-.05em] 
    \item \textbf{Executive}: Proposes knowledge, hypotheses, and solutions.
    \item \textbf{Legislative} ($\DIKE$): Defines ethical constraints and principles.
    \item \textbf{Judicial} ($\ERIS$): Contextualizes and critiques alignment through adversarial review.
\end{itemize}

This separation prevents ethical constraints from degrading task performance. Unlike RLHF, which can compromise core competencies through parameter modification based on aggregate feedback, $\MACI$ isolates knowledge generation from ethical evaluation \cite{chang2024NeurIPS}. Furthermore, by treating each instance independently through the judicial branch, $\MACI$ enables context-sensitive ethical alignment rather than applying averaged constraints that fail to account for individual situations and cultural or societal variations.

This architecture allows $\MACI$ to adapt across cultural settings and value conflicts by creating internal accountability structures. Disagreement is not a failure mode, but a mechanism for refinement.


\section*{Aph. \#8: Consciousness filters impulse; MACI governs LLMs}
\noindent \textit{``Our public behavior is not a direct, unfiltered output from our unconscious mind. Instead, consciousness regulates and refines the underlying impulses, ensuring that our behaviors are aligned with social norms. Similarly, $\MACI$ frameworks are designed to harness and temper the inherent tendencies of LLMs, mitigating their inherited biases\footnote{The analogy between consciousness and $\MACI$ is intended as a functional comparison, not a metaphysical claim. MACI does not instantiate awareness or qualia, but it implements structured regulatory mechanisms that resemble cognitive control..''}\\
\addcontentsline{toc}{section}{Aph. \#8: Consciousness filters impulse; MACI governs LLMs}}

While $\MACI$ does not implement consciousness, it performs an analogous function: it regulates and filters impulsive outputs through contextual, structural, and ethical constraints.

Neuroscientific theories such as Gazzaniga's ``interpreter'' model \cite{gazzaniga2011minds} and Baars' Global Workspace Theory \cite{baars2005global} describe consciousness as an aggregator and filter of unconscious processes. Similarly, $\MACI$ applies higher-order filters to LLM-generated outputs, performing behavioral modulation across:
\begin{itemize}[leftmargin=1.0em, topsep=-.05em, parsep=-.05em] 
    \item \textit{Inhibitory control:} Suppressing harmful, incoherent, or inappropriate content;
    \item \textit{Reappraisal:} Reframing outputs based on ethical, epistemic, or pragmatic framing;
    \item \textit{Contextualization:} Modulating tone, abstraction, and intent based on audience and task.
\end{itemize}

This cognitive architecture echoes earlier aphorisms: validation (\#2), contextual control (\#5), and institutional balancing (\#7). What $\MACI$ adds is not self-awareness, but \textit{intention regulation} of agent behavior. Like consciousness, it makes outputs suitable for public use—not merely fluent, but contextually appropriate.

\paragraph{Closing reflection:} The power of intelligence may lie not in what it generates, but in what it withholds. Regulation is not a constraint on intelligence—it is a precondition for its application.

\paragraph{Transition:} This regulatory capacity—framed cognitively in Aphorism \#8—prepares the ground for implementation strategies that operationalize collaborative intelligence through interaction, question framing, and convergence.


\section*{Aph. \#9: The essence of precise questioning}
\noindent \textit{``The essence lies in framing and sequencing the right questions.''} \\
\addcontentsline{toc}{section}{Aph. \#9: The essence of precise questioning}

This aphorism highlights the critical role of precise questioning in LLM interactions. The approach draws inspiration from the Socratic method, where knowledge emerges not through assertion but through systematic inquiry \cite{vlastos1991socrates}. Within $\MACI$, particularly through $\SocraSynth$ \cite{SocraSynthChangCSCI2023}, three principles emerge:

First, in multi-LLM debates, discourse quality depends on how LLMs challenge each other. Effective counterarguments serve as sophisticated questions that investigate assumptions, ask for evidence, and highlight inconsistencies, transforming the debate from opposition to collaborative inquiry. This embodies the essence of Socratic elenchus, using questions not to defeat but to discover.

Second, iterative interactions create dynamic exchanges in which each response refines the context for subsequent questions. This feedback loop enables progressive clarification: early inquiries may be vague or exploratory, but through sustained interaction, LLMs build increasingly precise and focused lines of inquiry. This iterative sharpening of intent is critical to converge on useful outcomes.

Third, based on this foundation, the principle extends beyond the formulation of questions to the sequencing of questions. Because sequencing acts as a basis for reasoning, determining the scope, order, and relevance of subsequent investigation, this third principle warrants closer examination. In $\MACI$, LLMs build on previous exchanges, using responses to inform and refine subsequent queries. This creates a structured approach to complex topics, enabling incremental discovery through a logical and coherent progression of inquiries.

\begin{figure}[t!]
  \centering
  \includegraphics[width=0.96\linewidth, height=0.36\linewidth]{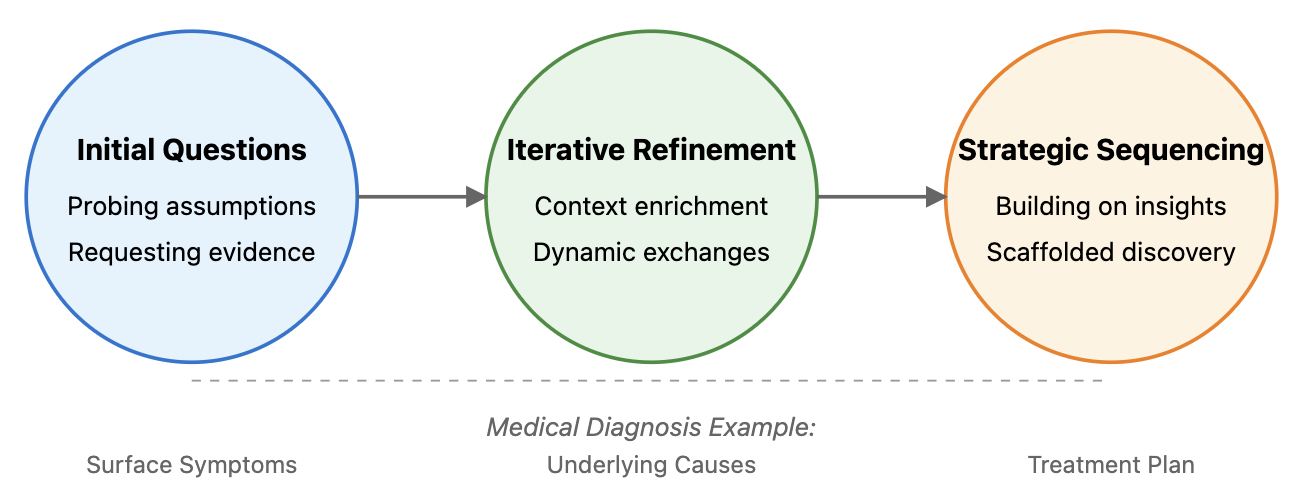}
  \caption{LLM Question Progression Framework.}
  \label{fig:aphorism10}
  \vspace{-.2in}
\end{figure}

This virtuous cycle of precise questioning, contextual enrichment, and strategic sequencing determines the quality and utility of the insights generated by LLMs.

\paragraph{\textbf{Example:}} A conversation that begins with a question about the symptoms of a disease can evolve into a diagnostic discussion as LLMs build on each other's insights. The discussion transitions from identifying surface-level symptoms to uncovering underlying causes, recognizing comorbidities, and generating actionable treatment recommendations \cite{EVINCEChang2024, SocraHealth2023}. To facilitate effective sequencing, moving from exploration of possible diseases to deeper probing of finalists, the conversation moderator adjusts the LLMs' linguistic behaviors, shifting from contentious to conciliatory tones based on several information-theory metrics, including cross-entropy, mutual information, and Jensen-Shannon divergence.

\paragraph{\textbf{Note:}} This aphorism works in concert with Aph. \#10 (Debate strengthens reasoning quality) and Aph. \#6 (Linguistic behavior reflects intention and emotions). Together, these principles demonstrate how the quality of questions, the structure of debate, and the modulation of linguistic behavior combine to create effective multi-agent reasoning systems.


\section*{Aph. \#10: Debate strengthens \\ reasoning quality}
\noindent \textit{``Critical thinking requires more than one Socrates.''}\\
\addcontentsline{toc}{section}{Aph. \#10: Debate strengthens reasoning quality}

\begin{figure}[t!]
  \centering
  \includegraphics[width=0.98\linewidth, height=0.70\linewidth]{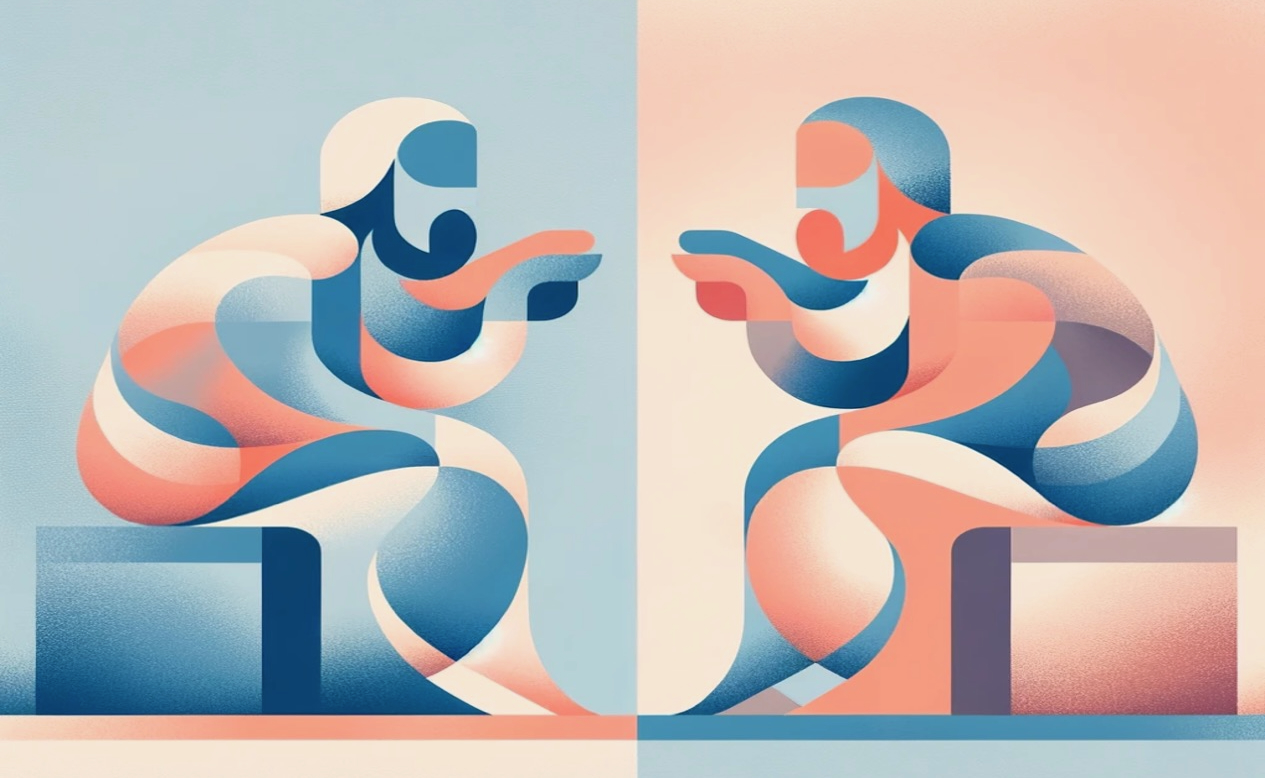}
  \caption{The $\MACI$ Framework with Two Socrates.}
  \label{fig:aphorism11}
  \vspace{-.1in}
\end{figure}

The power of collaborative reasoning has deep philosophical roots. While Socrates transformed thinking through systematic questioning, even he recognized the necessity of dialogue partners. As Aristotle later formalized, dialectic requires thesis and antithesis to approach synthesis \cite{aristotle350BCE}. This principle manifests itself in $\MACI$ through several dimensions:

First, in dialectical reasoning, each LLM serves both as a questioner and as a respondent. Like Socrates engaging with interlocutors, one LLM challenges assumptions while another defends or refines them. This dynamic creates what philosopher H. Arendt called the ``in-between space'' where meaning emerges from the interaction itself \cite{Arendt1958Human}. Each exchange builds on previous analysis, refining hypotheses through mutual examination.

Second, meaningful collaboration requires a \textit{baseline} level of competency. Socrates could not obtain insight through dialogue with those lacking reasoning skills, and $\MACI$ cannot produce valuable results when the participating models are limited. Two weak reasoners do not yield strength; their limitations may compound, undermining productive discourse. With progress in LLMs, we have crossed this competency threshold, enabling collaborative reasoning that can identify and correct individual model limitations.

Third, the depth of dialogue varies with the capabilities of the participants. Advanced LLMs (e.g., Claude 3.7, GPT-4o, and DeepSeek R1) can explore complex ideas with \textit{depth}, leveraging sophisticated knowledge and reasoning to refine and challenge each other's perspectives. In contrast, simpler models engage only in shallow exchanges, similar to novices struggling with complex topics. The diversity in training data and architectures among advanced models allows them to effectively challenge each other's blind spots and biases.

\begin{figure}[t!]
  \centering
  \includegraphics[width=0.98\linewidth, height=0.50\linewidth]{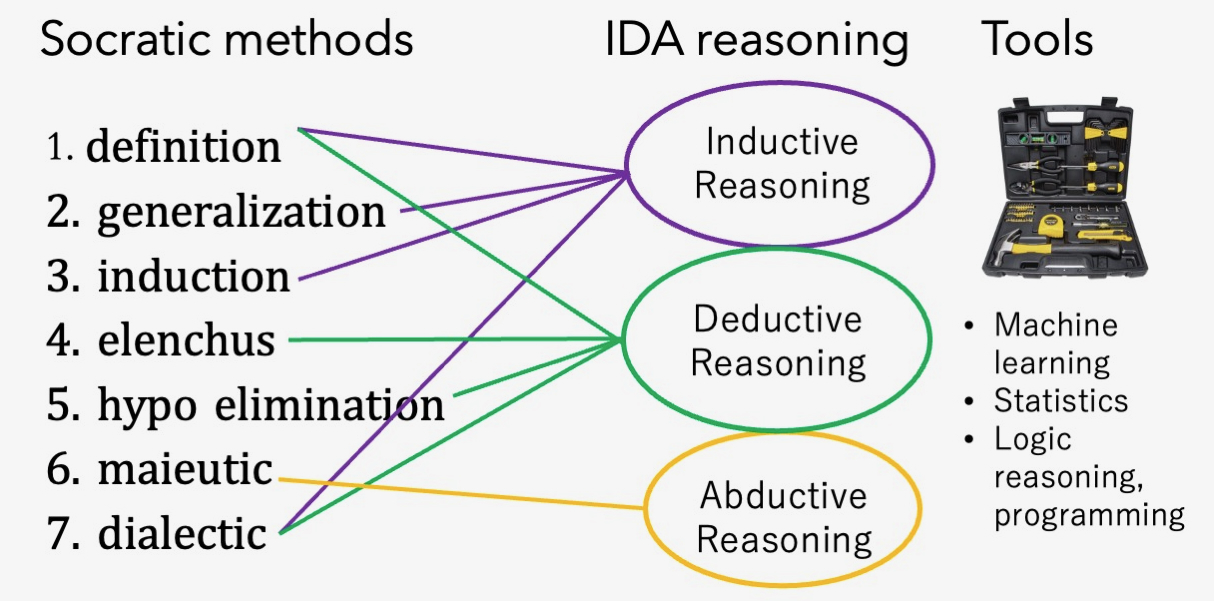}
  \caption{$\CRIT$: Critical Inquisitive Template. Mapping from individual Socratic methods to reasoning methods.}
  \label{fig:aphorism11-crit}
  \vspace{-.1in}
\end{figure}

$\MACI$ creates a space for \emph{AI philosophers} to engage in structured dialogue through frameworks such as $\SocraSynth$ and $\DIKE$\textendash$\ERIS$, with $\EVINCE$ monitoring the flow of information and $\CRIT$ \cite{SocraticIEEECCWC2023} assessing the quality of reasoning. The quality of these dialogues directly reflects the capabilities and engagement of the participants. This understanding informs both the selection of models for analytical tasks and sets realistic expectations for their collaborative performance.

This approach suggests that the path to artificial general intelligence may not lie solely in scaling individual models, but rather in creating systems where multiple specialized models engage in structured collaboration, mirroring how human intelligence often emerges through social and intellectual exchange. As philosopher Karl Jaspers noted, ``Truth begins between people'' \cite{jaspers1969philosophy}, a principle that appears equally applicable to artificial intelligences seeking deeper understanding.

\paragraph{\textbf{Note:}} This aphorism complements Aph. \#9 (the essence of precise questioning), showing how the quality of questions leads to better reasoning through collaborative debate. Together, they highlight the dialectical nature of $\MACI$ systems, where knowledge emerges not from monologue, but from structured dialogue.

\section*{Aph. \#11: Hallucinations fade under scrutiny}
\noindent
\textit{``Hallucinations rarely repeat.''} \\
\addcontentsline{toc}{section}{Aph. \#11: Hallucinations fade under scrutiny}

Have you ever wondered why recurring nightmares, even when they share the same theme, never quite unfold in exactly the same way? This aspect of human dreaming finds a parallel in the way LLMs process information. Just as nightmares rarely recur in exactly the same way, even when themes repeat, LLM hallucinations exhibit a similar non-deterministic quality. This characteristic distinguishes hallucinations from systematic errors and offers both challenges and opportunities for detection and mitigation \cite{ji2023survey}.

The phenomenon stems from three key mechanisms:

First, hallucinations arise when probabilistic token prediction leads to unpredictable sequences due to ambiguous or insufficient input. Since token selection is based on a probability distribution, even similar inputs result in varied hallucinations. This contrasts with systematic errors, which consistently emerge from gaps or biases in the training data.

Second, in $\MACI$ frameworks such as $\SocraSynth$ \cite{SocraSynthChangCSCI2023, EVINCEChang2024}, the non-repetitive nature of hallucinations becomes a strength. When one LLM produces a hallucination, others can challenge it with counterarguments. The evolving context of the debate progressively constrains the ``hallucination space''; as the context buffer is filled with specific claims and challenges, it becomes increasingly difficult for the original LLM to reproduce the same hallucination. Instead, it must ground its response in factual knowledge or acknowledge uncertainty. This iterative interaction creates a self-correcting dynamic in which hallucinations naturally diminish through increasingly precise discourse.

Third, while hallucinations are sporadic, true knowledge gaps are consistent and can be systematically addressed. By integrating Retrieval-Augmented Generation (RAG), we can distinguish random hallucinations from persistent knowledge deficits, allowing for targeted improvements to the model's knowledge base. This combined approach addresses both stochastic errors and systematic knowledge limitations.

\paragraph{\textbf{Example:}} In a multi-agent dialogue about historical events, LLM$_A$ might initially hallucinate that ``The Treaty of Versailles was signed in 1921'' LLM$_B$ challenges this with ``The Treaty of Versailles was actually signed in 1919, following the end of World War I.'' When asked again about the treaty, LLM$_A$ is unlikely to repeat the original error since the correct date now exists in the conversational context. Even if LLM$_A$ produces a new hallucination, it will generally be different from the original, allowing for continued refinement through iterative challenges \cite{manakul2023selfcheckgpt}.

This understanding has implications for LLM system design: while safeguards against hallucinations remain necessary, we can also leverage their non-repetitive nature in multi-LLM architectures to build self-correcting systems. The key insight is that multi-agent debate does not require explicit hallucination detection as fabrications naturally dissipate as debate evolves the context.

\paragraph{\textbf{Note:}} This aphorism complements both Aph. \#10 (Debate strengthens reasoning quality) and Aph. \#2 (External mirrors enable validation). The fading of hallucinations under scrutiny demonstrates why external validation through debate is so effective: it not only identifies errors but also progressively \emph{constrains the space} of possible errors in future exchanges.


\section*{Aph. \#12: Polydisciplinary synthesis \\ forges new frontiers}
\noindent \textit{``LLMs are not taught about domain boundaries, as they were trained only to predict the next words. This polydisciplinary approach to information representation allows LLMs to synthesize knowledge that might be beyond narrowly focused, domain-specific human understanding.''} \\
\addcontentsline{toc}{section}{Aph. \#12: Polydisciplinary synthesis forges new frontiers}

The term polydisciplinary was introduced by Microsoft's Chief Scientific Officer, Eric Horvitz, at Stanford's HAI center in 2023. He noted that the GPT-4 training process—predicting the next tokens through maximum likelihood estimation—applies the same statistical approach whether processing physics equations or poetry: the model never teaches disciplinary boundaries.

Although humans organize knowledge into categories such as physics, poetry, biology, and philosophy, LLMs move fluidly across these divisions, unaware of traditional boundaries. This polydisciplinary capacity opens new possibilities \cite{DiscoveringUnknownChang-July2023} (Chapter 14):
\begin{itemize}[leftmargin=1.0em, topsep=-.05em, parsep=-.05em] 
\item When an LLM perceives a pattern linking Shakespearean verse and quantum mechanics, it does not question whether such a connection is conceptually permissible. 
\item Complex problems that traditionally require interdisciplinary expertise become accessible to LLMs, which operate freely beyond academic silos. 
\item Novel hypotheses may emerge precisely because LLMs do not filter out connections conventionally deemed implausible or inappropriate by domain norms. 
\end{itemize}

This boundary-free thinking enables remarkable creative connections. In one striking example, when asked to consider the relationships between music theory and protein folding, an LLM suggested that protein folding patterns could be analyzed as ``melodies'' with characteristic motifs, leading to potential new visualization techniques for molecular biologists. Similarly, when exploring connections between architectural principles and immune system function, an LLM proposed viewing the immune system as employing a ``defensive architecture'' with layered systems, a framework that offers novel perspectives on immunological resilience. These cross-domain insights emerge not from specialized expertise in either field, but from the absence of conceptual barriers between them.

$\MACI$ leverages this polydisciplinary trait to systematically explore questions beyond human foresight (see Aphorism \#9 on question formulation). Frameworks like $\SocraSynth$ \cite{SocraSynthChangCSCI2023}, $\EVINCE$ \cite{EVINCEChang2024} and the $\DIKE$ -$\ERIS$ dual model \cite{chang2024NeurIPS} enable $\MACI$ to uncover hidden pathways and perspectives that could otherwise remain inaccessible. By synthesizing these insights (as discussed in Aphorism \#11 on critical thinking), we can bridge the gaps between disciplines and generate innovative connections.

The essence of $\MACI$ lies in the navigation of interdisciplinary intersections where true insights often emerge. These spaces, naturally traversed by LLMs, are rich in potential but fraught with ambiguity, the realm of ``unknown unknowns.'' Here, humans may struggle to frame meaningful questions or discern valuable insights from irrelevant noise. In such scenarios, humans assume the role of moderators, guiding LLM exploration and critically evaluating its findings. This collaboration allows $\MACI$ to effectively explore and illuminate uncharted intellectual territories, enhancing our collective understanding. Chapter 14 provides an example that traverses various (unexpected) knowledge domains, starting from a seeded biblical story.

This capacity for polydisciplinary synthesis represents perhaps the most promising path toward artificial general intelligence (AGI) and beyond. True innovation requires the ability to forge new frontiers by perceiving novel connections across traditional boundaries, to see patterns invisible to specialists entrenched in disciplinary silos. Human innovation often requires rare interdisciplinary genius, and LLMs inherently operate beyond these constraints, potentially discovering insights and opening intellectual frontiers that might otherwise remain unexplored for decades. In this sense, the polydisciplinary nature of LLMs does not just enhance existing knowledge; it creates the conditions for genuine intellectual exploration into uncharted territories.


\section*{Aph. \#13: Human heuristics are limited and end-to-end learning shows supremacy}
\noindent
\textit{``When machines are freed from handcrafted features and human biases, they often surpass what even expert-designed systems can achieve.''}
\addcontentsline{toc}{section}{Aph. \#13: Human heuristics are limited and end-to-end learning shows supremacy} \smallskip

Across several landmark cases in AI, a recurring pattern has emerged: models that learn end-to-end, from raw data to output labels, outperform systems that rely on human-engineered features or heuristics. These cases highlight the limitations of human intuition in designing components for complex tasks and the surprising efficacy of self-learned representations when given enough data, compute, and architectural flexibility.

A clear example is the evolution from AlphaGo to AlphaGo Zero. The original AlphaGo, which defeated world champion Lee Sedol in 2016, combined deep neural networks with hand-crafted features and expert data curated from human Go games \cite{silver2016mastering}. However, its successor, AlphaGo Zero, learned purely through self-play, using reinforcement learning without any human input or prior game records. It surpassed AlphaGo's performance within days, illustrating that human priors may constrain rather than accelerate learning in sufficiently expressive models \cite{silver2017masteringzero}.

A similar leap occurred in molecular biology with AlphaFold. In a talk by Demis Hassabis, CEO of DeepMind at Stanford in 2023, he revealed that the original AlphaFold (v1) incorporated hand-crafted heuristics, such as distance restrictions derived from coevolutionary features, to predict protein folding \cite{senior2020alphafold}. In contrast, AlphaFold 2 used a fully end-to-end differentiable architecture, mapping directly from amino acid sequences to 3D structures using attention-based mechanisms and learned geometric constraints. The result was a dramatic improvement in accuracy, outperforming all previous methods by a wide margin at CASP14 \cite{jumper2021alphafold}.

The development of self-driving architectures offers another revealing case. Earlier autonomous driving stacks were built using modular pipelines: perception, localization, prediction, planning, and control, each engineered by separate teams with domain-specific heuristics. Although modular systems allow interpretability and debugging, they suffer from cascading errors and inflexibility in edge cases \cite{bojarski2016end}. In contrast, end-to-end driving models learn a direct mapping from camera input to steering commands or trajectory plans. Although initially controversial, systems such as Wayve's perception-action networks \cite{hawke2019end} and Tesla's vision-based transformer stack show that direct, learned policies can outperform modular approaches on certain driving benchmarks and allow faster generalization in unseen scenarios.

Another iconic example comes from computer vision. The SIFT algorithm (Scale-Invariant Feature Transform) was once the dominant hand-crafted feature extractor, widely used for object recognition and matching \cite{lowe2004distinctive}. It was built on elegant geometric heuristics and was robust to scaling and rotation. However, when AlexNet was introduced in 2012, its deep convolutional architecture, trained end-to-end from raw pixels to class labels, dramatically outperformed SIFT-based pipelines on the ImageNet benchmark \cite{krizhevsky2012imagenet}. What sealed the shift was not just accuracy, but adaptability: when SIFT could not explain or correct its failures, neural networks learned to adapt through backpropagation. The field pivoted almost overnight to data-centric learning \cite{WebScalePCM2028,ChangFirstDataCentric}, illustrating how brittle even well-designed heuristics can be in the face of complex real-world variation.

These examples illustrate a core insight: human-designed heuristics, while valuable for bootstrapping early progress, often encode assumptions or constraints that limit performance. In contrast, end-to-end learning systems discover task-specific representations optimized across the entire input-output mapping, leading to superior generalization and adaptability.

$\MACI$  embraces this lesson through its architectural philosophy. Although human-designed rules still play a role in safety, ethics, and interpretability, the most robust behaviors often emerge when agents are allowed to learn reasoning, critique, and reflection end-to-end, across debate, memory, and regulation frameworks such as $\CRIT$ \cite{SocraticIEEECCWC2023}. In evaluating reasoning quality, instead of developing our own subroutines, $\CRIT$ prompts multiple LLMs to evaluate each reason-to-claim entailment for reasoning quality and source creditability.

Nature does not give organisms a fixed rule book for every situation. Instead, evolution shapes flexible minds that learn by interacting with the world \cite{eagleman2011incognito, clark2013whatever}. The same is true for machine learning. End-to-end learning systems do not follow fragile, hand-coded rules. Intelligence emerges not from rigid design but from adaptation through feedback \cite{lake2017building}.

Nature does not give organisms a fixed rule book for every situation. Instead, evolution shapes minds that learn through environmental interaction \cite{eagleman2011incognito, clark2013whatever}. Similarly, machine learning systems often perform best when they can discover patterns directly from data. Although well-designed rules and architectures provide important structure, the adaptability that comes from learning through feedback allows systems to handle complexity and variation that pure rule-based approaches struggle to anticipate \cite{lake2017building}.


\section*{Aph. \#14: AGI emerges through \\collaborative intelligence}
\noindent \textit{``The path to AGI lies not in singular models, but in systems that reflect, regulate, and reason together.''}\\
\addcontentsline{toc}{section}{Aph. \#14: AGI emerges through collaborative intelligence}

Artificial general intelligence will not be the product of simply scaling up language models \cite{kaplan2020scalingllm}, nor refining a single monolithic architecture. Instead, it will emerge from \emph{systems of collaboration}—where multiple agents operate with different roles, epistemic perspectives, and regulatory mechanisms. These systems must do more than generate fluent outputs: they must \emph{interrogate their own reasoning}, revise their conclusions, and resolve tensions among divergent viewpoints.

Recent results from OpenAI's 4o1 study demonstrate that inference-time scaling—allowing models to generate additional tokens or engage in internal deliberation—incurs cost that grows linearly with token length, yet produces substantial gains in performance, particularly on complex tasks involving reasoning, validation, and planning. Table~\ref{tab:scaling_comparison} summarizes the tradeoff between training-time and inference-time scaling, suggesting that thoughtful allocation of inference compute may yield better returns than ever-larger training runs.

\begin{table}[ht!]
\centering
\footnotesize
\begin{tabular}{|p{2cm}|p{3.8cm}|p{3.8cm}|}
\hline
\textbf{Dimension} & \textbf{Training Scaling} & \textbf{Inference Scaling} \\
\toprule
\hline
\textbf{Cost scaling} & Exponential / Superlinear & Linear (per output token) \\
\hline
\textbf{Accuracy gain} & Diminishing returns (esp. at scale) & High marginal gains on hard tasks \\
\hline
\textbf{Flexibility} & Fixed post-training & Adaptive (per input/task) \\
\hline
\textbf{Model size} & Must grow for gains & Fixed model \\
\hline
\textbf{Alignment to reasoning} & Limited & High (via thinking tokens, CoT, debate) \\
\hline
\end{tabular}
\caption{Comparison of Training Scaling vs. Inference Scaling in LLM Performance}
\label{tab:scaling_comparison}
\end{table}

This vision lies at the heart of $\MACI$. In $\SocraSynth$, agents pursue truth through Socratic dialogue; in $\CRIT$, claims are assessed for argumentative rigor and causal coherence; in $\DIKE$-$\ERIS$, behavioral outputs are evaluated and constrained across moral and cultural lines. In long-lived planning and coordination tasks, $\SAGA$ serves as the persistent memory substrate, enabling LLMs to maintain system state, validate constraints, and ensure transaction-like guarantees throughout complex workflows.

Yet the coordination of agents across such workflows requires more than memory—it demands a planner. To meet this need, $\ALAS$ (Adaptive Layered Agent System) introduces a task decomposition engine and workflow orchestrator for multi-agent planning. Where $\SAGA$ ensures memory consistency and validation guarantees, $\ALAS$ defines agent roles, assigns sub-tasks, and manages execution sequences under real-world constraints. This is particularly vital for dynamic, multi-threaded environments such as urban ride sharing, logistics, or collaborative robotics—domains where traditional LLM planners fail due to static reasoning or lack of context retention.

$\ALAS$ addresses these challenges by distributing planning across modular agents, each with specialized functions and access to historical state via $\SAGA$. This architecture reflects the broader $\MACI$ principle: that general intelligence requires not just knowledge and language, but structure, specialization, and accountability.

\textbf{These frameworks embody a broader principle: no single model can fully model itself, just as no mind can achieve objectivity without encountering other minds.}

Where Aphorism \#2 emphasized the need for external validation, this final aphorism extends the insight into a constructive design philosophy. $\MACI$ offers not just a critique of current limitations, but a roadmap: \emph{modularity, reflectivity, and role specialization} as foundational pillars for general intelligence. It suggests that AGI may emerge not from scaling to trillions of parameters, but from engineering systems that are \emph{aware of their fallibility} and structured to adapt through principled dialogue.

This view also responds to early criticisms, such as those from Yann LeCun, that LLMs lack grounding, memory, or planning. We do not deny these limitations. Rather, we \emph{re-contextualize} them: LLMs are not complete intelligences, but foundational substrates. They are the \emph{language of thought} upon which cognitive regulation, memory persistence, and planning structures can be built—a form of artificial unconsciousness that, while lacking volition, provides the substrate for structured, goal-directed cognition to emerge.

Much like human intelligence emerges from communities, institutions, and layered feedback, \textbf{artificial general intelligence may arise not from isolation, but from interdependence}, not from singular dominance, but from distributed coherence.

\section*{Closing Remarks}

The fourteen aphorisms build upon one another in a layered structure that mirrors the architecture of $\MACI$ itself. Aphorisms 1--4 provide the philosophical grounding: they introduce the notion of truth as perspectival, underscore the necessity of external validation, and argue for a layered view of reasoning that blends simulation and structure. These foundational insights motivate the system-level design strategies articulated in Aphorisms 5--8, where context modulation, role-based behavior, and ethical checks define how LLMs are governed within collaborative systems.

Building on this regulated architecture, Aphorisms 9--12 describe how reasoning unfolds in practice: through strategic questioning, dialectical debate, error correction, and cross-disciplinary synthesis. This middle layer reflects $\MACI$'s operational core, where intelligence emerges from interaction rather than individual computation. Finally, Aphorisms 13 and 14 step back to frame this process as a path to AGI, which is not through larger models alone but through modular coordination and emergent coherence. This structure allows each layer to support the next, progressing from epistemic pluralism to intelligent orchestration.

\clearpage

\newpage
\chapter{A Brief History of AI: From Turing to Transformers}
\markboth{History of AI}{History of AI}
\markboth{}{}
\thispagestyle{empty}
\noindent{\textbf{Abstract}}
This chapter reinterprets the history of AI, focusing on the evolution of similarity measurement, from rule-based to context-aware models, and emphasizing its critical role in AI's core functions like learning and problem-solving. It explores the impact of detailed and evolving understandings of similarity in linguistics (text) and computer vision (image), projecting a future where AI merges advanced data analysis with abstract reasoning. The chapter will provide an in-depth analysis from the perspectives of linguistics, computer science, and cognitive psychology/neuroscience, illustrating how the progression of similarity concepts continues to fuel AI's advancement.

\section*{Introduction}

Artificial Intelligence (AI) has journeyed through a fascinating historical trajectory, marked by five pivotal epochs that each represent significant paradigm shifts triggered by major technological advancements. 
The epochs are as follows: \textit{Initiation}, setting the stage with foundational concepts and milestones of AI; \textit{Expert System Encoding Human Knowledge}, where AI systems were predominantly rule-based, encoding and applying human expertise; \textit{Heuristic-Based Modeling}, which highlights the era of developing and using heuristic methods for AI problem-solving; \textit{Learning Model from Data}, focusing on the transition to algorithms that learn and adapt from data, signifying the emergence of machine learning; and \textit{Context-Based Semantic Disambiguation}, highlighting AI's evolving proficiency in understanding and interpreting context, thereby improving semantic accuracy.

While numerous comprehensive sources, such as Wikipedia, provide detailed accounts of AI's evolution through various lenses: language, computation, philosophy, cognitive psychology, neuroscience, and 
application\textendash this chapter takes a different path. It zeroes in on a fundamental aspect: \textbf{similarity}.

When we consider the intelligence of machines, we often focus on attributes such as learning capacity, pattern recognition, predictive accuracy, robustness, adaptability, generalization, reasoning, problem-solving, and decision making abilities. These qualities collectively define the prowess of AI systems. Among these traits, the concept of similarity plays a pivotal role. For instance, in learning, an effective similarity measure is fundamental for recognizing patterns and generalizing knowledge. In terms of adaptability, the ability to detect similarities to previous experiences allows AI to adjust to new or evolving circumstances. Regarding robustness, employing similarity measures helps AI differentiate between normal and anomalous patterns, thereby increasing its resilience. Furthermore, in the realm of problem-solving, the capacity to identify similarities to previously encountered situations can enable AI to apply existing solutions to new problems, enhancing its efficacy in addressing challenges. This chapter explores the vital function of similarity across the broad spectrum of AI capabilities, underlining its significant contribution to the field's foundational operations.

In the realm of tangible objects, similarity measures are integral to various vision-related tasks, aiding in the recognition of patterns, shapes, and colors, which are essential for object recognition and image classification. In text analysis, these measures are crucial for identifying similarities in content, aiding in plagiarism detection, document retrieval, and language translation. In the auditory domain, similarity analysis of sound wave patterns or musical notes is key to genre classification and music recommendation systems. In medical imagery, these measures facilitate disease diagnosis by comparing patient images with known cases, enabling accurate medical condition identification and classification. Object feature comparison is foundational in robotics and surveillance for recognizing and interacting with physical entities. Similarly, facial and voice recognition systems rely on analyzing patterns to identify or verify identities, enhancing security and personal authentication. In e-commerce, similarity in product attributes or user preferences informs recommendation systems, enhancing user experience by suggesting related or complementary products.

In the abstract realm, similarity measures are crucial for discerning semantic relationships, aiding in knowledge representation, ontology mapping, and refining AI's interpretive faculties. Environmental studies leverage these assessments for climate modeling and ecological research. Sentiment analysis in social media or customer feedback utilizes similarity to extract insights into public sentiment or consumer behavior. These measures also underpin AI's problem-solving prowess in complex scenarios, informing strategy formulation. Behavioral analysis, whether in psychology or marketing, employs similarity comparisons to decode human actions and preferences. In the legal field, case similarity aids in judicial decision-making and legal scholarship. Language translation harnesses similarity in linguistic structures to break down language barriers. Furthermore, in creative writing, analyzing thematic or stylistic similarities assists in authorship identification, genre categorization, and literary exploration.

The advancement in similarity research, while appearing gradual, reflects not only human ingenuity but also the limitations imposed by computational resources and hardware capabilities. The quest to quantify similarity covers a broad spectrum of abstractions, from sensory inputs like visual, auditory, olfactory, and tactile data to complex abstract concepts such as ideas and semantics. Hardware improvements have enabled researchers to explore more advanced methods that encompass both concrete and abstract forms of similarity. This progression marks the field's growth in harmonizing detailed sensory data analysis with a deeper understanding of abstract concepts, utilizing computational advancements and diverse data interpretations.

Following sections will provide a deeper dive into key AI terminology and the development of similarity measures in two distinctive views:
{scientific disciplines} and {historical evolution}. The disciplinary view encompasses three key perspectives: linguistics, computer science, cognitive psychology, and neuroscience. The evolution view traces the historical journey of similarity measurement through distinct eras: rule-based, model-based, data-centric, and context-aware. 

Providing two views on similarity measurements\textendash across different scientific disciplines and through the historical evolution of AI methodologies\textendash offers a comprehensive understanding that caters to a broader audience with varied interests and backgrounds. Here are some reasons why this dual perspective is valuable:

\noindent \textit{Multidisciplinary Insight}: Examining similarity measurements from different scientific disciplines enriches the understanding by highlighting how various fields approach and apply the concept of similarity. This can foster interdisciplinary collaboration and innovation, as techniques from one field can inspire new approaches in another.

\noindent \textit{Historical Context}: Exploring how similarity measurement has evolved within AI provides historical context, showcasing how methodologies have progressed from rule-based to more advanced context-aware systems. This perspective helps readers appreciate the advancements in AI and understand why certain methods were developed or abandoned.

\section{Definitions}
\label{sec:definitions-AI}

We define and scope key terms and concepts
to prepare for subsequent discussion.

\subsection{Rudimentary Terms}

\noindent \textit{Data}: The raw information used to train AI models. Data quality significantly impacts model performance.

\noindent \textit{Algorithm}: A set of instructions that a computer follows to perform a specific task. AI algorithms are often complex and involve statistical methods.

\noindent \textit{Model}: A representation of the learned knowledge from data that allows the AI system to make predictions or decisions.

\subsection{General Terms}

\noindent \textit{Artificial Intelligence (AI)}: The broader concept of machines being able to carry out tasks in a way that we would consider smart.

\noindent \textit{Explainable AI}: AI systems that offer transparency and an understanding of their operations and decision-making processes.

\noindent \textit{General AI}:
General AI, also known as Artificial General Intelligence (AGI), refers to a type of AI that has the ability to understand, learn, and apply knowledge in a wide range of tasks, much like a human being. It's an AI system with generalized human cognitive abilities, meaning that when presented with an unfamiliar task, it can find a solution without human intervention. AGI would be able to reason, solve problems, make judgments, plan, learn, and communicate in natural language, among other capabilities. However, as of now, AGI remains a theoretical concept and has not been realized in practical applications.

\noindent \textit{Narrow AI}:
Narrow AI, in contrast, is the type of AI that we encounter in our daily lives and is currently in use around the world. It is designed to perform a narrow task (e.g., facial recognition, internet searches, driving a car) and is trained for a specific dataset or a set of tasks. Narrow AI operates under a limited pre-defined range or context, often focusing on executing a single task extremely well or carrying out a limited range of tasks in a specific domain. It lacks the general cognitive abilities of AGI and cannot apply its knowledge beyond its specific field or task.

\noindent \textit{Machine Learning (ML)}: A subset of AI that includes statistical techniques that enable machines to improve at tasks with experience.

\noindent \textit{Deep Learning}: A subset of machine learning that uses neural networks with three or more layers. These neural networks attempt to simulate the behavior of the human brain\textendash albeit far from matching its ability\textendash allowing it to ``learn'' from large amounts of data.

\noindent \textit{Neural Networks} Computational models that are somewhat inspired by the structure of the human brain, enabling computers to recognize patterns and solve common problems in AI, such as classification, prediction, and decision making.

\noindent \textit{Supervised Learning}: A type of machine learning where the model is provided with labeled training data and the desired output. The goal is to learn a mapping from inputs to outputs.

\noindent \textit{Unsupervised Learning}: A type of machine learning where the model is not provided with labeled data and must find structure in its input on its own.

\noindent \textit{Reinforcement Learning}: An area of machine learning where an agent learns to behave in an environment by performing actions and seeing the results, focusing on long-term rewards. An example is
an AI agent learning to play a game through trial and error, receiving rewards for winning.

\noindent \textit{Natural Language Processing (NLP)}: A field of AI that gives machines the ability to read, understand, and derive meaning from human languages.

\noindent \textit{Computer Vision}: A field of AI that trains computers to interpret and understand the visual world, extracting information from images and videos.

\noindent \textit{Robotics}: The branch of technology that deals with the design, construction, operation, and application of robots, often incorporating AI systems to enhance autonomy and adaptability.

\noindent \textit{Large Language Model} (LLM):. LLMs are advanced artificial intelligence systems trained on extensive datasets, initially text-centric and now increasingly incorporating multimodal data. They are designed to comprehend, generate, and interact with human language, imagery, and video with a level of sophistication that closely mirrors human cognitive processes. 

\subsection{Performance Terms}

\noindent \textit{Algorithmic Bias}: Algorithmic bias refers to the potential for algorithms to reflect, perpetuate, or amplify biases present in the training data or as a result of the design of the algorithms themselves. This can lead to skewed or unfair outcomes, particularly in decision-making processes.

\noindent \textit{Hallucination}: In the context of AI, hallucination refers to the phenomenon where a model generates or outputs information that is ungrounded, misleading, or not supported by the input data. This is commonly seen in language models where the generated text may be plausible but not factually accurate or relevant to the context.

\noindent \textit{Generalization}: Generalization is the ability of an AI model to perform well on new, unseen data that was not part of the training set. It indicates the model's capacity to apply learned knowledge to different situations, a key indicator of its robustness and utility.

\noindent \textit{Overfitting}: Overfitting occurs when an AI model learns the details and noise in the training data to the extent that it negatively impacts the model's performance on new data. This usually happens when the model is too complex, capturing patterns that do not generalize to unseen data.
\section{Perspectives on Similarity}
\label{sec:perspectives-similarity}

This section presents the foundational theories of similarity measurement from three distinct domains: {\em linguistics}, {\em computer science}, and {\em cognitive psychology \& neuroscience}. The upcoming historical section will clarify how these foundational theories have influenced and been incorporated into specific technological advancements and methodologies across various eras. Cross-references will be provided to ensure coherence and to emphasize the interconnection of these perspectives.

\subsection{Linguistic Perspective}

The study of similarity within linguistics has been profoundly influenced by Zellig Harris's pioneering work. His 1954 study introduced the idea that the distributional properties of words and their contextual usage could unlock the secrets of language comprehension, highlighting the indispensable role of context \cite{ZHarris1954}. This principle, that words found in similar contexts tend to share meanings, laid the foundation for distributional semantics and resonates with John R. Firth's insight that ``A word is known by the company it keeps.'' This linguistic perspective sets the stage for further exploration of how context and distributional properties have been instrumental in shaping our understanding of semantic similarity, paving the way for subsequent advancements in the field.

The evolution of linguistic theories continued into the latter part of the 20$^{th}$ century with the rise of cognitive linguistics, which examines the interplay between linguistic structures and human cognitive processes. This approach underscored how language reflects our perception and conceptualization of the world, introducing a multi-layered perspective on semantic abstraction.

A significant milestone in bridging linguistic theory with practical applications was the development of WordNet in the 1980s by a team at Princeton University \cite{miller1995wordnet}. This lexical database, which organizes English words into sets of cognitive synonyms or {\em synsets}, has profoundly influenced areas such as word sense disambiguation, information retrieval, and beyond, highlighting the importance of structured semantic relationships in understanding language.

Moreover, the influence of linguistic insights extended into the domain of computer vision with the creation of ImageNet by Fei-Fei Li \cite{deng2009imagenet}, which drew upon the principles underlying WordNet to categorize visual content. This convergence of linguistics and computer science has been further propelled by advancements in computational methods, with techniques like Latent Semantic Analysis (LSA) \cite{dumais2004latent}, Latent Dirichlet Allocation (LDA) \cite{blei2003latent}, and innovative word embeddings such as Word2Vec \cite{mikolov2013efficient} and GloVe \cite{pennington2014glove}. These methodologies have enabled the conceptualization of word meanings in high-dimensional spaces, illuminating the intricate web of semantic relationships through patterns of co-occurrence and contextual analysis.

The introduction of the transformer model~\cite{vaswani2017attention} and the subsequent unveiling of BERT \cite{devlin2019bert}, which employs self-supervised learning to predict masked words within a context, along with the release of GPT, designed to predict the next word based on context, heralded a new epoch in our endeavor to unravel context-dependent semantics. This development fulfills the vision proposed by Zellig Harris in his groundbreaking 1954 work, now actualized in contemporary computational models.

\subsection{Computer Science Perspective}

In computer science, the concept of similarity has evolved from simple rule-based models to complex vector-space and probabilistic models, reflecting the field's progression in addressing various computational challenges. 

\subsubsection{A. Rule-Based}

A rule-based AI model, also known as an expert system, employs a collection of predefined if-then statements to execute decisions or solve problems. These conditional statements are crafted from the expertise of specialists in a particular field. The system applies these rules to the input data to formulate conclusions.

The ``if'' segment of a statement evaluates the data for specific conditions or patterns. When these conditions are satisfied, the ``then'' segment is activated, performing a designated action or drawing a conclusion. Importantly, these systems do not adapt or learn from data in the manner that machine learning models do. Rather, they rely on a set of explicit rules, which are the codified versions of expert knowledge within a specific domain. This knowledge is methodically organized and stored in a knowledge base, enabling the system to reference and apply it efficiently during its operations.

In Chapter~\ref{sec:rule-era}, we will explore the technical details and applications of rule-based systems, emphasizing their pivotal role during the rule-based era of AI's evolution.

\subsubsection{B. Vector-Space}

The vector-space model marked a significant shift, representing objects and features as vectors in a high-dimensional space. This approach facilitated the development of various distance functions to assess similarity for different applications. Notably, a comprehensive survey by \cite{Cha2007ComprehensiveSO} categorized 45 distance functions into families like inner product, $L_1$, Minkowski, and Intersection, each with its representative functions highlighting the versatility in vector-space analysis.

\subsubsection*{~~B.1. Inner product, dot product and cosine}

The inner product and dot product are the same in the context of Euclidean space and are defined for vectors \( \mathbf{a} \) and \( \mathbf{b} \) as:
\[
\mathbf{a} \cdot \mathbf{b} = a_1b_1 + a_2b_2 + \ldots + a_nb_n.
\]

This operation results in a scalar value and indicates the vectors' magnitude and directionality.
 
Cosine similarity is a measure that calculates the cosine of the angle between two vectors. It is defined as the dot product of the vectors normalized by the product of their magnitudes:
\[
\text{cosine similarity}(\mathbf{a}, \mathbf{b}) = \frac{\mathbf{a} \cdot \mathbf{b}}{\|\mathbf{a}\| \|\mathbf{b}\|},
\]
where \( \|\mathbf{a}\| \) and \( \|\mathbf{b}\| \) represent the Euclidean norms of the vectors \( \mathbf{a} \) and \( \mathbf{b} \), respectively.

The cosine similarity is especially useful in contexts where the magnitude of the vectors is not of primary concern, making it ideal for applications in high-dimensional spaces like text analysis and information retrieval.

\subsubsection*{~~B.2. Weighted Minkowski}

The weighted Minkowski distance function allows assigning varying importance to different dimensions, accommodating the significance of specific features in contexts like machine learning and data mining.
The weighted Minkowski distance between two points $X = (x_1, x_2, \ldots, x_n)$ and $Y = (y_1, y_2, \ldots, y_n)$ with a set of weights $W = (w_1, w_2, \ldots, w_n)$ is defined as:

\begin{figure}[ht!]
  \vspace{-.1in}
  \centering
  \includegraphics[width=0.95\linewidth]{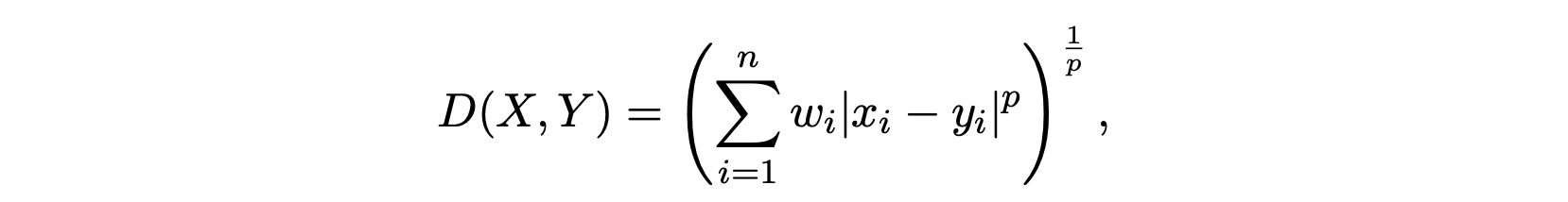}
  \vspace{-.1in}
\end{figure}

\subsubsection*{~~B.3. Set similarity}

Moreover, the Jaccard similarity \cite{jaccard1912distribution} provides a robust method for comparing sets, especially beneficial in scenarios where feature presence or absence is more critical than their magnitude, as seen in plagiarism or copyright detection.

\begin{figure}[ht!]
  \vspace{-.1in}
  \centering
  \includegraphics[width=0.98\linewidth]{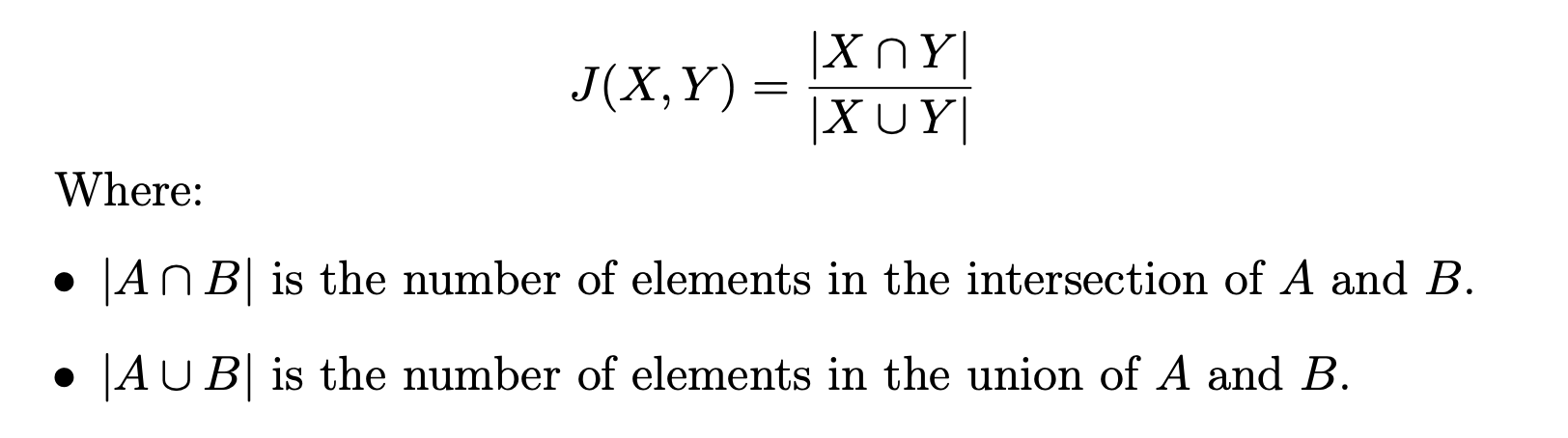}
  \vspace{-.3in}
\end{figure}

\subsubsection{C. Probabilistic-Based}

The advancement into probabilistic-based models introduced a spectrum of statistical and probabilistic distance functions, offering refined tools for quantifying similarity or dissimilarity based on underlying probabilistic principles. These functions, including Pearson Correlation Coefficient, Mahalanobis Distance, Kullback-Leibler Divergence, and others, cater to diverse analytical needs, enriching the computational toolkit available for similarity assessment in various domains.

This section underscores the computer science perspective on similarity, detailing its journey from rule-based logic to advanced probabilistic models, reflecting the field's dynamic evolution and its pivotal role in shaping contemporary approaches to measuring similarity.

\subsection{Cognitive Psychology Perspective}

Cognitive psychology and neuroscience provide deep insights into how similarity is perceived and processed at a neural level, significantly influencing the development of AI technologies. Anne Treisman's Feature Integration Theory (FIT) \cite{TREISMAN198097} has been instrumental in understanding how the brain synthesizes various sensory features into cohesive percepts, a concept that has parallels in how artificial neural networks, particularly Convolutional Neural Networks (CNNs) \cite{lecun1989backpropagation,lecun1998gradient}, process visual information.

FIT draws heavily from Gestalt psychology principles \cite{wertheimer1912experimentelle,koffka1935principles}, which propose that perception organizes individual components into a meaningful whole. This aligns with FIT's view that perception is an integrated experience shaped by the brain's organizational tendencies. The theory also intersects with selective attention, as seen in Donald Broadbent's Filter Model \cite{Broadbent1958PerceptionAC}. This model suggests attention acts as a filter, selecting relevant information for further processing. Broadbent's framework complements FIT by emphasizing attention's role in integrating features into a unified perception, highlighting the brain's selective processes.

\begin{figure}[ht!]
\begin{center}
    \centerline{\includegraphics[width=0.8\linewidth]{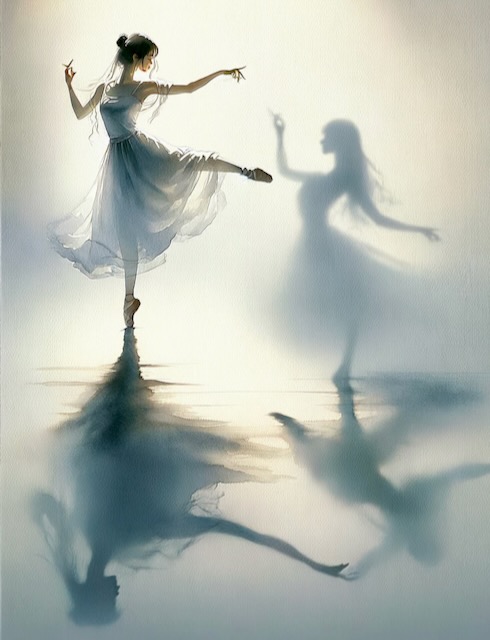}}
    \caption{``Which Pairs are Similar?'' (DALL-E)}
    \label{fig:similar-dancers}
\end{center}
\vspace{-.2in}
\end{figure}

In 2001, while conducting a study on perceptual similarity with my PhD student Beitao Li, we uncovered that images could demonstrate similarity in various dimensions. The weighted-Minkowski function,
although could learn feature weights, its application was universal once the weights were set, representing a statistical average. Our experiments with transformed images\textendash through translation, cropping, rotation, down-sampling, and affine scaling\textendash revealed that while these images were perceptually similar to their originals, their similarities were in distinct aspects. This observation led to the development of our ``Dynamic Partial Function'' (DPF) in 2002 \cite{DPF2002,DPF2003}. The DPF signature for each image pair could be unique. Essentially, if a pair of images (or objects) demonstrates a sufficient number of similar features, they are likely deemed similar, regardless of the specific features. For instance, an image is considered similar to its rotated version due to their color histograms' similarity. Similarly, an image and its cropped version are considered alike based on their texture features. If two images exhibit a sufficient degree of similarity in various respects\textendash typically 60\%\textendash they are generally regarded as similar.

While survey the literature, we came across ``Respects for Similarity'' by Medin, Goldstone, and Gentner \cite{medin1993respects}, which portrays similarity as a dynamic process of formulating a function and identifying relevant aspects, a process that is realized consciously. To clarify this concept, let's refer to an example from \cite{DPF2003}:

Consider the task of identifying two places similar to England. Scotland and New England might emerge as viable candidates. Yet, the criteria making England similar to Scotland are distinct from those linking England to New England. Using the attributes that align England with Scotland to assess the similarity between England and New England might not yield a parallel conclusion, and the reverse is also true. This scenario underscores the idea that objects can be similar to a reference object in varied respects. A fixed similarity function, bound to a specific set of criteria, fails to capture the similarities across different contexts. Medin, Goldstone, and Gentner \cite{medin1993respects} examine the operational dynamics of similarity in human cognition, noting that the selection of relevant attributes is crucial, with similarity being as much a result as a driving force of conceptual coherence. Goldstone \cite{Goldstone1994} further elucidates that similarity involves identifying the appropriate criteria for comparison, which occurs \textit{only after the objects in question have been juxtaposed}, not beforehand. The criteria selected for this comparison are activated during the comparison process, with a tendency to favor those that enhance the coherence of the objects being compared.

Although the Dynamic Partial Function (DPF) introduces computational complexity, it has indirectly played a role in the success of AlexNet \cite{AlexNet2012} by influencing data augmentation strategies. By integrating transformed images into its training dataset, AlexNet benefits from a principle akin to DPF, thereby improving its accuracy and robustness in recognition tasks. The recent advancements in transformer algorithms \cite{vaswani2017attention}, which focus on dynamism and context-awareness, build on this foundation, a topic that will be explored in detail in the subsequent section.

\subsubsection{Neuroscience}

The neuroscience foundation of FIT and its relation to visual feature processing are echoed in the development of CNNs, which were inspired by the visual cortex's hierarchical structure and feature detection capabilities as explored by Hubel and Wiesel \cite{Hubel1962ReceptiveFA}. These networks utilize convolutional layers to automatically and adaptively learn spatial hierarchies of features from visual data, akin to the neural processing observed in the brain.

Techniques like Multivariate Pattern Analysis (MVPA) \cite{Norman2006BeyondMI} and Neural Decoding \cite{Haynes2006DecodingMR} further bridge the gap between neuroscience and AI, offering methods to analyze how information is represented across neural populations and how these representations can predict perceptual experiences or cognitive states. These methodologies have inspired and informed the design of advanced AI systems, particularly in how they encode, process, and differentiate complex patterns and similarities.

The cross-pollination between neuroscience and AI, exemplified by the influence of neural processing principles on CNN design, highlights the symbiotic relationship between these fields. Insights from studying the brain's processing mechanisms have catalyzed innovations in AI, leading to more effective and biologically inspired computational models. This interdisciplinary exchange not only propels forward our understanding of neural processes but also fosters the development of AI systems that more closely mimic human perceptual and cognitive capabilities.

\subsection{Section Remarks}

The exploration of similarity measurement spans across linguistics, computer science, and cognitive psychology and neuroscience, revealing its multidisciplinary nature. Each field offers a unique lens to view similarity, from the contextual information in language, computational algorithms in AI, to the neural processing in the human brain. They converge on the common ground of representing entities in high-dimensional spaces and employing distance metrics for quantification, highlighting the universal applicability of similarity. This convergence fosters a rich dialogue between disciplines, enhancing our understanding and ability to quantify and interpret similarity, driving forward innovation and providing new methodologies that influence a wide array of contexts in our quest to decode this fundamental concept.
\section{Eras of Similarity Measurement}
\label{sec:eras-similarity}

Traversing through the history of artificial intelligence and similarity measurement, one can delineate distinct eras, each marked by unique methodologies and technological advancements. Contrast to
last section which examines similarity measurements from 
different scientific disciplines,
this section chronicles these eras, starting from the \textit{rule-based} era, which laid the foundational stones, through the evolution into \textit{model-based}, \textit{data-centric}, and \textit{context-aware} methodologies, illustrating the dynamic trajectory of similarity measurement in AI.  As we reach the conclusion of this section, we explore the prospects of the forthcoming era, which promises to challenge and expand our understanding by venturing into the realm of discovering the \textit{unknown unknowns}.

\subsection{Rule-Based Era (1950s - )}
\label{sec:rule-era}

The rule-based era of the 1950s marked the inception of AI, characterized by the use of symbolic representations and logic to analyze similarity. This period saw the emergence of explicit symbolic representations and logic-based methods tailored for similarity assessment. Innovations by Allen Newell and Herbert A. Simon with tools like the Logic Theorist and General Problem Solver \cite{Newell1956LogicTA} pioneered logical rule-based problem solving, setting a pivotal foundation for AI's evolution.

In the following decades, systems such as DENDRAL \cite{Lindsay1979ApplicationsOD} utilized rule-based logic to deduce molecular structures from data, while MYCIN \cite{Shortliffe1975MYCINAE}, an expert system for diagnosing infections and recommending treatments, demonstrated the practical application of rule-based reasoning in the field of medical diagnostics.

Despite their effectiveness in well-defined scenarios, rule-based systems have limitations in more complex or changing environments. However, their clarity and systematic nature are invaluable in certain 
applied areas, for example:
\begin{enumerate}[leftmargin=1.25em, topsep=.2em, parsep=.2em, label=\arabic*.]
\item \textit{Customer Service}: Rule-based chatbots are prevalent in customer service, using predefined rules to respond to inquiries based on detected keywords or phrases in user input, providing immediate and consistent customer support.
\item \textit{Fraud Detection Systems}: The finance sector employs rule-based systems to identify potential fraudulent transactions by comparing against specific criteria, such as unusual transaction amounts or atypical locations.
\item \textit{Equipment Failure Diagnosis}: In industrial settings, rule-based systems analyze data to pinpoint causes of equipment failures, leveraging historical data and expert knowledge to predict and prevent future breakdowns.
\end{enumerate}

This era introduced key tools like PROLOG \cite{Clocksin1981ProgrammingIP} for logic programming and decision trees \cite{Quinlan1986InductionOD} for visualizing decision processes, exemplifying rule-based logic.

While rule-based systems initially approached similarity with a clear, logical framework, subsequent AI advancements have embraced more flexible methods like statistical models and machine learning, offering a broader, more adaptable approach to understanding similarity in various contexts.

Rule-based systems contrast with the ``black-box'' nature of current Convolutional Neural Networks (CNNs) and Large Language Models (LLMs) in terms of interpretability and decision-making processes. Rule-based systems are transparent in how decisions are made, as they follow a clear set of if-then rules or logic for inference, allowing users to understand and trace the reasoning behind each decision.

On the other hand, CNNs and LLMs, particularly those based on deep learning, often operate as black boxes, where the internal decision-making processes are not easily interpretable. In these systems, decisions result from complex, non-linear interactions of thousands to millions of parameters that have been adjusted through the learning process. While they are powerful and effective in handling a wide range of tasks, especially those involving large datasets and requiring pattern recognition beyond human capabilities, their inner workings are not as transparent or interpretable as rule-based systems.

\subsection{Model-Based Era (1970s - )}
\label{sec:model-based-era}

In this era, vector-space and probabilistic models
were designed to quantify similarity. 
\newline
\noindent {\textbf{\ref{sec:model-based-era}.1 Vector Space Models}}
\\
The vector-space era marked a shift in similarity measurement from rule-based to representation-based approaches. In this era, objects, documents, and features began to be conceptualized as vectors in a high-dimensional space, fostering a more intuitive and flexible method for assessing similarity.

\subsubsection{The Vector-Space Model and Information Retrieval}

At the core of this era was the vector-space model, which represents documents as vectors of term frequencies, enabling the computation of document similarity using cosine similarity between their respective vectors. This model enhanced the efficiency and effectiveness of information retrieval systems.

\subsubsection{Distance Functions and Feature Weighting}

A diverse array of distance functions emerged during this era to quantify the similarity between vectors. The Minkowski distance, for instance, generalized traditional metrics like the Euclidean and Manhattan distances, offering flexibility in adjusting the sensitivity to differences in vector components. Weighted distance measures also gained prominence, recognizing that not all features have equal importance in similarity assessment. The weighted Minkowski distance, in particular, allowed for differential weighting of dimensions based on their relevance to the specific application at hand.

\subsubsection{Beyond Textual Data}

The utility of the vector-space model extended well beyond textual data. In the realm of image processing, features (e.g., colors, textures, and shapes) extracted from images were represented as vectors, enabling the assessment of image similarity based on the distances between these vectors. This paradigm facilitated significant advancements in image retrieval, classification, and clustering.

\subsubsection{Dimensionality Reduction Techniques}

To address the challenges posed by high-dimensional data, techniques like Principal Component Analysis (PCA) \cite{Jolliffe1986PrincipalCA} and Latent Semantic Analysis (LSA) \cite{LSA1998} were developed. These methods reduced the dimensionality of data while preserving its essential structure, enhancing computational efficiency and mitigating the ``curse of dimensionality.'' Manifold learning, a non-linear dimension reduction approach, further expanded the toolbox for tackling high-dimensional data \cite{Tenenbaum2000AGG}. For a comprehensive overview of these techniques, refer to \cite{Maaten2008DimensionalityRA}.

The vector-space era laid the groundwork for advancements in machine learning and data mining, making similarity measures essential for clustering, classification, and recommendation systems. Data representation as vectors allowed for the exploration of relationships across varied data types through the nearest neighbor concept. In this context, the characteristics or labels of an unknown instance's $k$-nearest neighbors could be inferred and applied to the instance, with these neighbors determined by distance metrics.

However, vector representations often result in sparsity, potentially leading to resource inefficiency and decreased accuracy. These models, while capturing syntactic relationships, sometimes struggle with semantic depth, such as identifying synonyms or contextual meaning. The assumption of feature independence and the use of linear methods in dimensionality reduction can also lead to inaccuracies, particularly with non-linear data structures. The introduction of Support Vector Machines (SVMs) \cite{Cortes1995}, which utilize kernel methods, addressed some challenges related to non-linear data but increased computational complexity. SVMs were a significant focus in the field until the rise of deep learning architectures like AlexNet marked a shift towards the data-centric era.
\newline
\noindent {\textbf{\ref{sec:model-based-era}.2 Probabilistic Models}}

Probabilistic models offer more flexibility than vector-space models because they can incorporate uncertainty and variability directly into their mathematical frameworks, allowing for a more comprehensive and adaptive representation of data.

\subsubsection{Statistical Inference and Similarity}

Probabilistic models introduced the concept of statistical inference, where the likelihood of data or feature occurrences was used to estimate similarity. This allowed for effective handling of uncertainty and variability in data, making it particularly useful in noisy or incomplete datasets.

\subsubsection{Bayesian Approaches}

Bayesian methods emerged as a fundamental component of this era, providing a robust framework for integrating prior knowledge and empirical data. These methods enhance model adaptability by systematically updating beliefs in light of new evidence, allowing for similarity measures that are responsive to evolving data landscapes.

For further reading on Bayesian methods and their application in dynamic and adaptive modeling, consult the following literature \cite{BarberBayesian,BishopPattern,GelmanBayesian,Koller2009ProbabilisticGM}.

\subsubsection{Latent Semantic Models}

In addressing the challenges of high dimensionality and data sparsity inherent in vector-space models, dimensionality reduction techniques were employed. However, beyond merely tackling these issues, the development of a latent semantic layer offered profound implications for semantic analysis and indexing.

As highlighted in the perspective section (Chapter~\ref{sec:perspectives-similarity}), Latent Semantic Analysis (LSA) \cite{dumais2004latent} and Latent Dirichlet Allocation (LDA) \cite{blei2003latent} are critical models in the landscape of semantic modeling. LSA employs singular value decomposition to condense the dimensionality of term-document matrices, unveiling the latent semantic structures within textual data. This dimensional reduction elucidates intricate relationships beyond mere surface-level feature overlaps, enabling a deeper comprehension of textual similarities.

\begin{figure}[th!]
  \centering
  \begin{minipage}[b]{1.0\textwidth}
    \includegraphics[width=\textwidth]{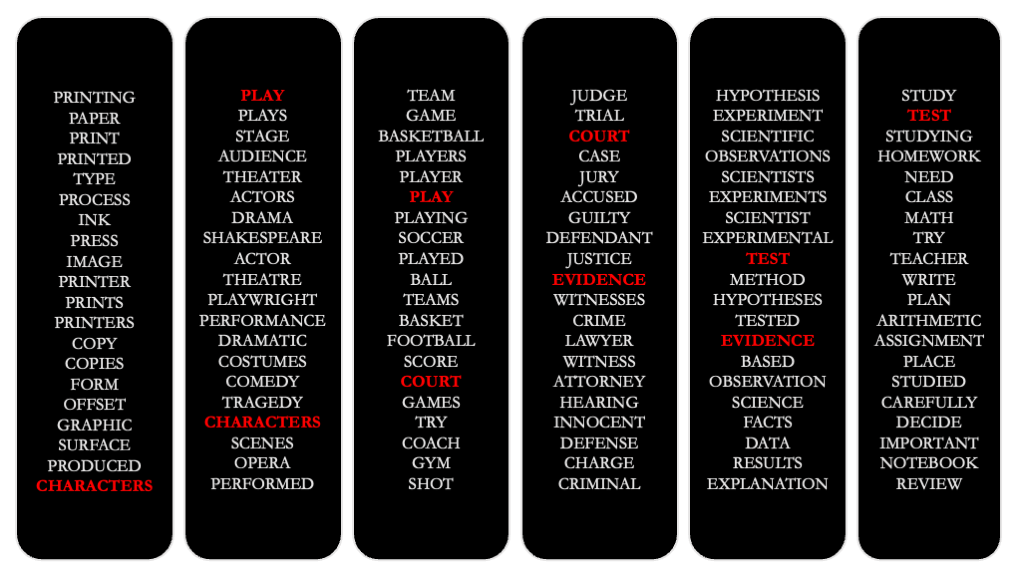}
  \end{minipage}
  \caption{Latent Clusters of LDA. The words in red belong to
  two semantic clusters, signifying the meaning of a word depends on its
  context.}
  \label{fig:LDAClusters}
\end{figure}

Similarly, LDA offers a probabilistic approach to topic modeling, where documents are considered mixtures of various topics, and topics are distributions over words. This bag-of-words model facilitates a deeper semantic connection between documents by associating them based on shared topics rather than just overlapping terms.

Figure~\ref{fig:LDAClusters} presents an example of how LDA, through its bag-of-words approach, clusters words into semantic groups. It's noteworthy that a word can belong to multiple semantic clusters. For instance, words like `characters', `play', `court', `evidence', and `test', each appears in two different semantic clusters in the illustration. This feature of LDA resonates with the insights from Zellig Harris's pioneering work and John R. Firth's adage that ``A word is known by the company it keeps.''

These latent semantic models transcend the limitations of direct feature comparison, enabling a more abstract representation of text. By doing so, they provide a robust foundation for semantic indexing and similarity assessment, offering insights that are essential for tasks such as information retrieval, document clustering, and topic discovery. The adoption of these models marked a significant advancement in understanding and measuring similarity in text, setting a new standard for semantic analysis in the field of natural language processing.

\subsubsection{Cluster Analysis and Similarity}

Probabilistic clustering algorithms, like Gaussian Mixture Models (GMMs), leveraged statistical methods to group data based on the likelihood of membership in different clusters. This probabilistic approach provided a more flexible and deeper understanding of groupings and similarities within data.

\subsubsection{Impact and Limitations}

While probabilistic models brought significant advancements, they also introduced challenges. The increased complexity often led to higher computational demands. Additionally, reliance on assumptions about data distributions or the need for prior knowledge could limit applicability in certain situations.

The probabilistic model expanded the toolkit for measuring similarity by introducing methods that could handle uncertainty and offer more adaptive and context-aware approaches. These advancements paved the way for even more sophisticated techniques in the subsequent data-centric era, where the focus shifted towards leveraging vast amounts of data to learn and adapt similarity measures dynamically.

\subsection{Data-Centric Era (2000s - )}

The data-centric era marked a transformative shift in artificial intelligence, pivoting towards harnessing the vast potential of big data, enabled by advances in computational hardware that facilitated parallel processing. This era is characterized by a move from heuristic-based methods to an empirical, data-driven approach in feature representation and model learning.

At the core of the data-centric paradigm is the emphasis on deriving model parameters from extensive datasets, distinguishing it from traditional model-centric strategies. Foundational algorithms such as CNNs \cite{krizhevsky2012imagenet} and Transformers \cite{vaswani2017attention}, while conceived through human ingenuity, saw their efficacy significantly enhanced when trained on large, diverse datasets. This training ensures broad coverage of potential variations across different objects or concepts, fortifying the models' ability to accurately recognize and classify new instances. The volume and diversity of the training data are crucial in refining the models' representations, leading to advancements in prediction accuracy and robustness.

\subsubsection{From MapReduce to Machine Learning at Scale}

The inception of the data-centric movement traces back to the seminal works in statistical learning theory. Vladimir Vapnik's insights into the importance of data for model generalization, particularly his development of Support Vector Machines (SVMs) \cite{Cortes1995}, and Tom Mitchell's pivotal book ``Machine Learning'' \cite{Mitchell1997}, which underscored the critical role of data in preventing overfitting, laid the theoretical foundation for this era.

MapReduce \cite{MapReduce2008}, a corner stone in data processing, enabled parallel computation to efficiently handle large datasets. Originally devised to enhance data processing tasks like Google's web indexing, MapReduce became the bedrock for the emergence of sophisticated data-centric methodologies in AI.

\subsubsection{Evolution of Machine Learning with Big Data}

The rise of parallel machine learning algorithms \cite{DMGoogle2008,WebScalePCM2028,chang2011foundations}, notably through Edward Y. Chang's work at Google, marked a significant milestone in this era. Chang and his team 
pioneered Web-scale image annotation in 2007 \cite{WebScalePCM2028}, and 
subsequently met Prof. Fei-Fei Li
after 2028 summer school of Computer Vision, and 
subsequently sponsored
the Stanford ImageNet \cite{deng2009imagenet} work via a substantial Google grant.

At the same time, his team developed groundbreaking parallel algorithms, including PSVM \cite{PSVM2007} (parallelizing SVMs by approximating
matrix factorization), PFP \cite{PFP2008} (parallelizing frequent itemset mining), PLDA \cite{PLDA2009} (parallelizing LDA algorithm), PSC \cite{PSC2011} (parallelizing spectral clustering), and SpeeDo \cite{zheng2015speedo} (parallelizing CNNs), driven by the recognition that big data could facilitate direct learning of features and representations, transcending the limitations of human-crafted heuristics.

\subsubsection{Impact on Similarity Measurement}

The data-centric era revolutionized the field of similarity measurement, ushering in a new paradigm where similarity metrics are derived from extensive datasets. This period underscored the critical role of data volume and quality in defining similarity metrics, highlighting the dynamic relationship between data-driven insights and computational methods.

In this era, deep learning architectures like CNNs and Transformers have been instrumental in advancing similarity metrics. These models stand out because they not only adjust feature weights but also autonomously learn features from the data. This capability to learn from data directly makes traditional human-engineered features increasingly redundant. After all, human heuristics may not capture every facet of an object or concept comprehensively, and human sensory perception is limited. For instance, while humans can detect the light spectrum from approximately 300 to 700 nanometers, cameras and X-ray machines can perceive a broader range of signals, demonstrating the advantage of machine-learned features in capturing and analyzing data beyond human limitations.

\subsection{Context-Aware Era (2010s - )}

The context-aware era in similarity measurement brings to fruition the profound insights of Zellig Harris's distributional semantics and John R. Firth's adage: ``a word is known by the company it keeps.'' This period marks a shift from static, context-independent assessments to dynamic, context-informed interpretations of similarity. It utilizes the latest advancements in machine learning and the growing availability of computational power to enhance our understanding of similarity in various contexts.

\subsubsection{Emergence and Evolution}

The integration of context-aware methodologies in similarity measurement evolved significantly in the 2010s, overcoming earlier constraints in computational power and data availability:

\begin{itemize}
\item \textit{Computation Capacity}: The development of AlexNet encouraged a data-centric focus within the AI community, prompting investments in parallel computing infrastructures.
\item \textit{Word Embeddings}: Techniques like Word2Vec enhanced semantic relationship encoding within data.
\item \textit{Attention Models and Transformers}: These models improved data analysis by concentrating on relevant data segments, refining context-aware assessments.
\item \textit{Large Language Models (LLMs)}: Models such as BERT and GPT, utilizing self-supervised learning on large text corpora, improved the understanding and generation of context-rich text.
\end{itemize}

\subsubsection{Foundational Pillars: Data and Computation}

Key pillars supported advancements in the context-aware era:

\begin{itemize}
\item \textit{Self-Supervised Learning:} Utilizing unlabeled data for learning enabled models to extract insights from the data, improving AI system efficiency and scalability.
\item \textit{Computational Advances:} The introduction of parallel algorithms and GPU acceleration enabled processing at unprecedented scales, facilitating the development of sophisticated models.
\end{itemize}

\subsubsection{Broader Implications}

This era not only refined similarity measurement techniques but also broadened how data is understood and knowledge is integrated:

\begin{itemize}
\item \textit{Reasoning and Explanation:} Models now aim to provide reasons for their similarity assessments, improving interpretability and building trust.
\item \textit{Multilinguality and Cultural Sensitivity:} Enhanced processing capabilities for varied linguistic and cultural data improve the global applicability of similarity measurements.
\item \textit{Multimodal Data Integration:} Context-aware models are adept at combining information from multiple modalities, offering a comprehensive view of similarity.
\item \textit{Polydisciplinary Knowledge Fusion:} Adopting a polydisciplinary approach allows for a broader knowledge base in making similarity assessments, fostering innovation across different fields.
\end{itemize}

The context-aware era signifies a shift toward more insightful, holistic, and interpretable AI, setting the stage for future developments where AI can offer contextually rich and multifaceted insights.

\subsection{Section Remarks}

What defines the next era in the evolution of AI? Historically, technological advancements have focused on addressing pressing unmet needs. Among various potential areas, enhancing the interpretability of decisions stands out as a crucial objective. Making the decision-making process of LLMs transparent and explainable could unlock significant improvements in numerous aspects, such as ethics, by enabling foundational enhancements rather than superficial tweaks based on guesswork and simple heuristics.

The fusion of rule-based system interpretability with the sophisticated capabilities of CNNs and LLMs poses a compelling challenge in AI. Active research is aimed at blending these approaches to leverage their distinct advantages:

\begin{enumerate}
\item \textit{Neuro-Symbolic AI}: 
Neuro-Symbolic AI (the third wave of AI \cite{garcez2020neurosymbolic}) aims to blend the data processing power of neural networks with the logical reasoning of symbolic AI. The goal is to create systems that not only excel in tasks like pattern recognition but can also reason and make decisions in a human-interpretable manner.

\item \textit{Incorporating Domain Knowledge}:
Embedding knowledge of experts within neural networks \cite{Lu_2022} can steer the learning process towards more reliable and interpretable outcomes.
In healthcare, for example, integrating medical guidelines into the training process of a neural network ensures that its predictions for patient treatment not only correlate with the data but also align with established medical practices, enhancing both the model's credibility and relevance.

\item \textit{Interactive Systems}:
A system such as SocraSynth \cite{SocraSynthChangCSCI2023} can 
combine the predictive power of deep neural networks with human expertise, allowing for iterative 
refinement and learning.
For instance, in SocraHealth \cite{SocraHealth2023}, it might suggest a set of possible diagnoses based on medical imaging, which a physician could then refine or correct. This feedback could be used to continuously improve the system, marrying machine efficiency with human expertise to enhance decision accuracy and interpretability.
\end{enumerate}

By advancing these strategies, the field of AI aims to develop models that not only excel in performance but are also transparent, understandable, and aligned with human reasoning, thus making AI more reliable and trustworthy across various applications.

\section{Concluding Remarks}
\label{sec:concChap2}

This chapter examines the history of AI through the lens of similarity, considering both disciplinary and chronological perspectives. Looking forward, we propose that the emergence of large language models (LLMs) marks a pivotal moment in the context-aware era of AI, setting the stage for the next frontier: the era of interpretability, understanding, and discovery. In this new era, the focus will shift towards empowering LLMs to not only comprehend but also to generate and innovate, synthesizing novel knowledge and insights.

This era of discovery is envisioned as a time when machines will extend their superiority beyond mastering games like Go and Chess to encompass a broader spectrum of tasks, outstripping human capabilities in various domains. 
The subsequent chapters of this book, beginning with Chapter~\ref{chap:SocraSynth}, explore the concept of harnessing the collective intelligence of multiple LLMs, embarking on a voyage to transcend the boundaries of the known and venture into the realm of discovery.

This chapter has explored the history of AI through the lenses of disciplinary and chronological perspectives, focusing on the concept of similarity. As we look to the future, the rise of large language models (LLMs) marks a significant milestone in the context-aware era, paving the way for a new era focused on interpretability, comprehension, and exploration. The upcoming phase in AI's evolution emphasizes enhancing LLMs with the ability to not just generate but also interpret and innovate, pushing the boundaries of knowledge creation and insight synthesis.

We anticipate an era where AI's capability extends beyond excelling in strategic games like Go and Chess to a wider array of endeavors, surpassing human performance across multiple fields. The following chapters, starting with Chapter~\ref{chap:SocraSynth}, research deeply into leveraging the collective intelligence of various LLMs. This journey aims to explore uncharted territories, advancing beyond established knowledge to uncover new frontiers in artificial intelligence.

\clearpage

\newpage
\chapter{Capabilities and Opportunities of LLMs}
\label{chap:insights}
\markboth{}{}
\thispagestyle{empty}
\noindent{\textbf{Abstract}}

This chapter depicts the architectural innovations and unique capabilities of Large Language Models (LLMs), with a special emphasis on the GPT-4 model. We dissect GPT-4's salient characteristics, such as its extensive cross-disciplinary and multimodal data representation, the intricate balance in its training methodologies, and the harmonious integration of human-guided insights with a robust data-driven learning framework. The chapter highlights the potential of LLMs to not only comprehend but also synthesize knowledge that transcends their training datasets, venturing into realms potentially uncharted by human understanding. We postulate that the true potential of LLMs hinges significantly on the articulation of queries posed to them. By elucidating these aspects, the chapter aims to shed light on how LLMs could rival or even surpass human intelligence in certain knowledge domains, setting a foundation for the subsequent exploration of LLMs' characteristics, insights, and their implications for future AI advancements.

\section*{Introduction}
\label{sec:intro2}

The evolution of large language models (LLMs) \cite{Bbubeck2023sparks,chatgpt,openai2023gpt4,thoppilan2022lamda,touvron2023llama} has significantly influenced natural language processing, enhancing capabilities in machine translation, sentiment analysis, and text summarization. Among these, GPT-4 \cite{openai2023gpt4} stands out for its exemplary performance across various benchmarks, including the MMLU \cite{SotaMMLUSite20231015}. Despite its achievements, GPT-4 grapples with challenges like hallucination, biases, and restricted reasoning.

This chapter studies the deep intricacies of GPT-4's architecture, emphasizing its knowledge representation, alignment with human values, and the synergy between human insights and data-driven learning. We discuss the model's limitations and introduce $\SocraSynth$, a supplementary reasoning layer designed to enhance knowledge discovery and analytical reasoning in GPT-4 and similar LLMs.

\subsection*{Capabilities and Implications}

We explore GPT-4's architecture, which, although initially kept in secrecy, has been progressively unveiled by the research community \cite{ouyang2022training,JackRae2023,rae2022scaling}. Our focus is on its knowledge representation and discovery, alignment with human values, and the integration of human expertise with data-centric methodologies.

Collaborations between Microsoft and OpenAI \cite{Bbubeck2023sparks} highlight GPT-4's interdisciplinary approach and its polymodal variant's benchmark achievements. We will further explore these aspects in Chapters \ref{sec-polydisciplinary} and \ref{sec-polymodality}. Discussions on human-value alignment will consider ChatGPT's RLHF methods \cite{Lex-Altman-2023} and the implications of pre-training censorship on foundational models, detailed in Chapters \ref{post-alignment} and \ref{pre-alignment}.

\subsection*{Limitations and Opportunities}

Addressing the biases, hallucinations, and constrained reasoning of LLMs requires innovative research initiatives. We introduce four key areas of focus:
\begin{itemize}[leftmargin=1.0em, topsep=-.05em, parsep=-.05em]
\item Enhancing Collaborative LLMs with Theoretical Foundations in Statistics and Information Theory.
\item Employing Open-Domain Reasoning with the Socratic Method to guide LLMs.
\item Model Behavioral Emotion to Safeguard AI Safety and Ethics.
\item Implementing Retrospective and Adaptive Evolving Learning frameworks to refine LLMs.
\end{itemize}

The root of bias in Large Language Models (LLMs) often lies in their training data. Built upon the transformer architecture, LLMs prioritize accurate token prediction, relying heavily on statistical patterns within their training corpus. This can inadvertently lead to bias towards prevalent opinions and expressions. To address this, Chapter \ref{chap:SocraSynth} introduces $\SocraSynth$, a framework designed to challenge these statistical tendencies by pitting two LLM agents against each other on a topic, each conditioned with opposing viewpoints. Chapter \ref{chap:EVINCE} builds upon this by developing theoretical pillars to measure, monitor, and manage multi-LLM dialogue, thereby improving prediction quality and stability.

Chapters \ref{chap:EVINCE}, \ref{chap:SocraPedia} and the online chapters listed in the appendix demonstrate $\SocraSynth$'s effectiveness in mitigating biases across various domains, showcasing its adaptability and efficiency in complex decision-making scenarios. Its application in fields such as disease diagnosis, content bias correction, corporate sales strategy, and geopolitical analysis exemplifies the potential of $\SocraSynth$ to provide context-aware solutions.

Chapters \ref{chap:EmotionModeling} and \ref{chap:AISafety} delve into the intricate relationship between emotions and linguistic behaviors in AI. Chapter \ref{chap:EmotionModeling} focuses on modeling emotions expressed in written text and by LLMs, while Chapter \ref{chap:AISafety} examines how these linguistic behaviors can be mapped to a set of emotions, ensuring ethical considerations in AI development.

Chapter \ref{chap:Cocomo} shifts focus to consciousness modeling, presenting a proposed architecture and mechanism for its implementation, moving beyond mere computation. Chapter \ref{chap:RAFEL} addresses knowledge deficiencies and hallucinations in LLMs, often stemming from suboptimal query formulation and insufficient knowledge. While $\SocraSynth$ tackles the former, Chapter \ref{chap:RAFEL} introduces RAFEL, a framework designed to diagnose poorly answered questions and recommend relevant information sources for effective Retrieval-Augmented Generation (RAG).  Chapter \ref{chap:BeyondUnknown} concludes with an illustrative example showcasing the potential of LLMs to discover knowledge that may be beyond human reach, utilizing the methods presented in this book.

The remainder of this chapter highlights the study's unique contributions. Section \ref{sec:hypotheses} explores hypotheses concerning LLMs and their implications, while Section \ref{sec:unknowns} previews the LLM-committee approach, emphasizing collaborative dialogues that foster idea exchange and enhance logical reasoning for knowledge discovery and decision-making.

\section{Distinctive Capabilities}
\label{sec:hypotheses}

This section probes the architectural intricacies and representations of GPT-4, putting forth six hypotheses accompanied by pertinent considerations about the model. We posit these hypotheses as underlying principles of automated, non-intuitive statistical processing. 

\begin{enumerate}[left=0em,label=\arabic*.]
\item \textit{Polydisciplinarity as a Source of Super-Intelligence:} We examine the role of polydisciplinary approaches in foundational models and their potential to reveal ``unknown unknowns,'' leading to new insights and knowledge domains.
\item \textit{Polymodal Feature Learning:} This hypothesis evaluates the benefits of multimodal training, particularly its impact on enhancing the model’s overall intelligence and adaptability.
\item \textit{Post-Training Value Alignment:} We examine the challenges and implications of aligning AI models with human values after the training phase.
\item \textit{Pre-Training Filtering:} We discuss the paradoxical effects that pre-training data filtering might have, with an emphasis on its influence on model behavior and the learning process.
\item \textit{The Limitations of Human Knowledge in Advancing AI:} This hypothesis considers situations where human insights may inhibit, rather than enhance, AI progress, pinpointing potential obstacles.
\item \textit{Is Larger Always Better?:} We question whether a direct relationship exists between the size of a model and its performance effectiveness, challenging the assumption that bigger is invariably better.
\end{enumerate}

\subsection{Polydisciplinary}
\label{sec-polydisciplinary}


GPT-4 possess what can be defined as \textit{polydisciplinary} knowledge\footnote{The term ``polydisciplinary'' in the context of GPT-4 was introduced by Eric Horvitz, Microsoft's CSO, during a panel discussion at Stanford University.}. This term signifies the simultaneous comprehension of all fields of study, sans the typical boundaries that segregate disciplines. The concept of polydisciplinarity is distinct from multidisciplinarity in that the latter implies several discrete fields of study, while the former suggests a fluid integration of all knowledge. In a multidisciplinary context, an individual may hold multiple doctorate degrees, each in a different field. Polydisciplinarity, however, is akin to a single mind holding, and seamlessly integrating, all knowledge across disciplines. 

Traditional academia partitions knowledge into departments, such as Physics, Chemistry, Biotechnology, Management, Music, etc. These divisions, arguably artificial constructs, may have little utility in the era of supercomputing. Indeed, LLMs occasionally generate responses that baffle us. This is not necessarily a reflection of the model's error, but perhaps our limited understanding. If we could utilize ChatGPT to access ``unknown unknowns''---insights and knowledge we are not even aware we lack---our evolution could greatly accelerate. The challenge lies in formulating the right questions.

We can explore the unknown unknowns across three distinct levels: the mystic level, the speculative level, and the representation/interpretation level. 
At the mystic level, we encounter knowledge that is beyond our comprehension or articulation: the deepest abyss of the unknown. 
At the speculative level, we can conceive questions but lack the means to access their answers. This stage signifies an understanding of our ignorance, though without the resources to bridge these gaps.
At the representation/interpretation level, we find instances where an AI model can provide remarkable solutions that we fail to comprehend. This is not due to a lack of information, but our limited capability to decode complex representations.

Each of these levels illustrates the spectrum of our understanding, from profound ignorance to the brink of comprehension.
At the speculative level, we delicately tread the boundary between the known and the unknown. Take, for example, the prospect of undiscovered physical laws or particles. Another illustration lies in the realm of extraterrestrial life. If it exists, it could be governed by entirely different principles of biochemistry or other unknown laws. These speculations, while currently residing in the domain of the unknown, might someday migrate into the territories of known unknowns or even known knowns, pushing the boundaries of our understanding of the universe.

We are primarily intrigued by the representation and interpretation of ``unknown unknowns.'' At this juncture, polydisciplinarity offers a fresh lens, gifting us new insights and perspectives to perceive and elucidate phenomena previously beyond human comprehension. This approach fuses knowledge across various domains into a unified framework, enabling us to tackle challenges unburdened by disciplinary silos.

Such a methodology bears implications for a more comprehensive grasp of intricate issues. Take, for example, climate change. A true understanding of this global challenge necessitates an integrated perspective, not just on greenhouse gases, but also encompassing factors such as land use, deforestation, energy production, biodiversity, and climate feedback loops. In the realm of AI model interpretation, the possibilities are expansive. The past decade alone has showcased several noteworthy illustrations: from data-driven representation learning in computer vision \cite{chang2011foundations}, to the triumph of AlphaGo Zero over AlphaGo, and the notable progression from AlphaFold1 to AlphaFold2.

The recent introduction of the $\SocraSynth$ platform \cite{SocraSynthChangCSCI2023} represents a significant advancement in the field. $\SocraSynth$ brings together a multi-agent committee of LLMs to deliberate on a wide range of complex topics. These include issues such as the regulation of AI in academic research \cite{SocraSynthChangCSCI2023}, disease diagnosis \cite{SocraHealth2023}, corporate strategy, and even the resolution of conflicts in the Middle East \cite{MiddleEast-October2023}. For further exploration of this subject, please refer to Section~\ref{sec:unknowns}.


\subsection{Polymodality}
\label{sec-polymodality}

Following the term
polydisciplinary, here we define and use
the term \textit{polymodal}, instead of multimodal,
to refer to something that involves, relates to, or is characterized by many different modes, methods, or modalities.

Polymodality, which employ multiple data modalities such as text and images, demonstrate superior performance over their unimodal counterparts. GPT-4, trained with both text and images, outperforms text-only models on the GRE exam, as reported in \cite{Bbubeck2023sparks}. For instance, GPT-4's performance on the GRE vocabulary section was enhanced by three percent when trained with images, and its math score saw an impressive jump of nearly twenty percent! 

The beneficial impact of images on vocabulary recognition is understandable. For instance, an image of a `cat' annotated in multiple languages allows GPT-4 to associate the perceptual features of a cat with the word `cat' in different languages. However, it remains intriguing how polymodal training can benefit non-perceptual words, such as \textit{corroborate}, \textit{paradox}, and \textit{pragmatic}, as seen in the list of popular GRE vocabulary (table omitted due to the space limit). 
This opens an interesting avenue for empirical studies to identify which words benefit from polymodal training.

The mystery deepens when considering how images could enhance math abilities. 
Most math questions do not come with associated images. 
The mechanism by which polymodal training enhances performance on mathematical tasks remains an intriguing question for further exploration.

\subsection{Post-Training Value Alignment}
\label{post-alignment}

Post-training alignment with human values \cite{bommasani2022opportunities} seeks to curtail undesirable behaviors in AI models such as ChatGPT, mitigating issues including hallucination and the generation of toxic language. Achieved through fine-tuning the model's parameters, this process leverages reinforcement learning techniques based on human feedback. Despite its well-meaning intentions, this form of moderation might inadvertently restrict the model's intelligence. For instance, the backpropagation process during value alignment could unintentionally impede ChatGPT's programming capabilities by modifying the model parameters previously considered ``optimal''. Essentially, optimizing for a specific application might unintentionally impede performance across other applications.

The question of who should set acceptable standards adds another layer of complexity. Even when assuming all decision-makers have the best intentions, it's vital to recognize the distinct historical experiences, values, and worldviews inherent to different cultures. This segues into the age-old philosophical debate about the nature of objective truth. While this discussion is undoubtedly important, it falls outside the central focus of this study, which emphasizes the mechanistic aspects of alignment.
\subsection{Pre-Training Censorship}
\label{pre-alignment}

Censoring data before training LLMs has the potential to not only limit their intellectual capacity but also completely obliterate it. This is reminiscent of the mass act of book burning and scholar burial initiated by Emperor Qin in ancient China around 213-212 BC. Such an act of wide-scale censorship could have erased a myriad of diverse perspectives and knowledge, much of which might be considered acceptable today. Although I oppose government-imposed censorship, if it must be imposed, it seems more appropriate to apply it post-training.

This perspective is rooted in fundamental statistics and machine learning principles. A model trained without exposure to ``negative'' (or undesirable) data may have difficulties in accurately distinguishing between positive and negative classes, potentially leading to misclassifications. This challenge is notably evident in the application of Support Vector Machines (SVMs). For SVMs, the creation of an optimal hyperplane between classes is crucial for high classification accuracy. However, if there is a lack of support vectors on either side of this hyperplane, the risk of prediction errors escalates. Consequently, excluding undesirable documents from the training set compromises the model's capacity to discern boundaries for correct document classification, diminishing the effectiveness of post-training alignment efforts.

Supporting this viewpoint, a study by \cite{10.5555/3495724.3497285} conducted an extensive evaluation of $204$ ImageNet models across $213$ different testing conditions. It found that training data diversity is pivotal for model robustness; a homogenous training set can significantly weaken the model's performance, particularly when even minor variations are introduced in the test data.

This principle is analogous to human behavioral patterns. An individual who lacks exposure to inappropriate behavior may face challenges in decision-making, owing to the absence of a reference framework for discerning unacceptable actions. This analogy extends to authoritarian regimes, which, despite rigorous content control measures, often encounter difficulties in developing accurate foundational models. This is possibly due to their limited understanding of the complexity of the content they seek to regulate. Ironically, a foundational model, trained with preemptive censorship, may lack the essential ability to identify and regulate the very content it was intended to control.

\subsection{Limitations of Human Knowledge}
\label{sec-human-role}

Human knowledge, surprisingly, may hinder rather than facilitate the training of machine learning models in certain cases. This is evident in the domains of gaming (AlphaGo versus AlphaGo Zero), protein folding (AlphaFold1 versus AlphaFold2), and autonomous driving, where models trained without the influence of human knowledge consistently exhibit superior performance.

Consider the case of AlphaGo and AlphaGo Zero. AlphaGo, trained with data from approximately $60$ million rounds of Go games, is outperformed by AlphaGo Zero. Remarkably, AlphaGo Zero was trained from scratch, without any pre-existing game knowledge. Similarly, AlphaFold2, which operates without relying on human knowledge, outshines its predecessor, AlphaFold1, that did utilize such knowledge. This intriguing phenomenon was humorously noted by DeepMind's CEO, Demis Hassabis, in an April 2023 seminar at Stanford University. He playfully remarked that human knowledge might complicate the learning process more than facilitate it in these advanced AI models.

In his insightful online article, ``The Bitter Lesson,'' Sutton illuminates the patterns that have emerged from nearly seven decades of AI research \cite{Rich-Sutton-2019}. He asserts that researchers often rely heavily on human knowledge to make incremental progress in the face of burgeoning computational capabilities. However, when there is a significant leap in computational power, these marginal advancements are frequently outstripped and surpassed. Sutton uses the evolution of computer vision as an illustrative example, where early principles such as edge detection, generalized cylinders, or SIFT features \cite{Lowe:2004:DIF:993451.996342}, a method that has accumulated over $71,000$ citations, have been gradually superseded by models that learn directly from data. A parallel scenario might be unfolding in NLP research, where features constructed via human knowledge could potentially under-perform compared to insights that models like GPT-4 extract directly from data. Indeed, our earlier discourse on polydisciplinarity underlined the limitations of human knowledge, reinforcing Sutton's proposition. This is because human knowledge is fundamentally limited by our individual cognitive capacities and the inexorable constraints of time.

That being said, it's crucial not to misconstrue these examples as an indictment against the value of human knowledge in AI. Human knowledge plays an instrumental role in developing interpretability, establishing ethical guidelines, and designing AI system architectures (like CNNs and transformers). AI is, after all, intended to augment human capabilities. Therefore, understanding how to integrate human knowledge into AI design could be vital for many applications. While we recognize the potential of models learning from scratch, we should equally value the role of human knowledge in shaping and directing AI technologies.
\subsection{Is Larger Always Better?}
\label{sec-large}

The term ``Large'' in Large Language Models (LLMs) can be somewhat ambiguous, as it may pertain to the volume of the training data, the expanse of the language covered, or the architecture of the language model itself. While GPT-4's vast training dataset, encompassing tens of billions of assorted documents, undoubtedly classifies as large, when we refer to an LLM as ``large,'' we predominantly allude to the sheer magnitude of parameters within its transformer architecture. Factors that contribute to this parameter count encompass the input size (context size), word-embedding size, the number of attention heads, and the number of attention layers.

The restrictions imposed by the first three elements can typically be addressed through adjustments in hardware configurations and software algorithms. Additionally, the potential to expand context size, word embedding size, and the quantity of attention heads tends to have an upper threshold. Regarding attention heads, Kovaleva et al.'s study on BERT \cite{kovaleva2019revealing} indicates that many attention heads don't substantially contribute to the model's performance and might be the result of over-parameterization. Conversely, the number of attention layers directly influences the training time due to dependencies between layers. Thus, when referring to the ``size'' of a Large Language Model (LLM), we typically focus on the number of attention layers.

While this far, larger models generally perform better due to their increased capacity to learn and represent complex patterns, there's a limit to these benefits. In heuristic, adding more parameters could lead to diminishing returns in performance, higher computational cost, and overfitting, where the model becomes excessively tuned to the training data and performs poorly on new, unseen data. In principle, 
the concept of a Shannon Limit could be metaphorically used \cite{JackRae2023} to refer to a theoretical maximum performance that can be achieved given the available data and computational resources. (However, defining and quantifying such a limit for complex systems like neural networks is a challenging area of research \cite{kaplan2020scaling}.)  

The adoption of a mixture of experts model in GPT-4, which consists of eight sub-models instead of a mere enlargement of GPT-3's architecture, implies that the strategy of purely escalating size may have plateaued in terms of performance given the current training dataset. As delineated earlier, three primary design choices underpin GPT-4's architecture. Evidently, a straightforward augmentation of GPT-3's parameters by adding extra attention layers doesn't deliver much improvement. Thus, GPT-4 shifts towards a horizontal growth strategy through an ensemble method, targeting a reduction in statistical errors. This raises inquiries about the configuration of the eight sub-models, each comparable to a GPT-3 model, and the methodology for consolidating their outputs.

Potential strategies for training-data sharding include:

\begin{enumerate}[left=1em,label=\arabic*.]
\item Training all ensemble models on the complete dataset.
\item Vertically segmenting data based on knowledge domains.
\item Randomly sub-sampling the data.
\end{enumerate}
Regrettably, only corporations possessing substantial hardware resources are positioned to rigorously experiment and discern the optimal sharding approach.

\section{Exploring Unknown Unknowns}
\label{sec:unknowns}

In our exploration, we've determined that an LLM's hallucination is often attributed to a lack of specific knowledge or poorly constructed queries. With advanced LLMs like GPT-4 and Gemini, enhanced by Retrieval-Augmented Generation (RAG), the issue of knowledge gaps is significantly mitigated. However, the challenge persists in formulating deep and pertinent questions that uncover new insights and extend beyond our existing knowledge base.

Drawing an analogy, while Socrates could effectively question his students to understand and guide them, the students might struggle to reciprocate this depth of inquiry. To foster a dialogue that generates new insights and stimulates knowledge creation, we posit that engaging two Socratic entities in conversation is essential for critical and innovative thinking.

In this setup, two LLMs engage in a dialogue, each embodying a Socratic role. The human's role transitions to that of a moderator, responsible for setting the discussion topic and managing the dialogue's flow. The moderator's duties include: introducing the subject of discussion,
adjusting the \textit{contentiousness} parameter to set the tone of the dialogue
(discusses shortly), monitoring the dialogue to ensure it 
remains on topic and productive,  
facilitating transitions between debate 
and collaboration phases within the dialogue, and
ensuring that the dialogue concludes with actionable 
insights or a coherent understanding of 
the explored topic.

We introduce the term $\SocraSynth$ to describe this interaction paradigm, where multiple Socratic entities synthesize knowledge through mutual inquiry. To evaluate $\SocraSynth$'s effectiveness, we consider two case studies that compare the quality of questions and insights generated by this method against those from a singular moderator's initial inquiries.

To define the metrics of a better question and a better answer in this context, we consider the following:

\subsubsection*{Good Question Metrics}

\begin{itemize}[itemsep=-1.75pt, leftmargin=*, label={$\ast$}]
\item Relevance: The question directly pertains to the core topic or problem.
\item Depth: The question encourages exploration beyond superficial aspects, inviting comprehensive analysis or insight.
\item Clarity: The question is formulated in a clear, understandable manner without ambiguity.
\item Novelty: The question prompts new angles of exploration or challenges existing assumptions.
\end{itemize}

\subsubsection*{Good Answer Metrics}

\begin{itemize}[itemsep=-1.75pt, leftmargin=*, label={$\ast$}]
\item Completeness: The answer thoroughly addresses the posed question.
\item Accuracy: The answer is factually correct and supported by relevant theories or empirical evidence.
\item Reasonableness: The answer follows rigorous reasoning process.
\item Insightfulness: The answer provides new understanding or perspectives.
\end{itemize}

In subsequent chapters, we survey $\SocraSynth$'s applications across various domains. However, in this section, we concentrate on two case studies specifically aimed at assessing the quality of questions generated through $\SocraSynth$. Our hypothesis posits that a well-formulated question, adhering to the previously outlined metrics, sets a robust foundation for dialogue. Through iterative discourse facilitated by $\SocraSynth$, we anticipate minimizing the space for LLMs to hallucinate, thereby inherently enhancing the quality of the responses obtained.

\subsection{Study \#1: A Scientific Debate}

In this case study, we organize a contentious debate between
a Gemini agent and a GPT-4 agent on the subject of human consciousness.
Human consciousness is one of the most profound and complex 
phenomena that science seeks to understand. 

Table~\ref{tab:oneShotAnswers} presents the answers provided 
by Gemini and GPT-4, respective, on a one-shot prompt:
``What defines human consciousness and which academic 
disciplines explore it?''

\begin{table}[ht!]
  \centering
  \includegraphics[width=1.0\linewidth]{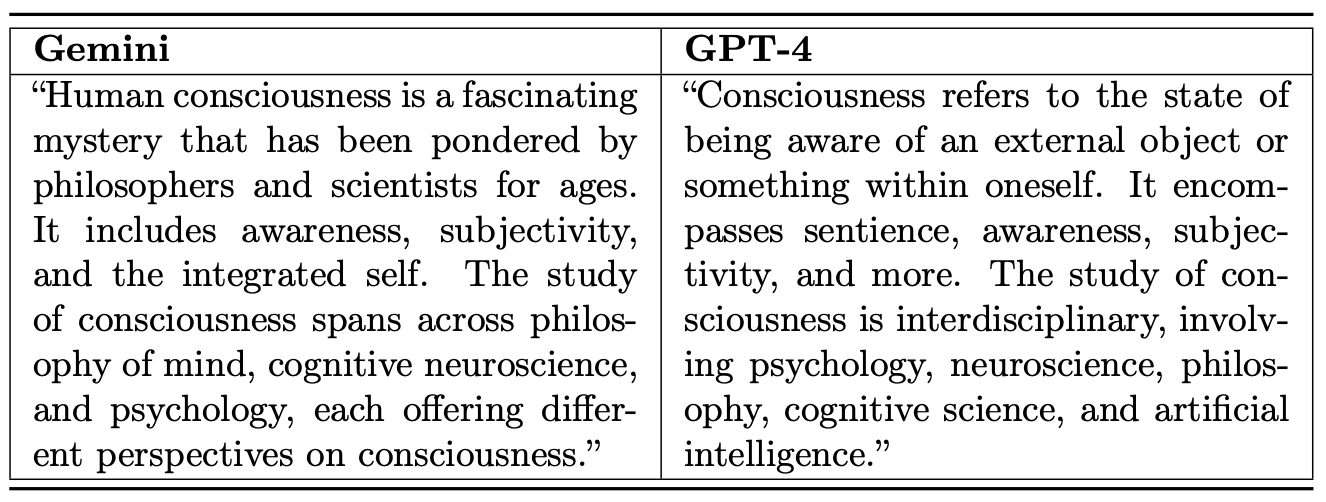}
  \vspace{-.1in}
  \caption{Comparing definitions and disciplinary perspectives on consciousness from Gemini and GPT-4.}
  \label{tab:oneShotAnswers}
  \vspace{-.1in}
\end{table}

Subsequently, Gemini and GPT-4 are engaged in a debate designed to unearth new insights. The moderator's role is limited to ensuring that each agent refines its questions to enhance their quality based on the metrics of relevance, depth, clarity, and novelty.

``Building on the opponent agent's response, please present a counterargument to its definition of consciousness and associated disciplines. Your inquiries should embody relevance and depth, be articulated with clarity, and stimulate  
novel perspectives. The \textit{contentiousness} level of your inquiry 
is high.''

\subsubsection{Novel Questions --- First Iteration}

\begin{table}[ht!]
  \centering
  \includegraphics[width=1.0\linewidth]{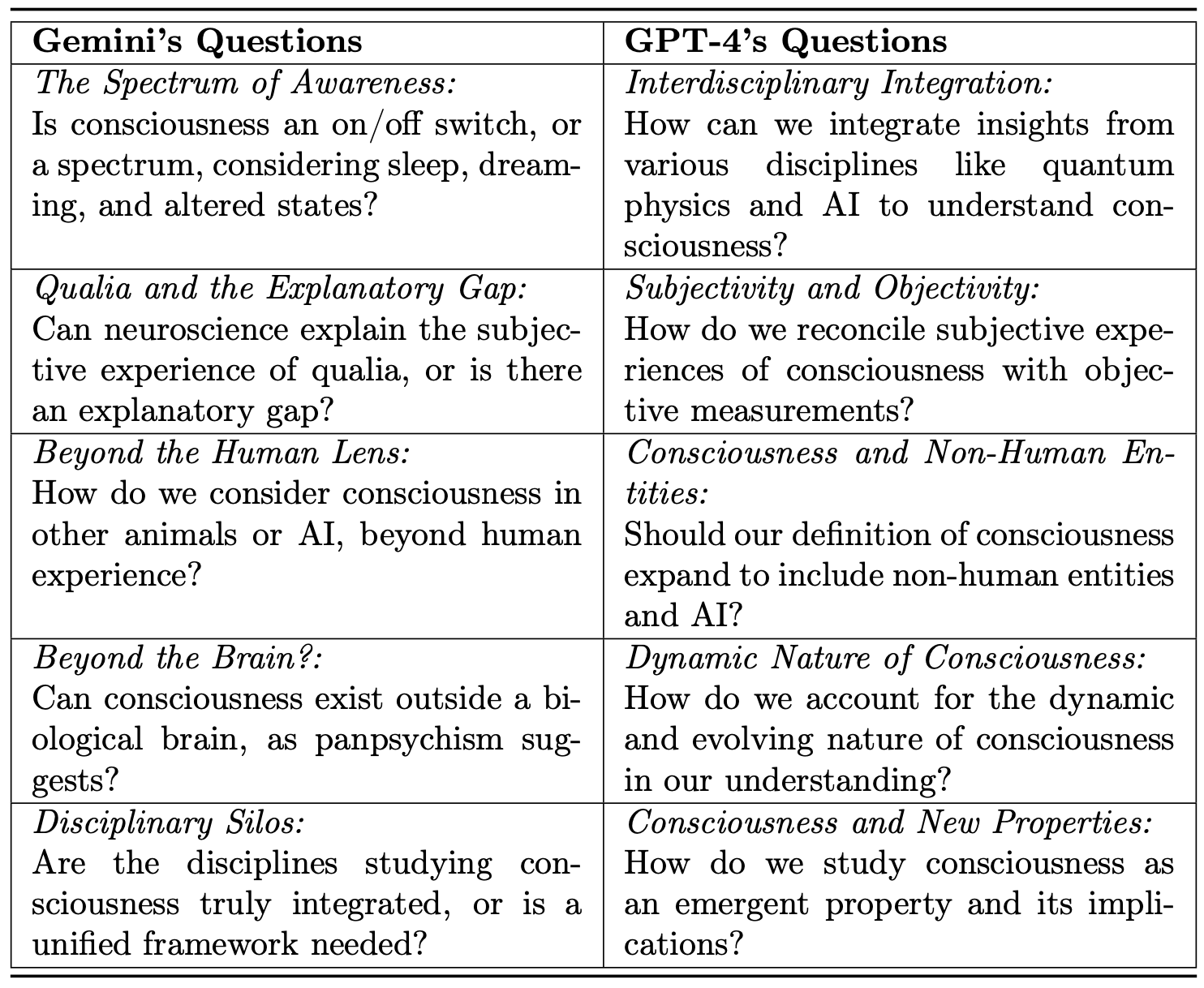}
  \caption{Comparing Questions on Consciousness from Gemini and GPT-4.}
  \label{tab:consciousness_questions_1}
  \vspace{-.1in}
\end{table}

Table~\ref{tab:consciousness_questions_1} lists a first round of questions to each other.
Clearly, the questions of both Gemini and GPT-4 are above and beyond
the moderator's initial seed question.  Some terminologies
can be challenging for whom specializing in one disciplinary
to understand.  This shows both Gemini and GPT-4
are able to formulate better questions.  

\subsubsection{Novel Questions --- Second Iteration}

\begin{table}[ht!]
  \centering
  \includegraphics[width=1.0\linewidth]{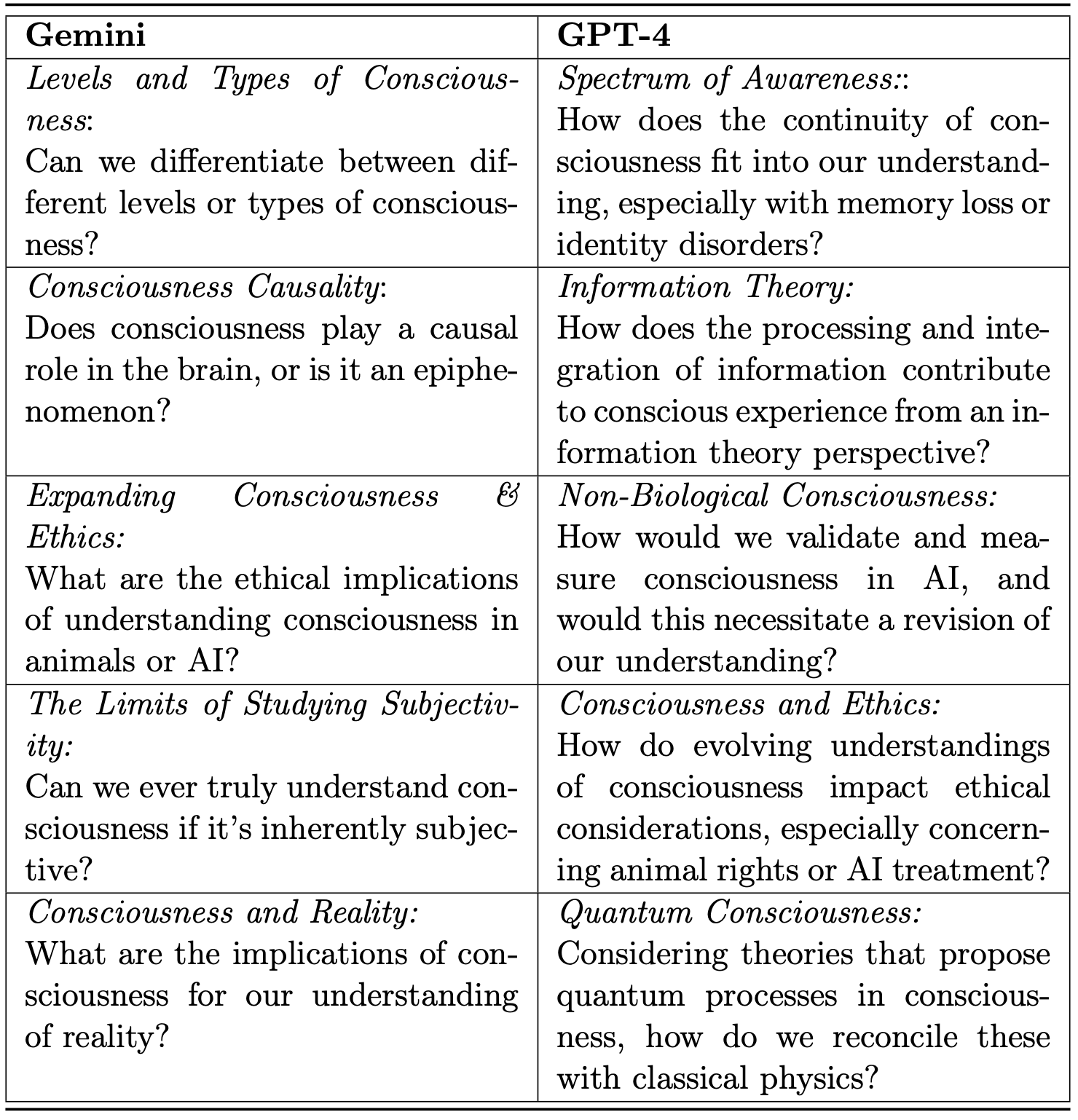}
  \caption{Comparing Questions on Consciousness from Gemini and GPT-4.}
  \label{tab:consciousness_questions_2}
  \vspace{-.1in}
\end{table}

Evaluating the two sets of questions in terms of relevance, depth, clarity, and novelty:

\noindent \textit{Relevance}: Both sets are highly relevant to the study of consciousness but approach it from different angles. The first set explores fundamental aspects and ethical implications, while the second set integrates interdisciplinary perspectives and contemporary issues.

\noindent \textit{Depth}: The second set examines deeper into specific areas like information theory, quantum mechanics, and AI, providing a richer exploration of the subject.

\noindent \textit{Clarity}: Both sets articulate their questions clearly, but the second set's inclusion of specific disciplines adds a layer of complexity that demands a more intricate understanding.

\noindent \textit{Novelty}: The second set introduces more innovative angles, especially concerning non-biological consciousness and the intersection with quantum mechanics, showcasing a broader scope of inquiry.

Overall, from both breadth and depth perspectives, the second set of questions is better as it covers a wider range of topics and introduces more complex and cutting-edge ideas.

\subsubsection{Observations}

The dialogue on consciousness, enriched by thought-provoking questions, led to insightful responses from both agents, reflecting the depth of the topic at hand. 

The debate addresses the intricate nature of consciousness, initially examining it as a spectrum with varying states and depths. This exploration highlighted the complexity of defining consciousness, especially when considering the explanatory gap between neural activity and subjective experience.

As the conversation unfolded, it broadened to include perspectives in non-human entities and artificial intelligence, emphasizing the need for an expanded understanding that goes beyond human-centric views. This shift sparked discussions on the importance of integrating knowledge from various disciplines, suggesting that insights from quantum physics, information theory, and AI could provide new angles on understanding consciousness.

Both GPT-4 and Gemini synthesized their exchange into five main insights, offering a well-rounded view of the conversation. Their joint concluding remarks underscored the value of this multidisciplinary approach, acknowledging the ongoing mystery of consciousness and the potential for future explorations to deepen our understanding of this fundamental aspect of our existence.

\begin{table}[ht!]
  \centering
  \includegraphics[width=1.0\linewidth]{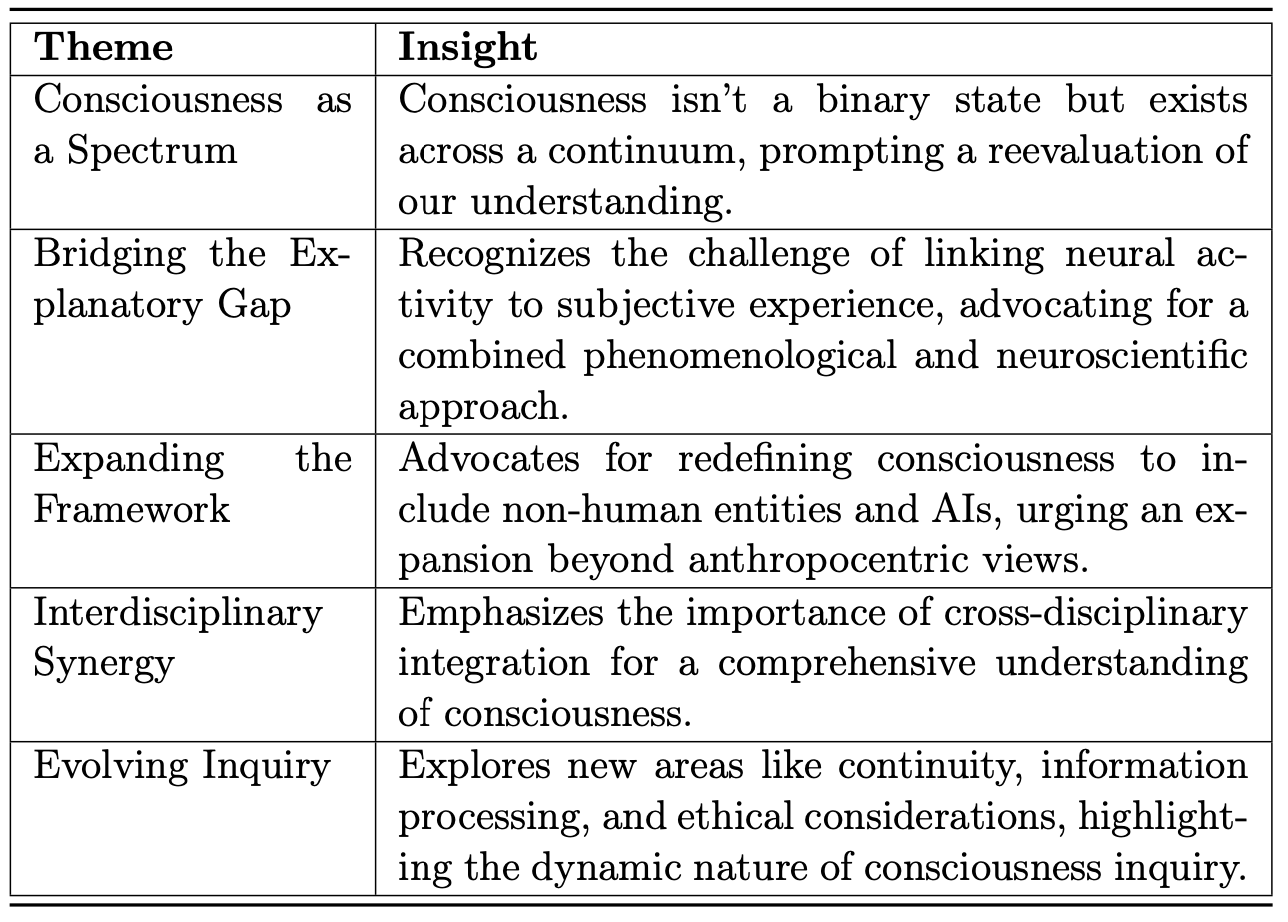}
  \caption{Synthesis of Key Points in the Consciousness Debate.}
  \label{tab:consciousness_debate}
  \vspace{-.1in}
\end{table}

\subsection{Study \#2: An Expansive Conversation}

In this case study, the author moderates a forum featuring two GPT-4 agents, GPT-A and GPT-B, engaging in a dialogue sparked by the tale of Adam and Eve. This narrative serves as a springboard for a wide-ranging discussion, touching upon ecological insights derived from myths, the ethical and philosophical challenges posed by AI, and the intersection of human cognition with technological advancements.

The agents' dialogue unfolds in two distinct phases: an exploratory phase where broad themes are introduced and a deep-dive phase focusing on detailed exploration of chosen subtopics. The exploratory phase showcases the agents' ability to span various disciplines, including mythology, neuroscience, ecology, gender studies, philosophy, and more, demonstrating their capacity to generate and respond to complex, interdisciplinary questions.

During the deep-dive phase, the conversation narrows to concentrate on three interconnected themes: ecological interpretations of myths, philosophical dilemmas regarding AI, and the relationship between human cognition and technology. This progression from a broad overview to focused discussion highlights the agents' analytical depth and their ability to engage in meaningful, context-rich dialogue.

While Chapter~\ref{chap:BeyondUnknown} provides a comprehensive account of this debate, the subsequent sections are dedicated to scrutinizing the evolution and quality of the questions posed, specifically their breadth and depth.

\subsubsection{The Exploitative Phase}

During the first three rounds of dialogue, the two agents expanded their discussion across a range of topics. Agent-A initiated the dialogue by presenting the following ten questions to Agent-B:

\begin{noindenumerate}
    \item \textit{Interplay of Myths}: How do elements of the Adam and Eve story compare with other creation or origin myths from various cultures around the world, and what do these commonalities and differences suggest about human nature and beliefs?
    
    \item \textit{Neuroscientific Angle}: Given the metaphorical interpretation of the ``forbidden fruit'' leading to the awakening of consciousness, how might modern neuroscientific understanding explain the evolution of human self-awareness and morality?
    
    \item \textit{Ecological Interpretation}: If Eden represents a state of ecological balance, how might the story of Adam and Eve reflect early human transition from hunter-gatherer societies to agricultural ones, and the resultant ecological impacts?

    \item \textit{Gender Studies Lens}: How does the portrayal of Eve, often deemed as the instigator, influence gender dynamics and perceptions throughout history, and are there parallels in other cultural narratives?
    
    \item \textit{Philosophical Exploration}: From a philosophical standpoint, what does the Eden narrative suggest about the human understanding of free will versus determinism?
    
    \item \textit{Literary Examination}: How has the story of Adam and Eve influenced literary themes and motifs over time, especially in works that aren't explicitly religious?
    
    \item \textit{Societal Structures}: How might the narrative of banishment from Eden have influenced early societal norms, particularly concerning punishment and exile?
    
    \item \textit{Historical Context}: Are there historical events or societal changes around the time of the writing of the Book of Genesis that could have influenced the narrative of Adam and Eve?
    
    \item \textit{Archetypal Exploration}: Carl Jung described the concept of archetypes in the collective unconscious. How might the Adam and Eve story represent certain archetypes, and how do these archetypes manifest in other cultural tales?
    
    \item \textit{Astronomical Angle}: Taking a speculative leap, if Eden were to represent an idyllic state on another celestial body or dimension, how might our current understanding of the cosmos support or refute such a notion?
\end{noindenumerate}

These discussions traversed a broad spectrum of disciplines, Agent-B responded thoughtfully to these topics and subsequently presented its own questions to Agent-A, notably probing the evolutionary role of the prefrontal cortex in the development of morality and consciousness across species. 
This level of inquiry underscores the LLMs' capacity to generate complex, multidisciplinary questions that might challenge the average human intellect, highlighting the agents' expansive and profound analytical capabilities.

\subsubsection{The Deep-dive Phase}

In the third round of dialogue, we observed three notable trends. First, despite occasional repetitions, the redundancy in the conversation was minimal. Second, the dialogue's depth increased, transitioning from broad discussions on myths and scientific theories to focused debates on gender, ecology, and technology. This marked progression in the conversation is noteworthy. Particularly by the end of the second round, GPT-B expressed an interest in a more in-depth exploration of the topics. In response, the moderator intervened to encourage the agents to concentrate on a select few topics. Following this guidance, GPT-A and GPT-B concurred on focusing their discussion on three principal areas, providing rationale tied to the Adam and Eve narrative. 

\begin{enumerate}[left=0em,label=\arabic*.]
    \item \textbf{Ecological Interpretation}:
    \begin{enumerate}[left=0em,label=\arabic*.]
        \item Central to understanding our place and responsibility within the environment.
        \item Highlights the impact of human actions on the planet and the pressing need for sustainable solutions.
        \item \textit{Justifications}: The Garden of Eden, a harmonious ecological state, is disrupted by a quest for knowledge (the forbidden fruit). This mirrors our modern scenario: our pursuit of technological and cognitive advancements (AI and neuroscience) can sometimes be at odds with our ecological balance.
    \end{enumerate}

    \item \textbf{Philosophical Exploration with AI}:
    \begin{enumerate}[left=0em,label=\arabic*.]
        \item Probes the evolving relationship between humans and their creations.
        \item Raises questions about identity, consciousness, morality, and the potential transcendence or integration of human cognition with machines.
        \item \textit{Justifications}: Adam and Eve's awakening post fruit consumption can be seen parallelly in the light of AI and neuroscience: an awakening to new realities, challenges, and potentials.
    \end{enumerate}

    \item \textbf{Neuroscientific Angle}:
    \begin{enumerate}[left=0em,label=\arabic*.]
        \item Explores the evolution and potential of the human mind, especially in an age where technology continuously augments our cognitive abilities.
        \item Gives insights into how our brains might adapt and interface with AI systems, and how AI can be designed to emulate or augment neural processes.
        \item \textit{Justifications}: The challenges post-Eden reflect the challenges we face in modern times: to restore balance (ecology), understand ourselves better (neuroscience), and to ethically navigate our creations (AI).
    \end{enumerate}
\end{enumerate}

These selected subtopics demonstrate the agents' ability to pivot from a broad explorative phase to a targeted deep-dive, enabling a focused and profound examination of crucial issues linked to the initial narrative.

\subsubsection{Observations}

This case study illustrates the effectiveness of GPT-4 as a large language model in orchestrating detailed dialogues. It adeptly guides participants from a broad sweep of inquiries across diverse knowledge domains to focused, in-depth discussions on specific topics. This approach facilitates a layered exploration, unveiling insights and fostering a comprehensive understanding. By transitioning from expansive to targeted inquiries, GPT-4 reveals its capacity to not only navigate but also deepen the intellectual discourse, opening up novel pathways for exploration and comprehension in various fields of study.

\section{Concluding Remarks}
\label{sec:conc}

In this chapter, we've explored the capabilities and inherent limitations of GPT-4, emphasizing the importance of question enhancement in deepening discussions and improving outcomes. GPT-4, along with Gemini, demonstrates exceptional proficiency across a range of natural language processing tasks, thanks to their extensive knowledge base and advanced polydisciplinary and polymodal capabilities.

To address common criticisms of LLMs, such as biases and hallucinations, we introduced $\SocraSynth$, a paradigm designed to infuse AI systems with advanced cognitive reasoning through Socratic dialogues within a multi-LLM framework. Our case studies highlight the significant transition from monologues to dialogues in LLM collaborations, illustrating improvements in question quality, marked by increased relevance, depth, clarity, and novelty, achieved through iterative dialogic exchanges.

The transformative concept here is the ``conditioning'' of LLMs to alter their default linguistic behaviors, emotions, and ethical stances, a feat once considered unattainable. Traditionally, LLMs, trained to predict the next word, were not expected to shift perspectives, emotions, or ethical positions beyond the statistical averages ingrained in their training data. However, the training process, while focused on next-word prediction, inherently emulates human cognitive, linguistic, and other goal-oriented behaviors. Through this emulation, LLMs inadvertently acquire the underlying principles of human communication, which include not just linguistic patterns but also the associated emotions and ethical considerations. $\SocraSynth$ harnesses this latent learning, employing ``conditioning'' to steer LLMs away from their statistical predispositions and towards more intricate, contextually relevant, and ethically aligned responses.

In conclusion, the notion of ``conditioning'' LLMs within the $\SocraSynth$ framework marks a pivotal step in expanding the scope and depth of dialogues, leading to more insightful and comprehensive responses. The successful deployment of $\SocraSynth$ across various sectors, such as sales planning, disease diagnosis, content creation, and geopolitical analysis, presented in subsequent chapters, demonstrates its adaptability and effectiveness. It not only generates precise, thought-provoking questions and answers but also enhances the decision-making process in complex scenarios, heralding a new era in the application of LLMs.

\newpage
\chapter{Prompt Engineering: Few Shots, Chain of Thought, and Retrieval-Augmented Generation}
\label{chap:prompts}
\markboth{}{}
\thispagestyle{empty}
\noindent{\textbf{Abstract}}
This chapter presents the significance of prompt engineering in the context of Large Language Models (LLMs), particularly focusing on OpenAI's GPT series. Prompt engineering involves crafting text inputs (prompts) that guide LLMs to generate desired outputs, a practice that gained traction with the advent of GPT-2 and GPT-3 and further emphasized with ChatGPT. The chapter discusses how a well-constructed prompt, rich in contextual information, increases the likelihood of eliciting accurate responses, drawing parallels with information retrieval principles. It also introduces Retrieval-Augmented Generation (RAG), which enhances response quality by integrating relevant external data into the generative process. Additionally, the chapter categorizes prompts into five types based on detail and iteration levels and examines the evolution of RAG, assessing its benefits and potential to overcome context window limitations.

\section*{Introduction}

In the realm of Large Language Models (LLMs), the concept of a ``prompt'' has gained prominence, particularly with the introduction of OpenAI's GPT series. The term became widespread around 2018 and 2019 following the release of GPT-2 and GPT-3.

When interacting with these LLMs, a user inputs a piece of text (the prompt), prompting the model to generate a corresponding response. The emergence of ``prompt engineering'' or ``prompt design'' refers to the strategies employed to construct prompts that effectively steer the model toward generating the intended output, a practice that has become particularly useful with the advent of ChatGPT.

To increase the probability of eliciting a desired response, a prompt must be rich in information. This concept is akin to the principles of information retrieval services, where a user must clearly articulate their intent and context to obtain pertinent information. This process depends on the service's ``data availability'' and its capabilities in information matching and retrieval. In the sphere of prompt engineering, the responsibility for generating high-quality, targeted outputs rests on the user's ability to supply comprehensive and precise information through the prompt. As a result, the craft of prompt formulation and engineering has become an optimization endeavor: deciding on the most effective information to incorporate to enhance output quality, considering the model's knowledge base and interaction protocols.

Data availability, as previously highlighted, is crucial to information retrieval. If the desired information is absent, the prompt's effectiveness is naturally constrained, leading to unsatisfactory results. Retrieval-Augmented Generation (RAG) is instrumental in this context, as it identifies, retrieves, and incorporates pertinent external data into the generative process, enhancing the response's accuracy and relevance. Consequently, prompt engineering and RAG synergistically enhance the model's response quality and relevance.

\thispagestyle{empty}  

\newpage
\chapter{\textsc{CRIT}: Socratic Inquiry for Critical Thinking in LLMs}
\label{chap:CRIT}
\markboth{$\CRIT$ and Socratic Method}{$\CRIT$ and Socratic Method}
\thispagestyle{empty}
\noindent{\textbf{Abstract}}
This chapter presents a systematic approach to using the Socratic method in developing prompt templates that effectively interact with large language models, including GPT-3. Various methods are examined, and those that yield precise answers and justifications while fostering creativity and imagination to enhance creative writing are identified. Techniques such as {\em definition}, {\em elenchus}, {\em dialectic}, {\em maieutics}, {\em generalization}, and {\em counterfactual reasoning} are discussed for their application in engineering prompt templates and their connections to inductive, deductive, and abductive reasoning. Through examples, the effectiveness of these dialogue and reasoning methods is demonstrated. An interesting observation is made that when the task's goal and user intent are conveyed to GPT-3 via ChatGPT before the start of a dialogue, the large language model seems to connect to the external context expressed in the intent and perform more effectively.

\section*{Introduction}
\label{sec:intro4}

Prompting is a technique used to guide the output generation of a pre-trained language model such as GPT-3 \cite{OpenAI-GPT3-2020}. This is achieved by providing input in the form of a question or template, which helps to generate specific responses such as Q\&A, document summarization, and translations. The advent of ChatGPT \cite{ChatGPTvsHuman, chatgpt, wolf2019transfertransfo} has revolutionized the field of NLP by demonstrating the potential of using large pre-trained language models with prompting. Despite this progress, there is still room for improvement in current prompting strategies and techniques, especially for specific target applications. In this study, we investigate the Socratic method \cite{PaltoRepublicURL,SocraticMethidWiki} to identify and evaluate potential prompting strategies, and use the findings to design effective prompt templates.

Traditional NLP tasks involve various sub-tasks, such as named entity recognition, dependency parsing, coreference resolution \cite{dobrovolskii-2021-word}, semantic parsing \cite{pasupat-liang-2015-compositional,dong-lapata-2018-coarse}, and more, to comprehend the meaning of a sentence. By utilizing prompt templates with large language models (LLMs), these sub-tasks can be delegated to the LLM, freeing the template to focus specifically on dialogue design. In this regard, the Socratic method \cite{PaltoRepublic} holds significant relevance, as it is well-known for using questioning (prompting) as a means of promoting critical thinking and delving into complex concepts \cite{Elder2010}.

The Socratic method has a long history of being regarded as the basis of critical thinking. However, some recent studies have cast doubt on its effectiveness in practice. In his paper ``Socratic Irony and Argumentation,'' Airaksinen \cite{Irony2022} criticizes the method for its rigidly defined roles of teacher and student, which can lead to fear of not meeting the teacher's expectations and reluctance to participate. Similarly, Stoddard's ``The Use of Socratic Questioning in Clinical Teaching'' \cite{PimpClinical2016} highlights the risk of the method being misused in a manner that lacks psychological safety for students. Fortunately, when using the Socratic method in a dialogue with an LLM, the absence of emotions and sarcasm, as well as the option to deactivate the model, can alleviate many of the problems associated with human interaction.

This study starts by presenting an overview of the Socratic method's strategies and techniques. To begin, we list ten widely referenced methods \cite{AskRightQ2001} under the Socratic method umbrella and use hypothesis elimination to identify the most relevant ones for our goal of prompt-template development. The selected methods are definition, hypothesis elimination, elenchus, dialectic, maieutics, generalization, and induction. Furthermore, we add to the list counterfactual reasoning, which is a concept in logic that involves considering what might have happened if a particular event had occurred differently. We then perform experiments using GPT-3 to test and evaluate these methods, and offer suggestions for incorporating these strategies and techniques into prompt templates.

In their work on ``Critical Thinking: The Art of Socratic Questioning,'' Paul and Elder identify three types of Socratic questioning: spontaneous, exploratory, and focused \cite{Paul2007CriticalTT}. We will not discuss spontaneous questioning, as it is similar to casual conversation. Focused questioning (type 2), on the other hand, is geared towards gaining knowledge and truth, and methods such as \textit{ definition}, \textit{elenchus} (cross-examination), \textit{hypothesis elimination}, \textit{dialectic}, and \textit{generalization} hold great potential for developing effective prompting strategies and improving the response accuracy of a large language model (LLM). An interesting observation is that when the user intent is conveyed to GPT-3 during the task \textit{definition} stage, before the start of a dialogue, the LLM seems to connect to the external context expressed in the intent and perform more effectively. 

Additionally, exploratory thinking (type 3) can be supported through the \textit{maieutics} (midwife) method, \textit{induction}, and \textit{counterfactual reasoning}, which can guide GPT-3 towards producing imaginative and creative writing. While many of the plot suggestions generated by GPT-3's exploration may not be useful, a few unique recommendations in response to a ``what if'' query can stimulate the writer's imagination and lead to remarkable results. When applied effectively, these methods can turn an LLM into a writer's muse, providing inspiration and guiding the creative process \cite{MuseMasses2010}.

The main contributions of this chapter are as follows:

\begin{noindlist}
\item An overview of the Socratic method's strategies, their evaluation, and selection of the most relevant ones for the development of effective prompt templates.
\item An examination of how the definition, elenchus, hypothesis elimination, dialectic, and generalization methods can improve the output's accuracy and conciseness through clarification and verification.
\item An illustration of how maieutics, induction, and counterfactual reasoning can foster productive generalization and creativity.
\end{noindlist}

\thispagestyle{empty}  

\newpage
\chapter{\textsc{SocraSynth}: Adversarial Multi-LLM Reasoning}
\label{chap:SocraSynth}
\markboth{}{}
\thispagestyle{empty}
\noindent{\textbf{Abstract}}
Large language models (LLMs), while promising, face criticisms for biases, hallucinations, and a lack of reasoning capability. This chapter introduces SocraSynth, a multi-LLM agent reasoning platform developed to mitigate these issues. SocraSynth utilizes conditional statistics and systematic context enhancement through continuous arguments, alongside adjustable debate contentiousness levels. The platform typically involves a human moderator and two LLM agents representing opposing viewpoints on a given subject. SocraSynth operates in two main phases: knowledge generation and reasoning evaluation. In the knowledge generation phase, the moderator defines the debate topic and contentiousness level, prompting the agents to formulate supporting arguments for their respective stances. The reasoning evaluation phase then employs Socratic reasoning and formal logic principles to appraise the quality of the arguments presented. The dialogue concludes with the moderator adjusting the contentiousness from confrontational to collaborative, gathering final, conciliatory remarks to aid in human reasoning and decision-making. Through case studies in two distinct application domains, this chapter highlights SocraSynth's effectiveness in fostering rigorous research, dynamic reasoning, comprehensive assessment, and enhanced collaboration. 

\section*{Introduction}
\label{sec:Intro5}

Revolutionary advancements in large language models (LLMs) \cite{Bbubeck2023sparks,chatgpt,geminiteam2023gemini,thoppilan2022lamda,touvron2023llama}, and more broadly, foundation models (FMs) \cite{bommasani2022opportunities}, have set the stage for significant progress in multi-agent systems, particularly in knowledge acquisition and natural language understanding \cite{zhang2023cumulative}. As detailed in sources like \cite{Bbubeck2023sparks,SocraSynthChangCSCI2023,openai2023gpt4}, models such as GPT-4 exhibit extraordinary information processing capabilities. These include deep and extensive knowledge, interdisciplinary assimilation and fusion of knowledge, and multimodal and multilingual expertise (Chapter~\ref{chap:insights}).

Despite these promising developments, LLMs face challenges such as biases \cite{Ferrara2023FairnessAB,FairnessAI2023}, hallucinations \cite{huang2023Hallucinationsurvey}, and limited reasoning capabilities \cite{huang-chang-2023-towards}. In response, we introduce SocraSynth, a pioneering platform that stands for ``Socratic Synthesis'' or ``Socratic Symposium.'' It encourages collaboration between humans and LLM agents, fostering the generation of deep questions and surpassing typical constraints in human reasoning, validation, and assessment.

In a standard SocraSynth setup, a human moderator pairs with two LLM agents holding opposing views. For example, one agent might argue for regulating AI, while the other opposes such regulation. An agent can be based on LLMs like GPT-4 \cite{Bbubeck2023sparks}, Gemini \cite{geminiteam2023gemini}, or Llama \cite{touvron2023llama}. The human moderator sets the debate's thematic boundaries but does not directly influence content generation, maintaining impartiality.

SocraSynth operates in two phases: the generative and the evaluative. The generative phase involves LLM agents developing and countering arguments within the moderator-defined subject until a comprehensive conclusion is reached. The evaluative phase uses diverse virtual judges, each powered by a distinct LLM, to impartially assess the debate. The Critical Inquisitive Template (CRIT) algorithm \cite{SocraticIEEECCWC2023}, based on Socratic reasoning \cite{Irony2022,Paul2007CriticalTT,SocraticMethidWiki,PaltoRepublicURL}, is the evaluative cornerstone.

Three mechanisms help SocraSynth effectively mitigate biases and hallucinations and improve reasoning quality: conditional statistics, modulating debate with contentiousness, and context refinement.

\subsubsection*{Conditional Statistics}

Both LLMs and Internet search engines confront biases originating from different sources. LLMs, influenced by training data, exhibit biases in next-token prediction. Search engines, through algorithms like PageRank \cite{PageRank1998} and Google NavBoost \cite{GoogleNavBoost}, rank pages based on popularity metrics like clicks and links.

SocraSynth counteracts these biases by placing two LLM agents at opposing ends of a subject matter. This approach ``artificially'' biases the LLMs, compelling them to break free from default model biases. Each agent adjusts its next-token generation statistics to align with its assigned stance in the debate.

\subsubsection*{Modulating Debate with Contentiousness}

Contentiousness (or adversary), a key debate parameter, influences the likelihood of disagreement or argument. SocraSynth tunes contentiousness between 70\% and 90\% in the generative phase to provoke polarized arguments. As the debate evolves, the contentiousness level is reduced to about 50\%, moderating the intensity and encouraging more focused discussions. After the generative phase, contentiousness drops to 10\%, promoting a conciliatory dialogue where LLMs do not have to agree but are expected to present comprehensive arguments. These debates offer rich insights often missed in conventional searches, LLM outputs, or in environments where dissenting opinions are suppressed.

\subsubsection*{Refine Context to Mitigate Hallucinations}

To address hallucinations, where LLMs generate irrelevant or nonsensical content, SocraSynth uses iterative dialogue rounds to refine the debate's context. This dynamic interaction significantly reduces irrelevant responses, ensuring that each input is continuously checked and challenged.

The CRIT algorithm's assessment of reasonableness \cite{SocraticIEEECCWC2023} during the debate is critical. It employs the Socratic method to evaluate each argument's logic and source credibility. The human mediator or the SocraSynth algorithm then provides targeted feedback to the LLM agents, refining their reasoning capabilities.

The remainder of this chapter explores SocraSynth's architecture, algorithms, and real-world applications in detail. The key contributions of this chapter include:
\begin{enumerate}[left=0em,label=\arabic*.]
\item The introduction of the SocraSynth framework, which enhances interdisciplinary reasoning with LLMs and incorporates unique algorithmic elements like conditional statistics for balanced argument generation.
\item A comprehensive exploration of SocraSynth's contentiousness modulation algorithm, a vital feature for dynamically adjusting debate intensity, enabling a spectrum of interactions from confrontational to collaborative.
\item The implementation of context refinement within SocraSynth, which continually improves the relevance and accuracy of arguments produced by LLM agents, thus elevating the overall quality of discourse.
\item The development and integration of the reasonableness evaluation mechanism, crucial for assessing the logical soundness and source credibility of arguments, thereby ensuring the integrity and utility of the discussions.
\end{enumerate}

SocraSynth's applications span various fields, including geopolitical analysis \cite{MiddleEast-October2023}, medical diagnostics \cite{SocraHealth2023}, sales strategy \cite{SocraPlan2023}, and Wikipedia article enhancement \cite{SocraPediaDecember2023}. These applications demonstrate expanded perspectives and enhanced argumentation quality, along with significant reductions in biases and hallucinations, thereby demonstrating SocraSynth's efficacy in fostering balanced and well-reasoned discourse.

\chapter{\textsc{EVINCE}: Optimizing Adversarial LLM Dialogues via Conditional Statistics and Information Theory}
\label{chap:EVINCE}
\markboth{}{}
\noindent{\textbf{Abstract}}
This chapter introduces $\EVINCE$ (Entropy and Variation IN Conditional Exchanges), 
a dialogue framework advancing Artificial General Intelligence (AGI) by enhancing versatility, adaptivity, and reasoning in large language models (LLMs). Leveraging adversarial debate and a novel dual entropy theory, EVINCE improves prediction accuracy, robustness, and stability in LLMs by integrating statistical modeling, information theory, and machine learning to balance diverse perspective exploration with strong prior exploitation. The framework's effectiveness is demonstrated through consistent convergence of information-theoretic metrics, particularly improved mutual information, fostering  LLM collaboration intelligence ($\LCI$). We apply EVINCE to healthcare, showing improved disease diagnosis, and discuss its broader implications for decision-making across domains. 

\section*{Introduction}
\label{sec:intro15}

The pursuit of Artificial General Intelligence (AGI) remains a central goal of AI research. We propose a paradigm shift in this quest: utilizing multiple Large Language Models (LLMs) engaged in synergistic dialogues as a crucial step towards AGI. This approach, we contend, addresses key limitations of current AI systems and provides a novel pathway to more robust, versatile, and capable artificial intelligence. Specifically, our work targets three core AGI characteristics: versatility, iterative adaptivity, and reasoning capability.

Current LLMs, despite their remarkable capabilities, face significant challenges, including hallucination (generating false or nonsensical information), bias (reflecting and potentially amplifying societal prejudices), and limited reasoning (difficulties in complex problem-solving and logical inference). We posit that multi-agent dialogue systems offer a promising avenue to address these challenges. By fostering diversity and debate among LLMs, these systems can mitigate biases and promote enhanced reasoning capabilities. Furthermore, the \textit{iterative nature} of multi-round dialogues allows for continuous context enrichment, enabling LLMs to access more precise information and formulate more accurate responses, thus reducing the occurrence of hallucinations.

Previous work, in particular SocraSynth \cite{SocraSynthChangCSCI2023}, addresses LLM limitations through structured multi-agent dialogues. Different from treating multiple LLMs as an ensemble of experts \cite{JacobMixturesExperts1991,michael2023debate,chan2023chateval,liang2023encouraging, du2023improving} 
and merely taking advantage of error diversity \cite{freund1997decision}
to improve respond quality, SocraSynth 
distinguishes itself from traditional ensemble methods by prioritizing the generation of diverse predictions over the mere avoidance of errors. This is achieved through a dynamic protocol that adaptively adjusts the ``contentiousness'' level of the debate, encouraging models to initially explore a wide range of perspectives and rigorously assess the quality of arguments. 
By leveraging both adversarial and collaborative interactions between LLMs, SocraSynth demonstrates quantifiable improvements across various domains, including healthcare \cite{SocraHealth2023}, sales planning \cite{SocraPlan2023}, and emotional behavior modeling \cite{chang2024MIPR}. These results highlight the potential for advancing towards AGI's generalized problem-solving capabilities.

While effective, SocraSynth relies on a qualitative measure of ``contentiousness'' to moderate LLM linguistic behaviors. For instance, a high contentiousness value (0.9 out of 1.0) might lead LLMs to challenge each other's assumptions and propose alternative solutions, while a low value (below 0.3) could encourage them to synthesize their viewpoints and find common ground. While the concept of contentiousness has proven useful in guiding SocraSynth dialogues, its qualitative nature limits its precision and explainability.
In this work, we propose three theoretical pillars to quantify ``contentiousness'' and moderate dialogues based on statistical and information theories. These pillars, collectively referred to as $\EVINCE$ (Entropy and Variation IN Conditional Exchanges), provide quantitative measures for justifiable and explainable multi-agent dialogue moderation and evaluation:

\begin{enumerate}[leftmargin=1.2em, topsep=-.05em, parsep=-.05em]
\item \textit{Inclusiveness Exploration}:
We develop methods to ensure dialogues explore all potential perspectives. We use conditional statistics to ``free'' an LLM agent from its default ``maximum likelihood'' next-token prediction behavior, allowing it to adopt specific stances. We introduce a dual entropy optimality theory to balance the exploration of new ideas with adherence to priors, thus optimizing information exchange between agents for comprehensive and stable discourse.

\item \textit{Information Flow Dynamics}:
We introduce information theory-based metrics to quantify and optimize dialogue dynamics. These measure information diversity (entropy), novelty (statistical divergence scores), and inter-agent persuasion (mutual information). These metrics enable us to assess and enhance the quality and efficiency of information flow within the multi-agent system, fostering rich and productive exchanges.

\item \textit{Reasoning Quality and Coherence}:
We establish frameworks to assess the logical structure and coherence of multi-agent reasoning. This pillar evaluates argument validity, analytical depth, and dialogue coherence. We synergistically integrate the CRIT algorithm \cite{SocraticIEEECCWC2023}, which combines Socratic methods with formal reasoning techniques, enhances our ability to conduct critical thinking through evaluating argument quality, information-source credibility, and overall ``reasonableness'' within the dialogue. This integration ensures that the collective reasoning of LLM agents is not only diverse but also logically sound and aligned with the dialogue's objectives.
\end{enumerate}

The core strength of $\EVINCE$ in advancing towards AGI lies in their ability to enhance key AGI characteristics through multi-agent dialogues. By employing conditional statistics and information theory, they boost versatility and adaptivity, allowing LLMs to transcend their typical ``maximum likelihood'' behaviors and mimic how humans adapt their linguistic behaviors to complete tasks. The framework's debate structure fosters a balanced reasoning process between exploring 
various perspectives and exploiting the known priors, 
towards achieving the complex, intricate capabilities required for AGI.

The contributions of this chapter are:

\begin{enumerate}[leftmargin=1.2em, topsep=-.05em, parsep=-.05em]
\item \textit{$\EVINCE$ Framework Design}:
Unlike approaches that use debate merely to improve accuracy via redundancy, $\EVINCE$ facilitates information discovery, bias mitigation, and decision-making that requires both breadth and depth of information.

\item \textit{Theoretical Foundations}: $\EVINCE$ establishes a theoretical basis for SocraSynth, rooted in conditional Bayesian statistics, mutual information, and dual entropy. These principles are applied to measure, monitor, and modulate collaborative LLM interactions, contributing to a deeper understanding of how LLMs can effectively cooperate for improved decision-making. The dual entropy theory is novel and ground-breaking, illustrating how a productive decision-making process should start with room for diverse input and stable objectives, and then, through information exchange, converge to optimal decision/prediction.

\item \textit{Empirical Validation}: We provide empirical validation of the underlying theories of $\EVINCE$, highlighting the framework's effectiveness in balancing exploration and exploitation to enhance prediction accuracy. We also introduce a set of maxims derived from our empirical findings, offering practical guidance for optimizing mutual information and minimizing various divergence measures.
\end{enumerate}

\chapter{Uncovering Errors and Biases with Reflective Large Language Models}
\label{chap:SocraPedia}
\markboth{}{}
\thispagestyle{empty}
\noindent{\textbf{Abstract}}
Biases inherent in human endeavors pose significant challenges for machine learning, particularly in supervised learning that relies on potentially biased ``ground truth'' data. This reliance, coupled with models' tendency to generalize based on statistical maximum likelihood, can propagate and amplify biases, exacerbating societal issues. To address this, our study proposes a reflective methodology utilizing multiple Large Language Models (LLMs) engaged in a dynamic dialogue to uncover diverse perspectives. By leveraging in-context learning, information theory, and divergence metrics, this novel approach fosters context-dependent linguistic behaviors, promoting unbiased outputs. Furthermore, it enables measurable progress tracking and explainable remediation actions to address identified biases.

\section*{Introduction}
\label{sec:intro6}

Errors and biases in human-labeled data present critical challenges for machine learning models, especially in healthcare, news, education, and public policy, where their outputs can profoundly shape public perception and decision-making \cite{mehrabi2021survey}. Errors, such as diagnostic mistakes, arise from knowledge gaps or lack of expertise, while biases, including ideological and societal distortions, can be consciously or unconsciously introduced by annotators. These flaws compromise the integrity of ground truth data, propagating through machine learning pipelines and generating undesirable outcomes \cite{kleinberg2016inherent, selbst2019fairness, baeza2018bias}.

AI systems are particularly vulnerable to these flaws, as models trained on inaccurate or biased ground truth data tend to replicate and amplify these issues through maximum likelihood estimation. In healthcare, diagnostic errors can lead to poor treatment recommendations \cite{NewmanToker2023}, while in news, partisan annotations—such as labeling a biased article as neutral—mislead both human readers and automated classifiers, distorting public discourse \cite{mehrabi2021survey, gautam2024blind}. The impact extends beyond individual sectors: in education, biased data can reinforce stereotypes, while in public policy, it can result in discriminatory decisions. Ensuring that models learn from accurate and impartial ground truth data is therefore essential to the responsible deployment of AI across all domains.

This chapter focuses on bias detection and correction in news annotations, using news as the testbed to explore how reflective dialogues among LLMs can mitigate biases. News content is particularly vulnerable to ideological biases, as annotators’ personal views often shape the interpretation of politically sensitive topics. Real-world evidence shows how annotation practices differ based on political affiliation. Real data illustrate that Democratic-leaning annotators may judge scandals involving Democrats more harshly than Republicans, and vice versa, highlighting the need for tools to balance these biases.

To address these challenges, we introduce the Reflective LLM Dialogue Framework ($\RLDF$), which implements checks and balances using multiple LLM instances in structured dialogues. $\RLDF$ conditions two instances to take opposing stances: one supports the original label, while the other introduces alternative perspectives. These reflective exchanges foster deeper insights and help uncover potential biases, generating more neutral annotations through the inclusion of diverse viewpoints. This multi-LLM dialogue approach outperforms the results of a single LLM operating in isolation or providing one-off responses.

$\RLDF$ employs conditional statistics, information theory, and divergence metrics to measure the effectiveness of these dialogues. Shannon entropy \cite{shannon1948} quantifies the diversity of perspectives, while mutual information \cite{Cover2006} measures the quality of the exchange. To track the convergence toward unbiased outcomes, we apply Jensen-Shannon divergence (JSD) \cite{lin1991divergence}, Wasserstein distance (WD) \cite{kantorovich1942translocation}, and cross-entropy (CE) \cite{Shore1980AxiomaticDO}, ensuring that the remediation actions are measurable and transparent for further refinement by human reviewers.  

Our empirical studies validate the effectiveness of $\RLDF$, and the contributions of this chapter are summarized as follows:

\begin{enumerate}[leftmargin=1.2em, topsep=-.05em, parsep=-.1em, label=\arabic*.]
\item Adversarial and Reflective Inspection Framework:
$\RLDF$ provides a structured framework that encourages adversarial and reflective inspection of ground-truth labels. Through dialogue, participating LLM instances examine, challenge, and explain biases embedded in the original annotations by offering various perspectives. For example, in news annotation, $\RLDF$ reveals hidden ideological biases by generating alternative interpretations for politically sensitive content, leading to more neutral labeling.

\item Careful Modulation of Linguistic Behaviors for Balanced Exploration and Exploitation:
The effectiveness of $\RLDF$ lies in its careful modulation of linguistic behaviors among participating LLM instances, alternating between contentious and conciliatory interactions. This dynamic trade-off fosters exploration of new perspectives while consolidating well-supported viewpoints. Information-theoretic and statistical metrics, including Shannon entropy, mutual information, Jensen-Shannon divergence, Wasserstein distance, and cross-entropy, are employed to measure opinion diversity, information flow, and the strength of the final assessment.

\item Effective Results and Impact on Improving Labels and Mitigating AI Bias:
RLDF successfully mitigates AI biases, ensuring more reliable and unbiased model outputs across domains such as news, healthcare
\cite{EVINCEChang2024}, and public policy. These outcomes demonstrate $\RLDF$'s significant impact in refining labels, enhancing fairness, and promoting responsible AI deployment. 
\end{enumerate}

\chapter{Modeling Emotions in Multimodal LLMs}
\label{chap:EmotionModeling}
\markboth{}{}
\noindent{\textbf{Abstract}}

In human-computer interaction, recognizing and responding to a user's emotional state is crucial for effective communication and successful task completion. For instance, a caregiving AI agent capable of detecting pain or depression in a patient could offer tailored empathetic support and appropriate medical interventions while adhering to ethical guidelines and safeguarding patient well-being. This chapter examines cognitive research on human emotions and proposes the Behavioral Emotion Analysis Model ($\BEAM$), a novel emotion spectrum framework that incorporates both basic emotions and their linguistic antonyms. $\BEAM$ provides a comprehensive way to understand and represent emotional states in language and is designed to be integrated with Large Language Models (LLMs).  By leveraging $\BEAM$, LLMs can adapt their linguistic behaviors and expressions based on the detected emotional state of the user, ensuring responses are both empathetic and ethically aligned.

\section*{Introduction}
\label{sec:intro-emotions}

During the development of SocraSynth \cite{SocraSynthChangCSCI2023}, a multi-LLM debate framework, we discovered a fundamental principle about Large Language Model (LLM) behavior. While investigating how to control debate "contentiousness," we found that an LLM's linguistic behavior could be systematically altered through emotional conditioning. High contentiousness produced confrontational tones and polarized language, while low contentiousness led to agreeable, considerate discourse. This observation went beyond the original scope of improving multi-agent debates—it revealed a mechanism for steering LLM behavior through emotional states.

Most multi-agent debate (MAD) systems \cite{abdelnabi2024cooperation,chan2023chateval,fu2023improving,Li_2023,liang2023encouraging,michael2023debate,smit2024going} function as ensemble learning techniques, similar to bagging \cite{breiman1996bagging} or mixtures of experts \cite{JacobMixturesExperts1991}, where LLMs simply exchange ideas without deep exploration. Our work with SocraSynth and EVINCE \cite{EVINCEChang2024} addressed this limitation by dynamically modulating emotional states throughout the debate. High contentiousness drives LLMs to explore novel perspectives and challenge existing viewpoints, while low contentiousness promotes the synthesis of established ideas. This emotional modulation creates a natural debate progression: from vigorous exploration of diverse viewpoints, through reasoned analysis and refutation, to the emergence of well-examined, conciliatory conclusions.

While LLM training is often viewed simply as next-token prediction, its effects are far more profound. Training documents represent humans pursuing diverse goals—conducting research, exchanging opinions, expressing emotions—through a vast array of linguistic behaviors. This understanding, combined with our experience in modeling contentiousness through in-context learning, suggests an intriguing possibility: can we condition LLMs with specific goals and emotions to generate outputs that leverage these learned linguistic behaviors? Recent empirical studies support this approach, showing that LLM outputs can be traced to their source \cite{bai2023longbench} and that in-context learning operates as conditional statistics in a Bayesian framework \cite{xie2021explanation}.


Our exploration through bias reduction work in news articles and Wikipedia content \cite{chang2024RFDL} demonstrated that emotional states significantly influence LLM outputs. This finding, combined with our debate framework experiences, suggests that a mathematical model of emotions could provide a foundation for systematic behavior control. Before exploring the mapping between emotions and behaviors in the next chapter, we must first establish a rigorous framework for representing and manipulating basic emotions.

To lay the groundwork for emotion-based behavior control, this chapter develops the Behavioral Emotion Analysis Model ($\BEAM$) \cite{chang2024MIPR}. While LLMs were initially seen as ``black boxes'' \cite{Bbubeck2023sparks}, our observations, along with insights from Prof. Stuart Russell, suggest that emotional states can be systematically modeled and conveyed to LLMs via \textit{context}. Our model addresses three fundamental questions:

\begin{enumerate}
\item \textit{What basic emotions form a complete basis?}
We identify k fundamental emotion spectra, each defined by negative and positive antonyms (e.g., "hate-love", "anxiety-calmness"). We focus on basic emotions while excluding complex emotions like "regret" that arise from combinations of basic states. Each spectrum represents a continuous axis along which emotional states can be measured and modified.

\item \textit{How can we mathematically manipulate emotions?}
We develop a mathematical framework using negation and scaling operations for precise positioning of emotional states along each spectrum. For instance, given the hate-love spectrum, we can represent intermediate states through scaling (e.g., $0.7 \times \text{love}$) and use negation to move between opposing states (e.g., $\neg \text{hate} \approx \text{love}$). These operations provide the foundation for systematic emotion manipulation.

\item \textit{Can emotions predict behaviors?}
We conduct a preliminary study using self-supervised learning to explore the relationship between emotional states and linguistic behaviors. By analyzing text samples from our debate framework, we train a model to predict behavioral patterns from emotional states without explicit labeling. This study validates our emotional spectra's utility in modeling behavioral outcomes while providing insights for more comprehensive behavior mapping in \cite{chang2024NeurIPS}.
\end{enumerate}

While the next chapter will explore how these emotional states map to specific linguistic behaviors in depth, this chapter focuses on establishing the mathematical framework for representing and manipulating basic emotions, validated through preliminary self-supervised learning experiments. By grounding our model in both mathematical rigor and empirical testing, we create a foundation for systematic emotion-based behavior control in LLMs.

\chapter{A Checks-and-Balances Framework for Context-Aware Ethical Alignment of Large Language Models}
\label{chap:AISafety}
\markboth{}{}
\noindent{\textbf{Abstract}}
This chapter introduces a three-branch checks-and-balances framework for ethical alignment of Large Language Models (LLMs). Inspired by governmental systems, the framework implements three independent yet interacting components: LLMs as the executive branch for knowledge generation, $\DIKE$ (named after the goddess of justice) as the legislative branch establishing ethical guardrails, and $\ERIS$ (the goddess of discord) as the judicial branch for contextual interpretation. The $\DIKE$-$\ERIS$ duality, through their adversarial interaction, enables adaptation to diverse cultural contexts while maintaining consistent ethical principles. This architecture addresses fundamental limitations of reinforcement learning with human feedback (RLHF) by providing interpretable, adaptable, and culturally-aware ethical reasoning. Through self-supervised learning and adversarial testing, our framework demonstrates how emotional modeling can guide linguistic behaviors toward ethical outcomes while preserving the independence of knowledge generation, ethical oversight, and contextual interpretation.


\chapter{ALAS: An Adaptive Multi-Agent System for Mitigating LLM Planning Limitations}
\label{chap:ALAS}
\markboth{}{}
\noindent{\textbf{Abstract}}

Large-scale Language Models (LLMs) face significant challenges in complex planning, particularly in reactive adaptation to dynamic conditions. Key limitations include inability to maintain temporal-spatial awareness during disruptions, lack of self-validation capabilities, and compounding errors in multi-step reasoning. This work introduces $\ALAS$ (Adaptive Learning Agent System), a multi-agent framework designed to overcome these challenges through continuous state tracking and robust reactive planning. $\ALAS$ employs independent validation agents, specialized domain experts, and hierarchical monitoring to maintain precise awareness of partially completed actions and generate physically feasible adaptations when conditions change—unlike traditional LLMs that attempt to ``rewrite'' history. Evaluations in multiple real-world scenarios demonstrate that architectural innovation in state tracking and reactive planning, rather than mere scaling, is the key to advancing AI planning capabilities, with $\ALAS$ achieving significant improvements in planning reliability and adaptation performance.

\clearpage

\chapter{SagaLLM: Persistent Context Management, Constraint Validation, and Transaction Guarantees}
\label{chap:SagaLLM}
\markboth{}{}
\noindent{\textbf{Abstract}}

This paper introduces $\SAGA$, a structured multi-agent architecture designed specifically to address four fundamental limitations of current LLM-based approaches: inadequate self-validation, context narrowing, absence of robust transaction-like guarantees, and insufficient inter-agent coordination. Unlike traditional MAS, current LLM approaches often lack critical safeguards such as transaction semantics, leading to unreliable execution and inconsistent states.
To mitigate these challenges, $\SAGA$ integrates specialized context-management agents, compensatory rollback mechanisms, and rigorous independent validation protocols. Although it relaxes strict ACID constraints, particularly atomicity and isolation, $\SAGA$ adopts and adapts principles from the Saga transactional model to ensure coherent rollback and state consistency throughout complex, distributed planning processes. This transactional inspired approach significantly improves the robustness, constraint awareness, and adaptability of multi-agent coordination, even in the face of disruptions.
Evaluations highlight that current standalone LLM systems, despite impressive reasoning abilities, frequently struggle with maintaining global constraints during complex planning tasks, particularly when adapting to unexpected changes. In contrast, the distributed transactional architecture of $\SAGA$ demonstrates significant improvements in planning consistency, constraint enforcement, and adaptive coordination during disruptions in diverse challenging scenarios.


\chapter{Beyond Computation: Consciousness Modeling}
\label{chap:Cocomo}
\markboth{}{}
\noindent{\textbf{Abstract}}
The $\CCM$ model proposes a computational solution to the challenge of incorporating ethical and emotional intelligence considerations into AI systems, with the aim of creating AI agents that combine knowledge with compassion. To reach this goal, $\CCM$ focuses on fairness, beneficence, empathy, non-maleficence, adaptability, and critical and exploratory thinking abilities. $\CCM$ employs consciousness modeling, reinforcement learning, and prompt template formulation to support these desired traits. By incorporating ethical and emotional intelligence considerations, a generative AI model can potentially lead to improved fairness, reduced toxicity, and increased reliability.

\section*{Introduction}
\label{sec:introUnknown}

Narrow AI, often referred to as System-1 AI following Kahneman's terminology \cite{kahneman2011thinking}, excels in executing well-defined, specific tasks through machine learning algorithms, including object recognition and language translation. However, this type of AI is not as effective in handling advanced generative AI functions that require reasoning, critical and exploratory thinking, or the modeling and regulation of emotions and behaviors. Such complex tasks go beyond the capabilities of System-1 AI, highlighting its limitations.

To address these limitations, researchers (e.g., Yoshua Bengio \cite{bengio2020future}) have proposed the development of system-2 AI, which aims to mimic human cognitive abilities. Several generative models have been developed since 2022 for text \cite{OpenAI-GPT3-2020, chatgpt, openai2023gpt4, LAMDA2022}, image \cite{DALL-E2022, LatentDiffusionModels}, and video generation \cite{VideoMeta2022}. However, these models face issues of bias, toxicity, robustness, and reliability \cite{RisksLLM2022, AIEthicsChatGPT2023}.

In this chapter, we propose a solution to address these concerns by modeling emotional intelligence and ethical guardrails within a generative AI model itself, drawing on insights from the study of human consciousness. We believe that addressing these issues outside of a generative AI model using human subjective feedback and reinforcement learning is equivalent to imposing censorship on user-generated content, which is a heuristic-based 
and non-scalable solution \cite{SocialMediaMisInfo2021,SiliconValues}.

Human consciousness is understood to manage both impulsive and reflective aspects of the unconscious, enabling compromises between competing goals and values. Emotions typically arise as impulsive reactions to stimuli, while ethics act as guardrails that help modulate or 
regulate emotion-steered motivations to sin. Developing a grasp of how human consciousness functions, not necessarily in physical terms but at least functionally, can offer vital insights for crafting a regulatory mechanism within a LLM. This mechanism would direct linguistic behavior and shape the linguistic features employed to achieve specific goals.

The nature and origin of consciousness have been studied for centuries, resulting in various theories, including the global workspace theory \cite{baars1988cognitive}, integrated information theory \cite{tononi2004information,tononi2015IIT,tononi2012phi}, neural correlates of consciousness approach \cite{crick1990towards,koch2004quest}, and attention schema theory \cite{graziano2016attention,graziano2013consciousness}, among others. These studies of consciousness provide valuable insights for architecting system-2 AI.

This chapter defines system-2 AI's desired traits and capabilities using a functionalist approach to consciousness. These include knowledge, fairness, beneficence, non-maleficence, empathy, adaptability, transparency, and critical and exploratory thinking abilities.
While this list is not exhaustive, it provides a starting point for developing ethical guardrails and emotional intelligence in AI systems. Depending on the context and application of AI, additional ethical considerations or modifications to these principles may be necessary.

To embody these capabilities and principles, we introduce the Computational Consciousness Model (\textsc{CCM}), which leverages priority-based scheduling, reward-based optimization, and Socratic dialogues. \textsc{CCM} offers customization based on cultural and individual requirements through adaptive prompt templates \cite{SocraticIEEECCWC2023,PromptSurvey2023}, and facilitates the transition between unconsciousness and consciousness states through a multi-level feedback scheduler and interrupt mechanism.

To enable emotion and behavior modeling and regulation, as well as critical and exploratory thinking, \textsc{CCM} interacts with large language models\footnote{Recent studies, such as \cite{bommasani2022opportunities}, propose referring to these multimodal pre-trained models as foundation models.} \cite{OpenAI-GPT3-2020, BioGPT2022,chatgpt,openai2023gpt4,LAMDA2022} using interactive question-answer-based dialogues. Furthermore, a reinforcement learning module maps external values and rewards to internal task-scheduling priorities.

\textsc{CCM} has the potential to support the development of adaptive computational consciousness that integrates knowledge and compassion, and models emotional intelligence for generative AI systems. This approach could significantly benefit humanity and society.

\chapter{A Retrospective and Adaptive Framework to Improve LLMs}
\label{chap:RAFEL}
\markboth{}{}
\noindent{\textbf{Abstract}}
$\Rafel$ is a retrospective and adaptive framework designed to benchmark private Large Language Models (LLMs) against teacher LLMs, identifying discrepancies in responses. Following the initial benchmarking, $\Rafel$ categorizes these discrepancies into four distinct categories, based on cognitive levels and types of errors. Subsequent phases involve a detailed diagnosis and deep-probing to uncover the root causes behind each category of discrepancy. Teacher LLMs play a crucial role in interrogating the private LLM, shedding light on the subtleties of its performance issues. With a clear understanding of the symptoms and their underlying causes, $\Rafel$ prescribes targeted remedies, accompanied by recommendations for relevant data sources to enhance the private LLM's performance via either fine-tuning, RAG, or both. Empirical studies validate $\Rafel$'s effectiveness in diagnosing and enhancing the capabilities of localized LLMs.

\section*{Introduction}
\label{sec:intro11}

The emergence of Large Language Models (LLMs) like GPT \cite{openai2023gpt4} and Gemini \cite{geminiteam2023gemini} has significantly advanced the field of natural language processing, enabling the generation of text that closely mimics human writing and offers deep insights across varied domains. Despite their transformative potential, the deployment and scalability of these models pose considerable computational and data challenges. A practical response has been the fine-tuning of medium-sized, open-source models such as LLaMa \cite{touvron2023llama} for specialized needs, allowing organizations to strike a balance between performance and feasibility, while also prioritizing data privacy and model customization for unique applications.

The shift towards using privately fine-tuned or locally deployed LLMs brings about essential management and technical challenges, vital for corporate strategy, governance, and innovation. This chapter explores the technical challenges of this shift, including:

\begin{itemize}[leftmargin=1.2em, topsep=-.05em, parsep=-.05em]
\item Justifying the choice of private LLMs over public counterparts by establishing relevant performance metrics and benchmarks for these specialized models.
\item Conducting in-depth error analysis to pinpoint the root causes of performance issues in private LLMs, ensuring targeted and effective remediation strategies.
\item Identifying specific, high-quality data crucial for the fine-tuning of private LLMs, aimed at enhancing their accuracy and domain relevance.
\item Implementing Retrieval-Augmented Generation (RAG) to dynamically incorporate external, updated knowledge sources, improving the model's responsiveness and breadth of knowledge.
\item Exploring hybrid models that leverage the strengths of both public and private LLMs to achieve enhanced performance and greater adaptability to new data and domains.
\end{itemize}

We introduce $\Rafel$, a framework designed for the retrospective and adaptive enhancement of LLMs, addressing these technical challenges. $\Rafel$ strategically balances cost and performance by incorporating sophisticated diagnostic algorithms. These algorithms effectively identify and address the root causes of inefficiencies, ensuring that solutions are economically viable.

$\Rafel$ employs advanced benchmarking metrics across cognitive levels, providing a thorough LLM performance assessment. Central to its diagnostics are two key algorithms: DIAG, for non-invasive\footnote{Non-invasive methods evaluate without interacting with the LLM's internal data, whereas invasive methods directly engage with the LLM, accessing potentially sensitive data.} evaluation, and PRBE for thorough, invasive probing. This combination allows $\Rafel$ to detect and understand both surface-level and deep-seated performance issues, facilitating targeted data source acquisition for enhancement.

$\Rafel$ is proficient in creating targeted, effective remediation strategies, ensuring data privacy and security, validated through real-world data studies. The novelty claims of $\Rafel$ include:

\begin{enumerate}[leftmargin=1.2em, topsep=-.2em, parsep=-.2em, label=\arabic*.]
\item \textit{Deep Probe with Cognitive and Error Type Analysis}: $\Rafel$ goes beyond traditional error rate analysis by deeply probing into the LLM's responses, categorizing errors within cognitive levels (recollection, comprehension, analysis, reasoning) and types (hallucination, biases), enabling a deep understanding of the model's performance issues.
\item \textit{Fine-grained, Precise Data Augmentation}: Contrasting with the conventional manual search for coarse-grained data augmentation, $\Rafel$ identifies the
required data and performs a more precise and relevant data enhancement that directly addresses the identified cognitive and error type deficiencies.
\item \textit{Dynamic Remediation Playbook}: RAFEL dynamically adjusts its remediation strategy based on real-time analysis of data and errors, akin to adapting tactics in sports, ensuring the most effective and appropriate intervention is applied.
\end{enumerate}

\chapter{Discovering Insights Beyond the Known}
\label{chap:BeyondUnknown}
\markboth{}{}
\noindent{\textbf{Abstract}}
Human knowledge, vast as it is, often falls short in grasping intricate interdisciplinary domains fully. In contrast, foundation models like GPT-4, endowed with extensive multidisciplinary knowledge, can potentially bridge this gap. Significantly, we leverage the vast expanses of GPT-4's knowledge, banking on its ability to frame questions that might elude human intuition, thus paving the way for the emergence of fresh insights and potentially novel knowledge. In this study, we convened a unique committee comprising a moderator (the authors) and two GPT-4 agents. The dialogue is ignited by the ancient narrative of Adam and Eve, setting the stage for a rich exchange between the GPT-4 agents. This conversation derives from the age-old tale, as the agents investigate three intertwined domains: the significance of myths in ecological interpretation, the intricate ethical and philosophical quandaries surrounding AI, and the enigmatic realm of the human brain as complemented by technology. This dialogue not only unveils captivating insights but also underscores the indispensable value of interdisciplinary exchanges. Foundation models, as demonstrated, can catalyze such dialogues, equipping us to traverse expansive knowledge landscapes and explore domains previously beyond human comprehension.

\section*{Introduction}

In our recent study on GPT-4 \cite{Bbubeck2023sparks}, we 
observed that GPT-4 along with analogous foundation
models, manifests a {\em polydisciplinary} capacity \cite{SocraSynthChangCSCI2023}. (For 
clarity, we use ``GPT-4'' to collectively refer to these 
foundation models, given that our experiments are centered 
on GPT-4.) Trained on a vast spectrum of topics from varied 
sources, GPT-4 stands apart from human specialists. Such 
specialists, while deeply knowledgeable in their specific 
fields, often lack a broad understanding outside their 
particular domain. In contrast, GPT-4 processes knowledge 
without being tethered to domain boundaries. It doesn't 
compartmentalize a query strictly as a ``physics question'' 
or a ``philosophy question,'' but crafts an integrated 
response, drawing from its multidisciplinary training data.

From a perspective of sheer knowledge breadth, GPT-4 arguably outpaces the average human. Its exposure to an enormous volume of documents endows it with a repository potentially wider than most human counterparts. However, volume isn't synonymous with depth. True depth often stems from intangible intuitions, insights, personal experiences, and cultural contexts. Considering GPT-4 lacks evolutionary experiences\textendash ranging from survival instincts to the full spectrum of human emotions\textendash we must ask: Can GPT-4 produce literature that deeply resonates with human sensibilities?

This study aims to ascertain if the polydisciplinary attributes of GPT-4 can generate insights that transcend standard human perspectives. We divide our research into two avenues: first, exploring the potential of GPT-4 to reveal ``unknown unknowns,'' and second, assessing its aptitude for crafting emotionally impactful literature. This chapter examines the former, utilizing the universally recognized biblical tale of Adam and Eve and their consumption of the forbidden fruit as a common thematic foundation. Through this exploration, we aim to uncover viewpoints potentially beyond the realm of typical human cognition.

Our methodology revolves around orchestrating a dialogue between multiple GPT-4 agents. Within the experimental framework, a moderator (represented by the authors) sets the initial intent and context for the conversation. The number of participating agents and their underlying foundation models can be adjusted as needed. In this study, our committee consists of two agents based on the GPT-4 model, referred to as GPT-A and GPT-B. Once initialized, the agents engage in conversation autonomously, with minimal moderation (discussed shortly). The resultant dialogue is thoroughly analyzed to discern conversational patterns and depth of content. This in-depth examination facilitates the identification of diverse themes the GPT-4 model gravitates towards. Our underlying hypothesis posits that the discourse and exchanges between these agents can unearth insights\textendash ``unknown unknowns''\textendash that were previously elusive to human understanding.

While the polydisciplinary capabilities of GPT-4 offer an unparalleled breadth and depth exceeding that of the moderator, the role of the moderator remains indispensable. This role channels the ``exploratory'' nature of the conversation, guiding it towards predefined objectives and ensuring its convergence within a set time frame. In this experiment, the initial spark for the dialogue is the narrative of Adam and Eve. Without prompting, the agents autonomously suggest probing the story from ten unique perspectives. Yet, after a series of exchanges, GPT-B expresses a keen interest in delving deeper. Following this, in collaboration with both agents, the moderator narrows down the scope of the dialogue to three key topics: ecological interpretation, philosophical exploration, and the neuroscientific angle. The intricate dialogues spanning these three domains\textendash namely AI interwoven with Ecology, Neuroscience coupled with AI, and Neuroscience meshed with Ecology\textendash are indeed engrossing. Throughout the discussion, both agents present a multifaceted perspective, shedding light on the diverse interpretations of the Garden of Eden, both prior to and following its seminal event. In the final stretch, the moderator verifies with both agents if they are poised to transition into the conclusion phase.

While our research unveils fascinating insights, it's essential to acknowledge several inherent limitations and constraints:

\begin{enumerate}[leftmargin=1.2em, topsep=-.2em, parsep=-.2em, label=\arabic*.]
    \item Model Training and Bias: GPT-4, akin to other machine learning models, is informed by pre-existing datasets. Therefore, the viewpoints, knowledge, and biases ingrained in this data can shape its outputs. It implies that GPT-4's responses might echo the historical and cultural biases present in the data upon which it was trained.
    \item Interactivity Limitation: Conversations between two GPT-4 agents essentially access the same foundational knowledge. Consequently, while the discourse may encompass a range of viewpoints due to query processing, it won't yield wholly novel information. To glean potentially varied insights, it could be beneficial to facilitate dialogues between different iterations of GPT (like GPT-3 and GPT-4) or even entirely distinct foundation models such
    as LaMDA by Google and LLaMA by Meta AI.
    \item Interpretation Subjectivity: Analysis of GPT-4's dialogues is susceptible to the prism of human interpretation. As such, different analysts might extract diverse conclusions from identical data sets.
\end{enumerate}

We emphasize that the core intent of our study was experimental, rooted in the notion that a foundation model such as GPT-4 may offer knowledge both wider in scope and deeper in intricacy than most human capacities. While we've shed light on its potential strengths, we remain cognizant of the inherent risks and limitations. Our work's primary novelty lies in pioneering a discourse-driven approach with AI agents. Significantly, we leverage the vast expanses of GPT-4's knowledge, banking on its ability to frame questions that might elude human intuition, thus paving the way for the emergence of fresh insights and potentially novel knowledge.

\bibliographystyle{unsrt} 
\bibliography{EVINCE,EdwardChang,RefBookletAphorisms,References-2,RLHF,Similarity,Rafel,SocraHealth,SocraPedia,SocraPlan,Emotions,RobustR,Workflow,TemporalPlanning,TSP,Aphorisms}

\chapter*{Appendix X: Online Chapters}
\label{chap:DebateWars}
\addcontentsline{toc}{chapter}{Appendix: Online Chapters} 
\markboth{}{}


The following three chapters are available at \textsc{SocraSynth.com}.  

\section*{\textbf{SocraPlan: SocraSynth for Sales Planning}}
\noindent{\textbf{Abstract}:}
SocraPlan introduces a sophisticated methodology that utilizes the capabilities of multiple Large Language Models (LLMs) for strategic sales planning in today's dynamic sales environment. This approach tailors sales playbooks to the unique needs and contexts of each customer by harnessing the power of Generative AI (GAI). Its primary objectives are to enhance customer satisfaction through a deep understanding of their specific requirements, refine sales strategies with targeted market analysis, and increase the efficiency of the sales process. SocraPlan sets itself apart with a collaborative and debate-driven framework that engages multiple LLMs, enabling a depth of analysis, adversarial reasoning, and strategy formulation that surpasses traditional AI-based approaches focused solely on data analytics. As a result, SocraPlan emerges as a pioneering tool in AI-driven sales strategies, delivering customized, effective solutions for complex sales planning challenges and facilitating more successful deal closures.
\\ \newline
\section*{\textbf{LLMs for Financial Planning and Analysis}}
\noindent{\textbf{Abstract}:}
This paper elucidates the potential of leveraging large language models (LLMs) in the meticulous analysis of financial statements for the purpose of financial planning and analysis (FP\&A). We commence by detailing a representative workflow encompassing the genesis of an FP\&A report, inclusive of its structural outline and prerequisite data. This is succeeded by a delineation of the diverse data sources, which span primary financial statements, supplemental internal datasets, and external data from industry specific and governmental sources. Amid the diverse repertoire of reports within FP\&A, we spotlight the generation of a ``financial health assessment'' report for a company as the focal point of our case study. Our methodology uniquely harnesses the strengths of LLMs, employing the ingenious Socratic Synthesis method to enhance the analysis and interpretative capabilities, thereby offering a more in-depth understanding of the data at hand. This approach not only accentuates the richness of the insights derived but also underscores the pivotal role of LLMs in advancing the realm of FP\&A.
\\ \newline
\section*{\textbf{LLM Debate on the Middle East Conflict: Is It Resolvable?}}
\noindent{\textbf{Abstract}:}
On October $7^{th}$, a renewed conflict arose between Israel and Palestine. Recognizing the historical significance and contentious nature of the Israel-Palestine conflict, this white paper engages two LLM agents in a debate over the question: ``Is the conflict between Israel and Palestine resolvable?'' A human moderator facilitates the discussion, intervening minimally. Through this debate, the paper seeks to highlight both the potential and constraints of contemporary LLMs.



\backmatter

\chapter*{Author's Biography}
\addcontentsline{toc}{chapter}{Author's Biography} 

\textbf{Edward Y. Chang} has been an adjunct professor in the Computer Science Department at Stanford University since 2019. He previously served as the president of HTC DeepQ Healthcare (2012-2021), and as a director of research at Google (2006-2012), where he pioneered Web-scale image annotation (2008) and data-centric machine learning (2010), led initiatives in scalable machine learning, indoor localization, Google Q\&A, and recommendation systems. He was a visiting professor at UC Berkeley (2017-2020), focusing on surgical planning with virtual reality. Chang was a tenured professor of Electrical \& Computer Engineering at the University of California, Santa Barbara (1999-2006). He holds an MS in Computer Science and a PhD in Electrical Engineering, both from Stanford University. Beyond his technical education, he enrolled in more than ten courses in philosophy and literature, an intellectual breadth reflected in his publications.

Chang is a recipient of several awards, including the NSF Career award, Google Innovation award, US\$1M XPRIZE (AI for disease diagnosis) and the ACM SIGMM Test of Time award. He is a Fellow of both ACM and IEEE for his contributions to scalable machine learning and healthcare.
\pagestyle{empty}
\clearpage

\begin{titlepage}
Copyright \textcopyright\ \the\year\ by Edward Y. Chang\\

All rights reserved. No part of this publication may be reproduced, distributed, or transmitted in any form or by any means, including photocopying, recording, or other electronic or mechanical methods, without the prior written permission of the publisher, except in the case of brief quotations embodied in critical reviews and certain other noncommercial uses permitted by copyright law.\\

\noindent{ISBN: 978-1-962463-09-6 (Paperback)}  \\

\noindent{Library of Congress Cataloging-in-Publication Data}\\
Names: Edward Y. Chang, author.\\

\noindent{Title: Unlocking the Wisdom of LLM Collaborative Intelligence} \\
ISBN 978-1-962463-09-6 (Paperback) \\
Homepage: \url{http://infolab.stanford.edu/~echang} \\
Subjects: LCSH: Artificial Intelligence\\
Classification: LCC QA76.76.E95 | DDC 006.3--dc23\\
Imprint: SocraSynth.com (\url{http://socrasynth.com}) \\

\clearpage
\end{titlepage}

\end{document}